\documentclass[final]{cvpr}
\usepackage[utf8x]{inputenc} 
\usepackage[T1]{fontenc}    
\usepackage{url}            
\usepackage{booktabs}       
\usepackage{amsfonts}       
\usepackage{nicefrac}       
\usepackage{microtype}      
\usepackage{times}
\usepackage{epsfig}
\usepackage{graphicx} 
\usepackage{graphics}
\usepackage{grffile}

\usepackage{amsmath}
\usepackage{amssymb}

\usepackage{comment}
\usepackage{amsmath,amssymb} 
\usepackage{color}
\usepackage{array}
\usepackage[export]{adjustbox}
\usepackage{tabularx}
\newcolumntype{L}{>{\raggedright\arraybackslash}X}
\usepackage{caption, booktabs}
\usepackage{floatrow}
\usepackage[caption=false]{subfig}
\usepackage{placeins}
\usepackage{bm}
\usepackage{hhline}
\usepackage{xcolor}
\usepackage{multirow}
\usepackage{makecell}
\usepackage{footnote}

\usepackage{paralist, tabularx}
\usepackage{enumitem}
\usepackage[pagebackref=true,breaklinks=true,colorlinks,bookmarks=false]{hyperref}

\pdfoutput=1
\newcommand{\ALE}{\textit{ALE}}      
\newcommand{\RALE}{\textit{RALE}}      
\newcommand{\Ag}[1]{ALE_{\gamma}\hspace{-2pt}\left(#1\right)}
\newcommand{\Rg}[1]{RALE_{\gamma}\hspace{-2pt}\left(#1\right)}

\def\eqnvspace{{\vspace{-2mm}}}

\newcommand{\Paragraph}[1]{\vspace{1mm} \noindent \textbf{#1} \hspace{-0mm}}

\def\cvprPaperID{5458} 
\def\confYear{CVPR 2021}

\begin{document}

\title{Depth Completion with Twin Surface Extrapolation at Occlusion Boundaries}

\author{Saif Imran \hspace{1cm} Xiaoming Liu \hspace{1cm} Daniel Morris\\
Michigan State University\\
{\tt\small \{imransai, liuxm, dmorris\}@msu.edu}\\
{\small \url{https://github.com/imransai/TWISE}}
}

\maketitle
\begin{abstract}
  
Depth completion starts from a sparse set of known depth values and estimates the unknown depths for the remaining image pixels.  Most methods model this as depth interpolation and erroneously interpolate depth pixels into the empty space between spatially distinct objects, resulting in depth-smearing across occlusion boundaries.  Here we propose a multi-hypothesis depth representation that explicitly models both foreground and background depths in the difficult occlusion-boundary regions.  Our method can be thought of as performing twin-surface extrapolation, rather than interpolation, in these regions.  Next our method fuses these extrapolated surfaces into a single depth image leveraging the image data.  Key to our method is the use of an asymmetric loss function that operates on a novel twin-surface representation.  This enables us to train a network to simultaneously do surface extrapolation and surface fusion.  We characterize our loss function and compare with other common losses. Finally, we validate our method on three different datasets; KITTI, an outdoor real-world dataset, NYU2, indoor real-world depth dataset and Virtual KITTI, a photo-realistic synthetic dataset with dense groundtruth, and demonstrate improvement  over the state of the art. 
  
  \end{abstract}
  

\section{Introduction}
Depth completion problems involve estimating a dense depth image from sparse depth measurements of active depth sensors, often guided by a high-resolution modality; {\it e.g.}, RGB sensors. 
Solving depth completion has extensive applications, {\it e.g.}, scene understanding~\cite{schwarz2010lidar}, object shape estimation~\cite{cui2013algorithms}, and $3$D  object detection in autonomous driving~\cite{qi2017frustum}.  

Step-like object discontinuities are an inherent property of $3D$ scenes, and are challenging to model well with depth completion and depth super-resolution methods. 
It is important to maintain depth discontinuities to facilitate object shape and pose estimation. 
Most prior works rely on conventional regression losses for depth completion which, albeit promising results in depth accuracy, suffer from depth {\it smearing} and hence shape distortion of objects. 
While~\cite{imran2019depth} tackles this depth mixing problem, it has high computational and memory demand for accommodating many channels at high resolution. 
Another interesting work \cite{park2020non} learns non-local spatial affinity with spatial propagation network to eliminate smearing, but suffers from significant inference times and poor generalization due to sparse patterns (See Tab.~\ref{tab:eff_sparsity}).

\begin{figure}[t!]
\captionsetup{font=small}
    \scalebox{1.02}{
        \includegraphics[trim=98 15 142 5,clip,width=1.04\linewidth, center]{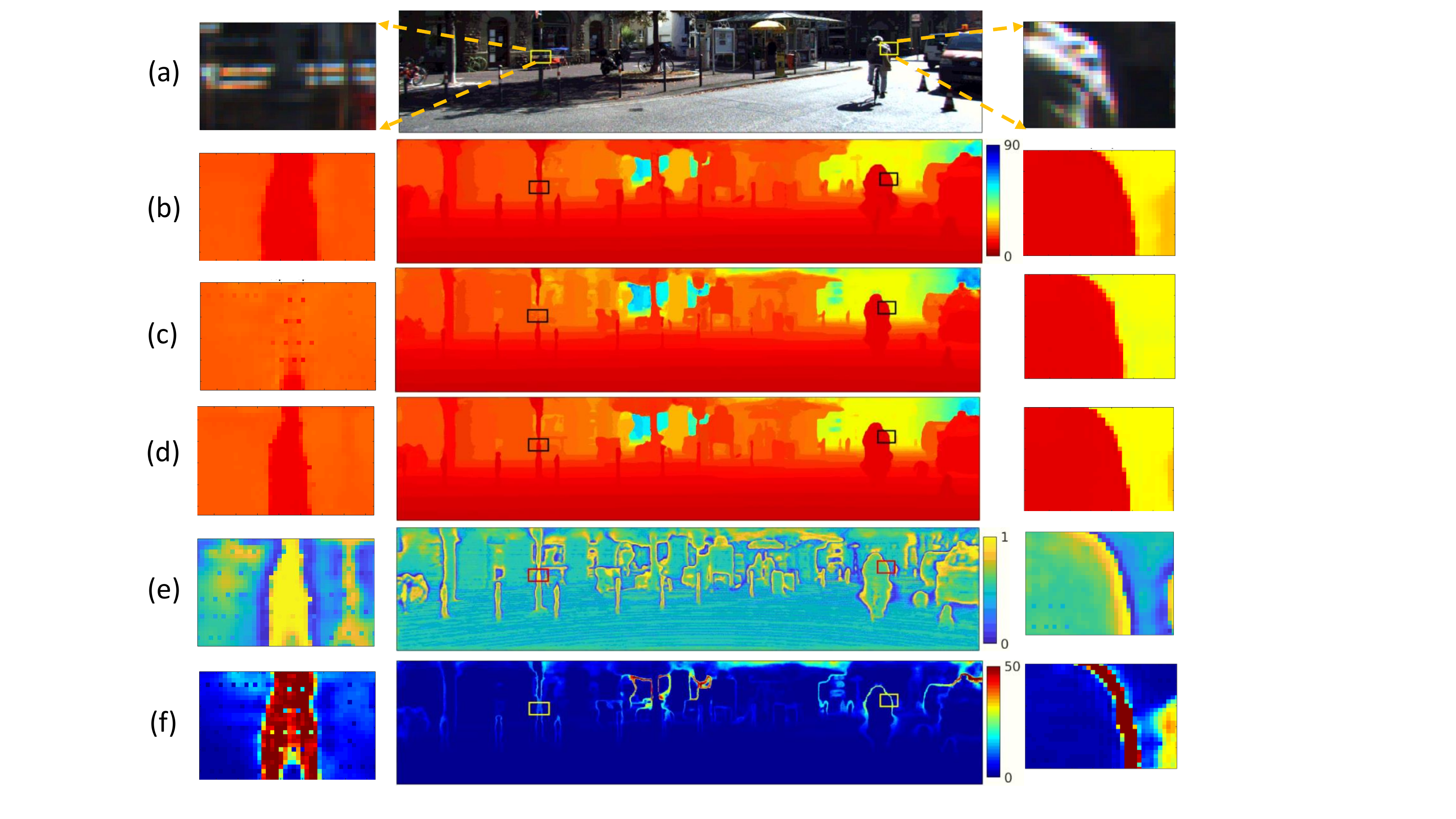}
        \vspace{-4mm}}
    \caption{\small Our depth completion algorithm can input LiDAR data and image (a), and extrapolate the estimates of foreground depth $d_1$ (b) and background depth $d_2$ (c), along with a weight $\sigma$ (e). Fusing all three leads to the completed depth (d). The foreground-background depth difference (f) $d_2-d_1$ is small except at depth discontinuities.\vspace{-4mm}}
    
    \label{fig:tease_fig}
\end{figure}

\begin{figure}[t!]
\captionsetup{font=small}
    \centering
     \scalebox{1.1}{
    \begin{tabular}{@{}c@{}c@{}c@{}c@{}}
        \includegraphics[trim=50 30 60 30,clip, width=0.22\linewidth]{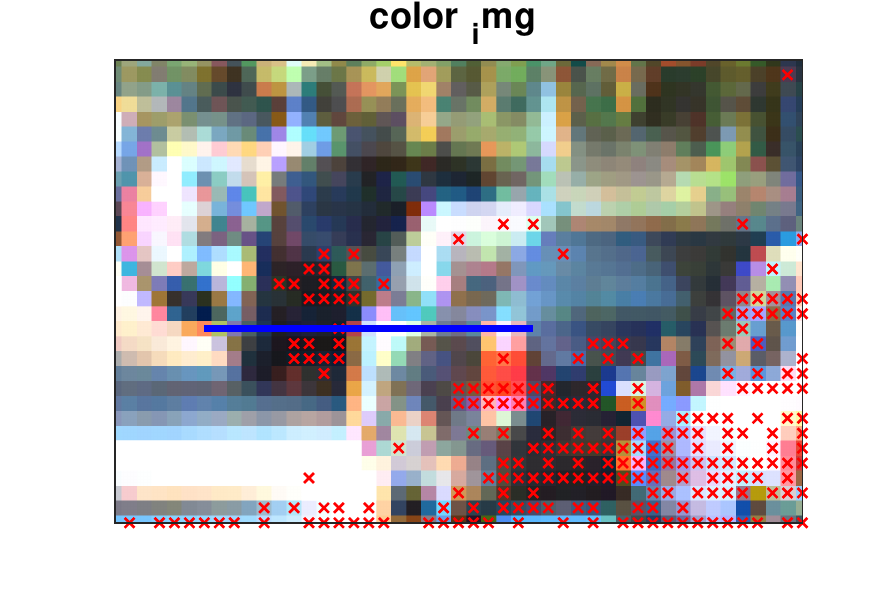} & 
        \hspace{0.1mm}
        \includegraphics[trim=50 30 60 30,clip,width=0.22\linewidth]{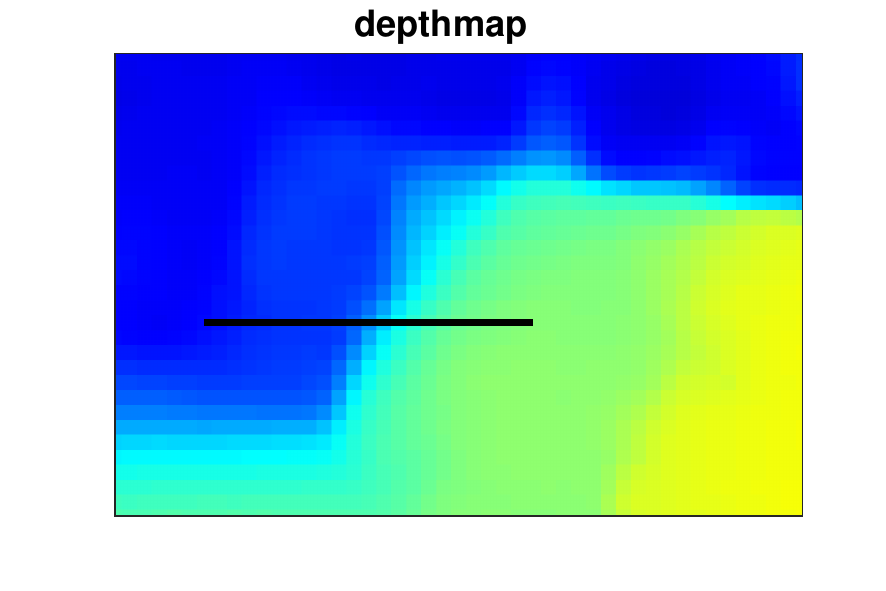} & 
         \includegraphics[trim=50 30 60 30,clip,width=0.22\linewidth]{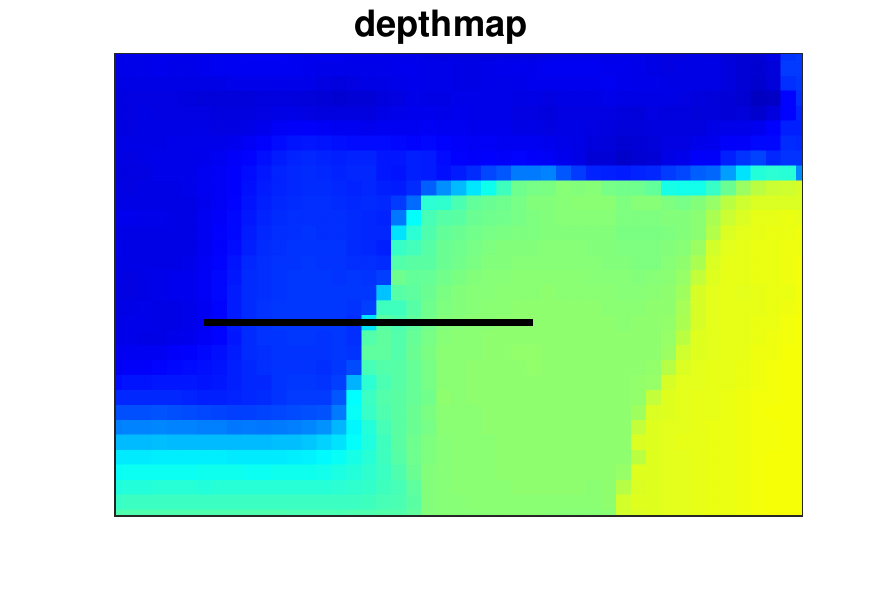} & 
         \includegraphics[trim=50 30 60 30,clip, width=0.22\linewidth]{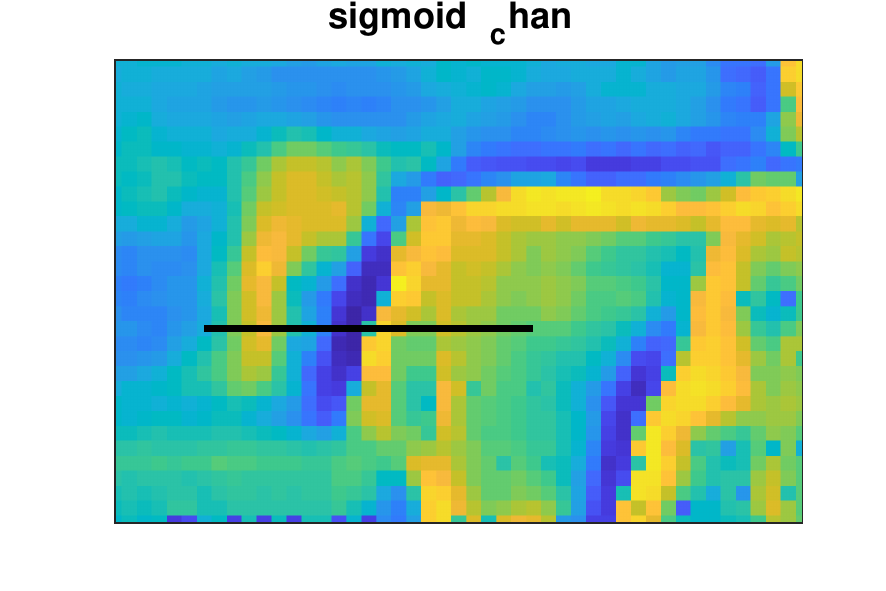} \vspace{-2mm}\\
        \footnotesize (a) & \footnotesize (b) & \footnotesize (c) & \footnotesize (d) \vspace{-1mm}  \\
        & \includegraphics[trim=0 2 22 5,clip,width=0.20\linewidth]{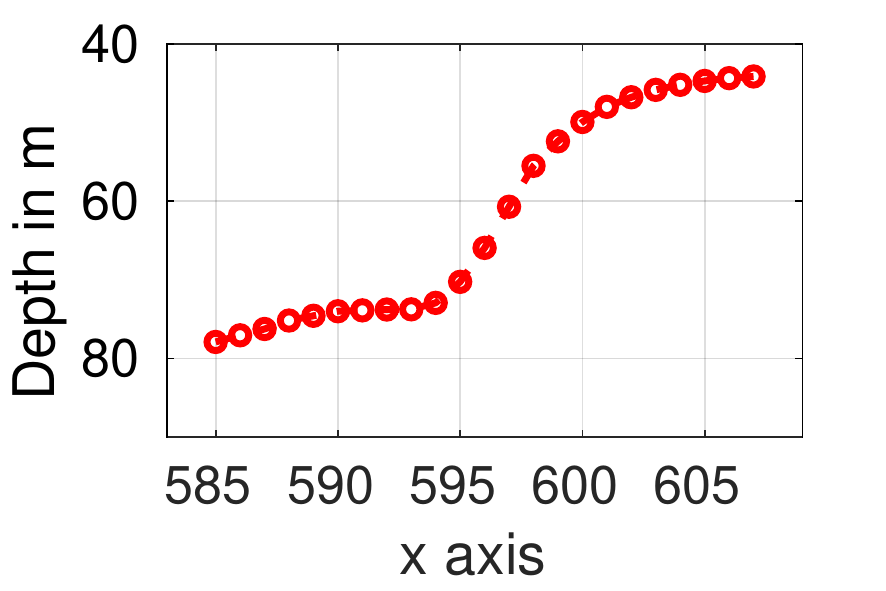} & \hspace{-0.02\linewidth}
        \includegraphics[trim=0 2 0 35,clip,height=33pt, width=0.28\linewidth]{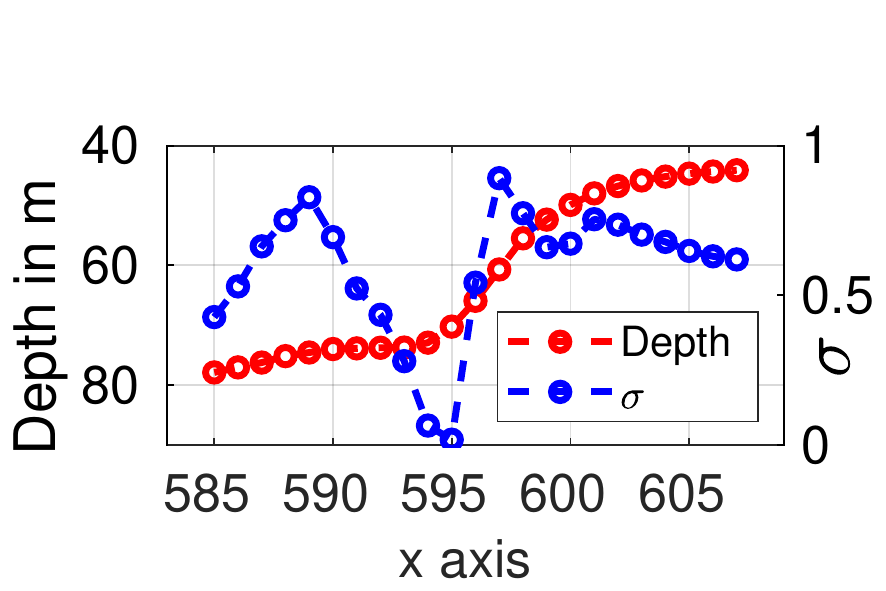} &
        \includegraphics[trim=0 2 22 5,clip,width=0.20\linewidth]{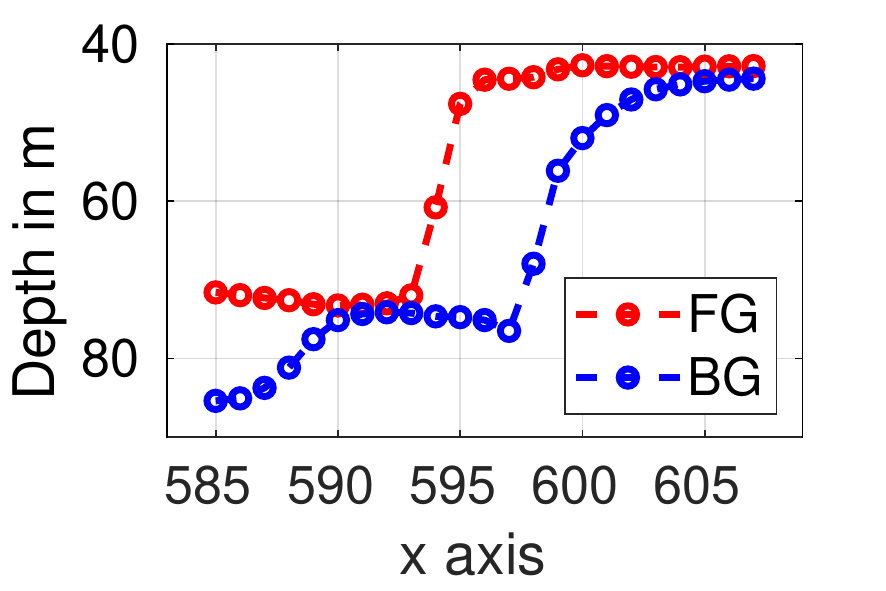} \vspace{-2mm}\\
        & \footnotesize (e) & \footnotesize (f) & \footnotesize (g)
        \end{tabular}
        }
        \vspace{-2mm}
    \caption{\small \textbf{Depth smearing across boundaries}. We show the ground truth depth (colored red) overlaid on an image (a),  depths estimated by the SoTA method~\cite{Li_2020_WACV} (b), our fused depths (c),  our estimated weights $\sigma$ (d),  a depth slice of~\cite{Li_2020_WACV} (e),  fused depth and $\sigma$ slices (f), and foreground and background slice (g). Our extrapolation ability in (g) results in the sharp depth boundary in (f), rather than the smeared depth in (e). \vspace{-2mm}}
    \label{fig:dep_smearing}
\end{figure}

One fundamental challenge of recovering depth discontinuity is that pixels at boundary regions suffer from ambiguity as to whether they belong to a foreground depth or background depth. 
Some methods seek to reduce the impact of ambiguities through intelligent regularization of the loss function \cite{hu2019revisiting} or ranking loss \cite{xian2020structure} at potential occlusion boundaries; and others explicitly train detectors for depth discontinuities~\cite{huang2019indoor} via full dense ground truth data.  
Unfortunately large scale datasets like KITTI have only partial dense ground truth depth (see Fig.~\ref{fig:dep_smearing} (a)) which is sparse on boundaries, making it difficult to explicitly label occlusion boundaries. Some recent works have leveraged prior information, {\it e.g.}, estimated semantic maps \cite{zhu2020edge, guizilini2020semantically, wang2020sdc} and estimated depth maps \cite{ramamonjisoa2020predicting} for object boundary recovery/refinement.  
Our approach instead seeks to explicitly model ambiguity and leverage it in depth completion.
We noted that \emph{Depth coefficients}~\cite{imran2019depth} can also model ambiguities on account of its non-parametric probability distribution, but maintaining many high-resolution channels is computationally and memory expensive, and also suffer from the binning resolution.  
Instead of using multiple channels with binning, our method, named TWIn-Surface Estimation (\emph{TWISE}), uses a two-surface representation which is much more efficient and can {\it explicitly} model ambiguity by finding difference between the twin surface depths. 
We believe that naturally encoding the foreground and background pixels at the boundary would enable the effective learning of the step-wise discontinuity with lower memory and computational requirement. 
  
In order to train a twin-surface estimator, we propose a pair of asymmetric loss functions that naturally bias estimates toward foreground and background depth surfaces.  The asymmetry in the losses are key to separation of foreground and background depths at ambiguous pixels.  We also incorporate a fusion channel that automatically combines the foreground and background depths into a final depth estimate for each pixel, by selecting a foreground/background depth at the ambiguous regions and mixing the two depths at non-ambiguous regions.

Of particular concern is the lack of dense and reliable ground-truth depth data in outdoor scenes needed for accurate evaluation of depth estimates. KITTI, a realistic outdoor scene dataset, offers semi-dense ground-truth, created by accumulating LiDAR points but suffers from noisy depth samples (outliers) at boundaries and dynamic objects \cite{uhrig2017sparsity}. Indoor dataset like NYU2 provides dense GT only by using some colorization techniques that can cause smoothing at object boundaries. 
Currently the preferred evaluation metric of choice for ranking depth completion methods is RMSE. 
In this paper, we study the effects of outlier noise present in ground-truth data on RMSE and note that MAE is a more consistent metric for both cases of noisy and clean ground-truth, as validated on the synthetic VKITTI dataset. 

The contributions of this paper are as follows:

\begin{compactitem}
\item We propose a twin-surface representation that can estimate foreground, background and fused depth.

\item We adopt a pair of assymmetric loss functions to explicitly predict foreground-background object surfaces. 

\item We validate our theory in KITTI, a challenging outdoor scene dataset for depth completion, and show the superiority of our method on several metrics, and also show it generalizes well to variable sparsity and offers competitive inference times over the SoTA. 

\item We suggest that in presence of outliers, MAE is a more consistent metric to rank methods compared to RMSE, and we validate this claim with extensive experiments in VKITTI, a synthetic dataset for urban driving scenario.  
\end{compactitem}

\section{Related Works}
\Paragraph{Depth Completion}
Deep neural networks (DNNs) have been applied to the depth completion problem, in works such as Sparse-to-Dense~\cite{ma2019self}, DDP~\cite{yang2019dense}, and Spade RGBsD~\cite{jaritz2018sparse}. These works show that by using standard encoder-decoder architecture (ResNet and MobileNet), it is possible to improve depth estimation accuracy via regression losses like $L_2$, $L_1$ and inverse $L_1$ losses. Deep-Lidar~\cite{qiu2019deeplidar} estimates surface normal and dense depth using multiple DNNs to assist in further fine-tuning dense depth. 
Both~\cite{qiu2019deeplidar} and~\cite{yang2019dense} rely on synthetic data and various labels for learning depth representations. 
Recently, works have opted to optimize depth using $3$D geometric constraints like depth-normal consistency~\cite{AAAI1816421,xu2019depth} to improve depth completion. Xu {\it et al.} create geometric consistency between the surface normal and depth in $3$D, but use another refinement network for improved depth estimation~\cite{xu2019depth}. 
Another recent trend is to learn spatial propagation of pixels in $2$D depth space for depth completion problems in fixed~\cite{cheng2019cspn++} or variable receptive field~\cite{park2020non,  xu2020deformable}. Although results are highly encouraging, these methods suffer from poor inference times and generalizability on variable sparsity. 
Researchers have also looked into learning $3$D features for depth completion using continuous convolution in $3$D space~\cite{chen2019learning}, point cloud completion~\cite{xiang20203ddepthnet}, $3$D graph neural networks~\cite{xiong2020sparse} for dynamic construction of local neighborhood regions.

\Paragraph{Depth Representations}
Depth maps, as $2.5$D representations, have been used for RGBD fusion and instance segmentation~\cite{shao2018clusternet,gupta2014learning}.  They naturally encode sensor viewing rays and adjacency between points. They are compact representations and their regular grids can be processed with CNNs in an analogous way to image super-resolution~\cite{tai2017image,tai2017memnet}. This is the representation of choice for colorization techniques and fusion~\cite{Silberman:ECCV12} as well as depth completion.

We propose a $2$-layered representation of depth to model occlusion boundaries. The concept of layered representation of depth has been well known in graphics community.
LDIs (Layered Depth Images) are first proposed by Shade {\it et al.}~\cite{shade1998layered} as  intermediate representation for efficient image-based rendering. 
These are gathered by accumulating depth values via z-buffering from multiple depth images of nearby view points. Tulsiani {\it et al.}~\cite{tulsiani2018layer} infer $2$-layered depth representation (recovering depth of visible and non-visible scene) from a single input image by learning view-synthesis from multiview camera guided supervision. 
Hedman {\it et al.}~\cite{Casual3D2017} propose a $3$D photo reconstruction algorithm that builds multi-layered geometric representation of the scene by warping several depth maps and stitching color and depth panoramas for front and back-scene surfaces. In all these cases, multi-layered representation is constructed/learned from multi-camera viewpoints/depthmaps of the scene. In our case, we estimate these $2$-layered representation on a single camera viewpoint with our proposed loss functions.

\Paragraph{Loss Functions in Depth Completion}
A key component of depth completion is the choice of loss functions.
Recent work has explored loss functions including $L_2$~\cite{chen2018estimating,ma2019self}, $L_1$~\cite{mal2018sparse}, inverse-$L_1$~\cite{jaritz2018sparse}, Huber loss~\cite{chen2019learning} and Softmax loss on depth~\cite{liao2017parse}. Another elegant way is to use combination of $L_1$ + $L_2$ \cite{park2020non}, which can leverage the benefits of both $L_1$ and $L_2$ losses.
While these loss functions can achieve low error on metrics including RMSE, MAE, iMAE, often it comes at the cost of smoothing depth estimates across object boundaries. 
In addition to the aforementioned losses, people increasingly use Chamfer distance on point cloud~\cite{xiang20203ddepthnet}, depth-normal constraint~\cite{xu2019depth}, Cosine loss~\cite{qiu2019deeplidar}, in a multi-learning framework to improve depth completion accuracy.  Nevertheless, smoothing across sharp boundaries remains a concern in many of these methods. Imran {\it et al.}~\cite{imran2019depth} show that cross-entropy (CE) loss can generate sharp boundaries, although performs worse in the RMSE metric.

We learn foreground and background depth by proposing two assymetric loss functions, and the final depth using a fusion loss. The assymetric loss function has been used in Vogel {\it et al.}~\cite{vogels2018denoising}, for the  different purpose of denoising input images. We propose to use assymetric loss functions to learn biased estimators of FG/BG surface, and learn to select/blend (fusion loss) between FG/BG surface, and that, we claim, helps to recover depth discontinuity.

\section{Methodology}

Depth completion involves two quite different challenges which can be at odds.  The first is to interpolate missing pixel depths within objects leveraging nearby sparse depths.  The second is to accurately find the occlusion boundaries of objects and ensure that interpolated pixels belong to either the foreground or background object.  We propose a method that aims to perform both tasks well.  

Our approach divides depth completion into two simpler problems, each of which can be more easily learned by a network.  The first problem is depth interpolation without boundary determination.  Rather than estimating a single surface which must model step functions at depth discontinuities, Our key novelty is to estimate twin surfaces.  A foreground surface extrapolates the foreground object depth up to and beyond boundaries, while a background surface extrapolates the background depth up to and behind the occluding object.  
Then the second problem is to find the boundary and determine a single depth by fusing these two surfaces. 
We find the color image is particularly useful in aiding surface fusion. Both of these components are illustrated in Fig.~\ref{fig:dep_smearing}.  

\subsection{Ambiguities and Expected Loss}
\label{sec:ambiguity}

Ambiguities have a significant impact on depth completion, and it is useful to have a quantitative way to assess their impact.  Here we propose using the \textit{expected loss} to predict and explain the impact of ambiguities on trained networks. 

By an ambiguity we mean, not that there isn't a unique true solution, but rather that from a measurement it is difficult for the algorithm and/or human to decide between two or more distinct solutions.  
Ambiguity can be more formally defined as follows.
Given measurement data that sparsely samples the scene, the number of ambiguities is equal to the number of different true scenes, {\it i.e.} true depth maps in our case, that could have generated the sparse measurement.  This number depends on what variations occur in actual data.  For simplicity we treat each pixel ambiguity independently of other pixels, and so the ambiguities for a pixel are the possible depth values it could take that are consistent with the measurement.

We anticipate the level of ambiguity to vary across a scene.  For example, pixels on flat surfaces will be well-constrained by nearby pixels and have low ambiguity.  In contrast, pixels near depth discontinuities may have large depth ambiguity. There is often insufficient data from the depth image to decide whether the pixel is on the foreground or background. 

A corresponding color image can help resolve ambiguities as to which object a pixel belongs.  However, exactly how to leverage color images to resolve ambiguities in CNNs is one of the open challenges in depth completion.  
Our work aims to offer a solution to this problem by {\it explicitly} estimating ambiguities and resolving them {\it within} the network.

To assess the impact of ambiguities on our network, we build a quantitative model.  Consider a single pixel whose depth, $d$, we seek to estimate.  Next assume that the pixel has a set of ambiguities, $d_i$, each with probability $p_i$.  This probability measures of how likely it is that the ground truth will take the corresponding depth, given our modeled scene assumptions.  Now consider a loss function on the error for each pixel, $L(d-d_t)$, where $d_t$ is the ground truth depth.  The \textit{expected loss} as a function of depth is:
\begin{equation}
    E\{L(d)\} = \sum_i p_i L(d-d_i).
    \label{eq:eale}
\end{equation}
This expected loss is important because if a network is trained on representative data then it will be trained to minimize the expected loss. Thus by examining the expected loss we can predict the behavior of our network at ambiguities, and so justify the design of our method.

\subsection{Asymmetric Linear Error}
\begin{figure}[t!]
\captionsetup{font=small}
    \centering
    \scalebox{1.12}{
    \begin{tabular}{@{}c@{}c@{}c@{}}
      \includegraphics[trim=20 20 15 5,clip,width=0.25\linewidth]{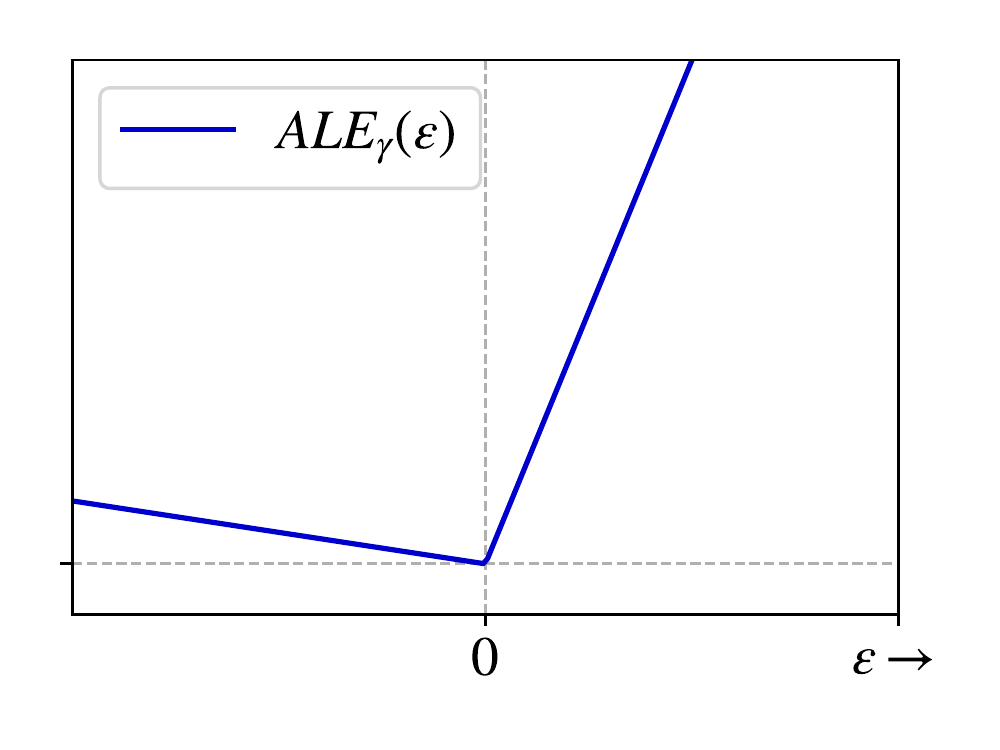}    &  
       \includegraphics[trim=20 20 15 5,clip,width=0.25\linewidth]{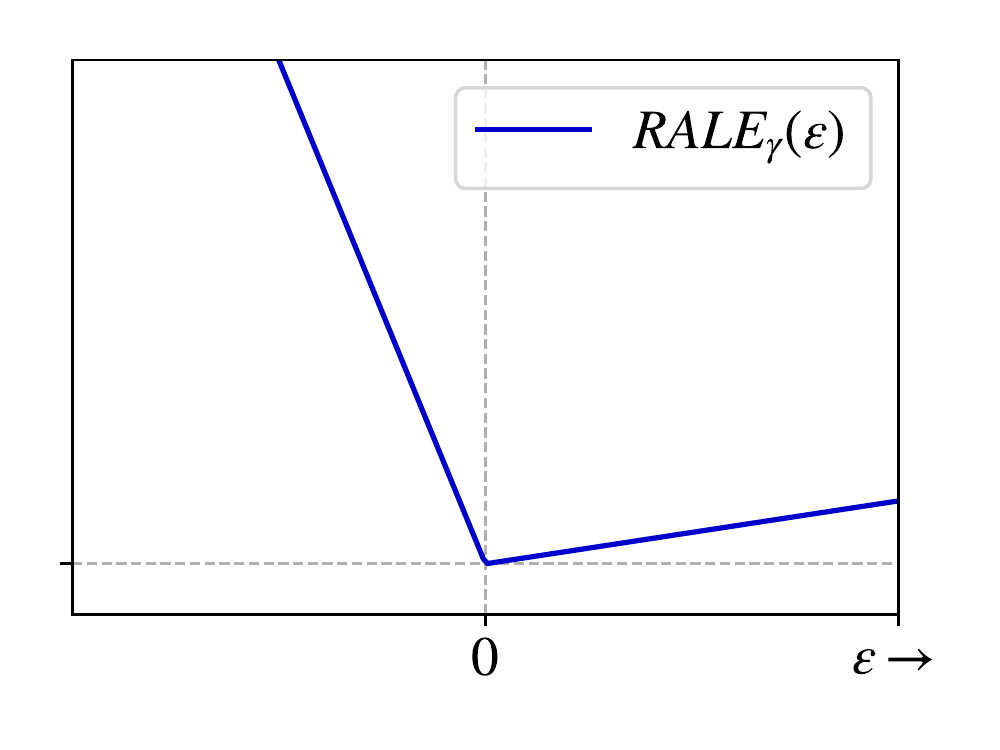} &
    \includegraphics[trim=20 19 15 17,clip,width=0.39\linewidth]{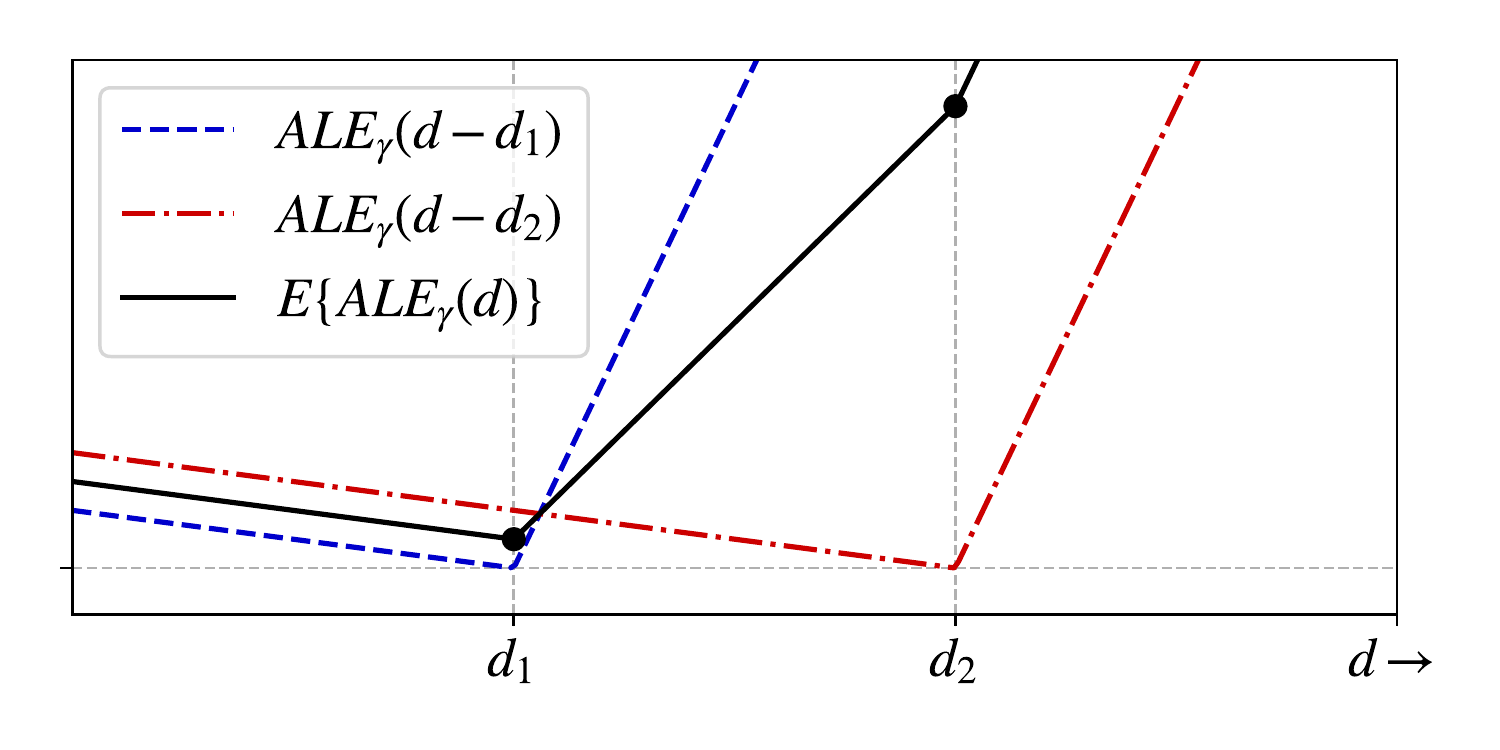} \vspace{-1.5mm}\\
       {\footnotesize (a)} &  {\footnotesize(b) }&  {\footnotesize(c)}
    \end{tabular}
    }
    \vspace{-2mm}
    \caption{ \small (a) The \ALE{} from Eq.~(\ref{eq:ALE}) is asymmetric around its minimum at the origin. (b) The \RALE{} from Eq.~(\ref{eq:RALE}) is a reflection of the \ALE{}. We use the \ALE{} for foreground surface estimation and the \RALE{} for background estimation.  (c) A pixel depth is shown with two ambiguities at depths $d_1$ and $d_2$ and probabilities $p1$ and $p1$ respectively. The black line shows the expected \ALE{} which is the probability-weighted sum of two \ALE{} functions, see Eq.~(\ref{eq:eale}).  The expected \ALE{} will have a minimum at one of the marked corners occurring at $d_1$ and $d_2$.  The minimum will be at $d_1$ if Eq.~(\ref{eq:gamma}) is satisfied, as it is in this case with $p_1=p_2$, and so acts as a foreground depth estimator. \vspace{-3mm}}
    \label{tab:ae_error}
\end{figure}

Our method uses a pair of error functions which we call the Asymmetric Linear Error (\ALE{}), and its twin, the Reflected Asymmetric Linear Error (\RALE), defined as: 
\begin{equation}
    \Ag{\varepsilon} = \mbox{max}\left(-\frac{1}{\gamma}\varepsilon, \gamma \varepsilon\right),
    \label{eq:ALE}
\end{equation}    
\begin{equation}
    \Rg{\varepsilon} = \mbox{max}\left(\frac{1}{\gamma}\varepsilon, -\gamma \varepsilon\right).
    \label{eq:RALE}
\end{equation}
Here $\varepsilon$ is the difference between the measurement and the ground truth, $\gamma$ is a parameter, and $\mbox{max}(a,b)$ returns the larger of $a$ and $b$.  The \ALE{} and \RALE{} are generalizations of the absolute error, and are identical to the absolute error when $\gamma=1$.  The difference is that the negative side of \ALE{} is weighted by $\nicefrac{1}{\gamma}$ and the positive weighted by $\gamma$.  The \RALE{} is simply the reflection of the \ALE{} over the $\varepsilon=0$ line.  Both are illustrated in Fig.~\ref{tab:ae_error} (a,b).

Note that if $\gamma$ is replaced by $\nicefrac{1}{\gamma}$, both the \ALE{} and \RALE{} are reflected.  Thus, without loss of generality, in this work we restrict $\gamma\geq 1$.

\subsection{Foreground and Background Estimators}

We make a further simplifying assumption in our analysis that there are at most binary ambiguities per pixel.  A binary ambiguity is described by a pixel having probabilities $p_1$ and $p_2$ of depths $d_1$ and $d_2$ respectively.  When $d_1<d_2$ we call $d_1$ the foreground depth and $d_2$ the background depth.  Such a binary ambiguity is likely to occur near object-boundary depth discontinuities.

To estimate the foreground depth we propose minimizing the mean \ALE{} over all pixels to obtain $\hat{d}_1$, the estimated foreground surface.  To predict the characteristics of $\hat{d}_1$ from a trained network at ambiguous pixels, we examine the expected \ALE{}, as shown in Fig.~\ref{tab:ae_error} (c).  This is piecewise linear and has two corners, one at $d_1$ and the other at $d_2$.  The lower of these will determine the minimum expected loss, and hence what an ideal network will predict.  Using Eqs.~(\ref{eq:ALE}) and (\ref{eq:eale}), we obtain expected losses: $L(d_1) = p_2(d_2-d_1)/\gamma$, and $L(d_2) = p_1 (d_2-d_1)\gamma$. From this it is straightforward to see $L(d_1)<L(d_2)$ when:
\begin{equation}
    \gamma > \sqrt{\frac{p_2}{p_1}}.
    \label{eq:gamma}
\end{equation}
This equation shows the sensitivity of the foreground estimator to $\gamma$; the higher $\gamma$, the lower the probability on foreground $p_1$ needed for the minimum to be at the foreground depth $d_1$.  

To estimate the background depth, $\hat{d}_2$, at boundaries we propose minimizing the expected \RALE{}.  The same analysis will apply to this as to the \ALE{}, and we obtain the same constraint on $\gamma$ as in Eq.~(\ref{eq:gamma}), except that the probability ratio is inverted.  

Fig.~\ref{fig:tease_fig} (b) shows an example foreground depth estimate, (c) the background depth  and (f) the depth difference.  We observe that at pixels far from depth discontinuities, as well as the sparse input-depth pixels, the foreground depth is very close to the background depth indicating no ambiguity.

\subsection{Fused Depth Estimator}
We desire to have a fused depth predictor that can do both interpolation and extrapolation at surfaces depending on ambiguous and non-ambiguous regions. The foreground and background depth estimates provide lower and upper bounds on the depth for each pixel.  We express the final fused depth estimator $\hat{d}_t$ for the true depth $d_t$ as a weighted combination of the two depths: 
\begin{equation}
    \hat{d}_t = \sigma\hat{d}_1 + (1-\sigma)\hat{d}_2.
    \label{eq:depth_est}
\end{equation}
where $\sigma$ is an estimated value between $0$ and $1$.  We use a mean absolute error as part of the fusion loss:
\begin{equation}
    F(\sigma) = |\hat{d}_t - d_t| = |\sigma\hat{d}_1 + (1-\sigma)\hat{d}_2 - d_t|.
\end{equation}
The expected loss for this is
\begin{equation}
 \begin{split}
  L_e(\sigma)= & E\{F(\sigma)\} =  p|\sigma\hat{d}_1 + (1-\sigma)\hat{d}_2 - d_1| + \\ & (1-p) |\sigma\hat{d}_1 + (1-\sigma)\hat{d}_2 - d_2|.     
 \end{split}
 \label{eq:fuse_loss}
\end{equation}

Here, $p$ = $p_1$, and $p_2$ = $1 - p$. This has a minimum at $\sigma=1$ when $p>0.5$ and a minimum at $\sigma=0$ when $p<0.5$.  Of course this assumes that depth is either $d_1$ or $d_2$.

Depth fusion occurs by optimizing the loss of Eq.~(\ref{eq:fuse_loss}) to predict a separate $\sigma$ for each pixel.  In this way our fusion step is an explicit determination of whether a pixel is foreground or background or a combination.  An example estimated $\sigma$ is shown in Fig.~\ref{fig:tease_fig} (e).

\subsection{Depth Surface Representation} 
We have developed three separate loss functions whose individual optimizations give us three separate components of a final depth estimate for each pixel.  Based on the characterization of our losses, we require a network to produce a $3$-channel output. Then for simplicity we combine all loss functions into a single loss:
\begin{equation}\eqnvspace
  \begin{split}
      L(c_1,c_2,c_3) = & \frac{1}{N}\sum_j^N( \Ag{c_{1j}} + \Rg{c_{2j}} \\
     & + F(s(c_{3j}))).  
  \end{split}
    \label{eq:channel_loss}
\end{equation} 
Here $c_{ij}$ refers to pixel $j$ of channel $i$, $s()$ is a Sigmoid function, and the mean is taken over all $N$ pixels.  We interpret the output of these three channels for a trained network as $c_1\rightarrow\hat{d}_1$, $c_2\rightarrow\hat{d}_2$ and $s(c_3)\rightarrow\sigma$, and combine them as in Eq.~(\ref{eq:depth_est}) to obtain a depth estimator $\hat{d}_t$ for each pixel.

\subsection{Implementation Details}
\Paragraph{Architecture}
This work presents novel loss functions linked to a multi-channel depth representation.  
These can be easily incorporated into a variety of network architectures with minimal change to the network.  
Specifically we selected the multistack network~\cite{Li_2020_WACV}, with the author-provided code.  
We choose this network due to its fast inference time, lower number of parameters than~\cite{ma2019self}, and its near-SoTA performance.  
The changes we made were three output channels and instead of one at each stacked hourglass network, and we use our loss function for the optimization. 
We used $64$ channels in the encoder-decoder network as that provided their highest performing results. More details are shared in the supplementary material.

\Paragraph{Training and Inference}
We followed the training protocol in~\cite{Li_2020_WACV} with multi-scale supervision on our $3$ channels. 
The total loss is a weighted sum of the multiple resolution losses $L_i$, where $L_1$ is the full resolution $3$-channel loss in Eq.~(\ref{eq:channel_loss}), $L_2$ is half-resolution and $L_3$ quarter resolution:
    $L = \omega_1 L_1 + \omega_2 L_2 + \omega_3 L_3$.
The multiscale stage training protocol sets $\omega_1 = \omega_2 = \omega_3 = 1$ during the first $10$ epochs, reduces $\omega_2 = \omega_3 = 0.1$, and continues to train for another $10$ epochs. For the last $10$ epochs we set $\omega_2 = \omega_3 = 0$ and complete training after $30$ epochs. Using Adam optimizer with an initial learning rate of $10e-3$ and decrease to half every $5$ epochs, we train a full sized image with gradient accumulated every $4$ samples in a batch.
We use PyTorch~\cite{paszke2017automatic} for our implementation.

\begin{table*}[t!]
\captionsetup{font=small}
    \centering
    \setlength\doublerulesep{0.5pt}
    \scalebox{0.83}{
    \begin{tabular}{|c|c|c|c|c|c|c|c|}
    \hline
    Method & MAE & RMSE & iMAE & iRMSE & TMAE~\cite{imran2019depth} & TRMSE~\cite{imran2019depth} & Infer.~time 
    (sec.) \\
    \hline
     Ma {\it et al.}~\cite{ma2019self} & $249.95$/$269.2$ & $814.73$/$878.5$ & $1.21$/$1.34$ & $2.80$/$3.25$ & --/$190.15$ & --/$297.48$ & $0.081$ \\
     Depth-Normal \cite{xu2019depth} & $235.17/236.67$ & $777.05/811.07$ & $1.79/1.11$ & $2.42/2.45$ & --/-- & --/-- & --\\
     DeepLidar~\cite{qiu2019deeplidar} & $226.50$/$215.38$ & $758.40$/\boldsymbol{$687.0$} & $1.15$/$1.10$ & $2.56$/$2.51$ & --/$162.75$ & --/$266.79$ & $0.097$ \\
     3DepthNet~\cite{xiang20203ddepthnet} & $226.2$/$208.96$ & $798.40$/$693.23$ & $1.02$/$0.98$ & $2.36$/$2.37$ & --/-- & --/-- & -- \\
     Uber-FuseNet~\cite{chen2019learning} & $221.19$/$217.0$ & $752.88$/$785.0$ & $1.14$/$1.08$ & $2.34$/$2.36$ & --/-- & --/-- & -- \\
     MultiStack~\cite{Li_2020_WACV} & $220.41$/$223.40$ & $762.20$/$798.80$ & $0.98$/$1.0$ & $2.30$/$2.57$ & --/$157.90$ & --/$270.15$ & \boldsymbol{$0.018$} \\
     DC-3co~\cite{imran2019depth} & $215.75$/$215.04$ & $965.87$/$1011.3$ & $0.98$/$0.94$ & $2.43$/$2.50$ & --/$141.67$ & --/$238.5$ & $0.112$ \\
     CSPN++~\cite{cheng2019cspn++} & $209.28$/-- & $743.69$/-- & $0.90$/-- &  $2.07$/-- & --/-- & --/-- & $0.200$\\
     DDP \cite{yang2019dense} & $205.40$/-- & $836.00$/-- & $0.86$/-- & $2.12$/-- & --/-- & --/-- & --\\
     NLSPN~\cite{park2020non} & $199.59/198.64$ & $\boldsymbol{741.68}/771.8$ & $0.84/0.83$ & \boldsymbol{$1.99$}/\boldsymbol{$2.03$} & --/$138.81$ & --/$248.88$ & $0.225$ \\ 
     \hline
     TWISE  & \boldsymbol{$195.58$}/\boldsymbol{$193.40$} & $840.20$/$879.40$ & \boldsymbol{$0.82$}/\boldsymbol{$0.81$} & $2.08$/$2.19$ & --/\boldsymbol{$131.60$} & --/\boldsymbol{$239.80$} & $0.022$  \\
     \hline
    \end{tabular}}
    \vspace{-2mm}
    \caption{\small Depth completion on the Test/Validation sets of KITTI, with $64$R LiDAR and RGB input (units in mm). }
    \label{tab:kitti_results}
\end{table*}

\vspace{6mm}
\begin{figure*}[t!]
\captionsetup{font=small}
    \centering
    \scalebox{0.95}{
    \begin{tabular}{@{\hskip-2.5mm}c@{}c@{}c@{}c}
         \sidesubfloat[]{\hspace{0.7mm}\includegraphics[trim=20 120 20 120,clip, width=0.248\linewidth]{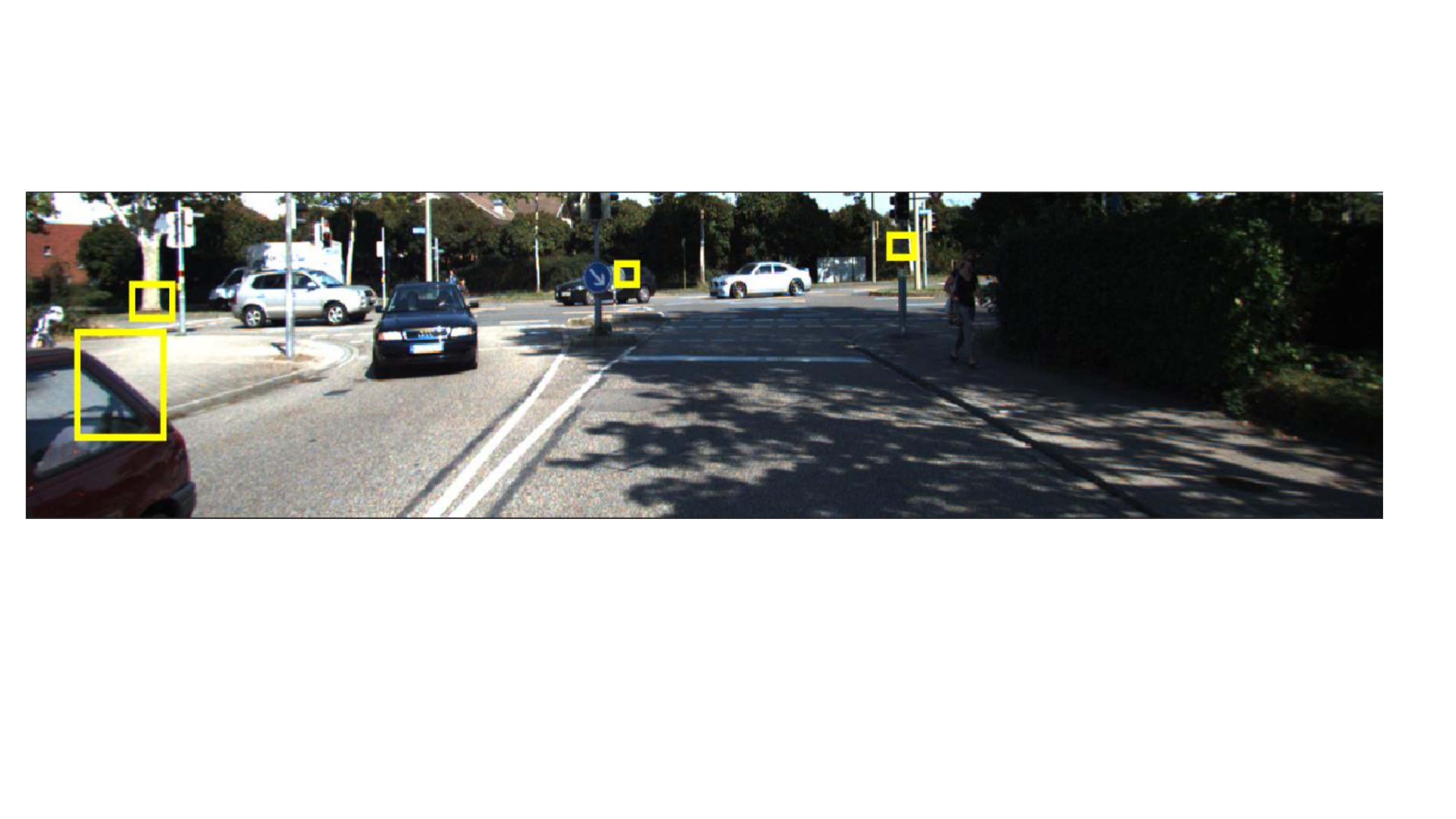}} & \sidesubfloat{\hspace{0mm}\includegraphics[trim=20 120 20 120,clip, width=0.248\linewidth]{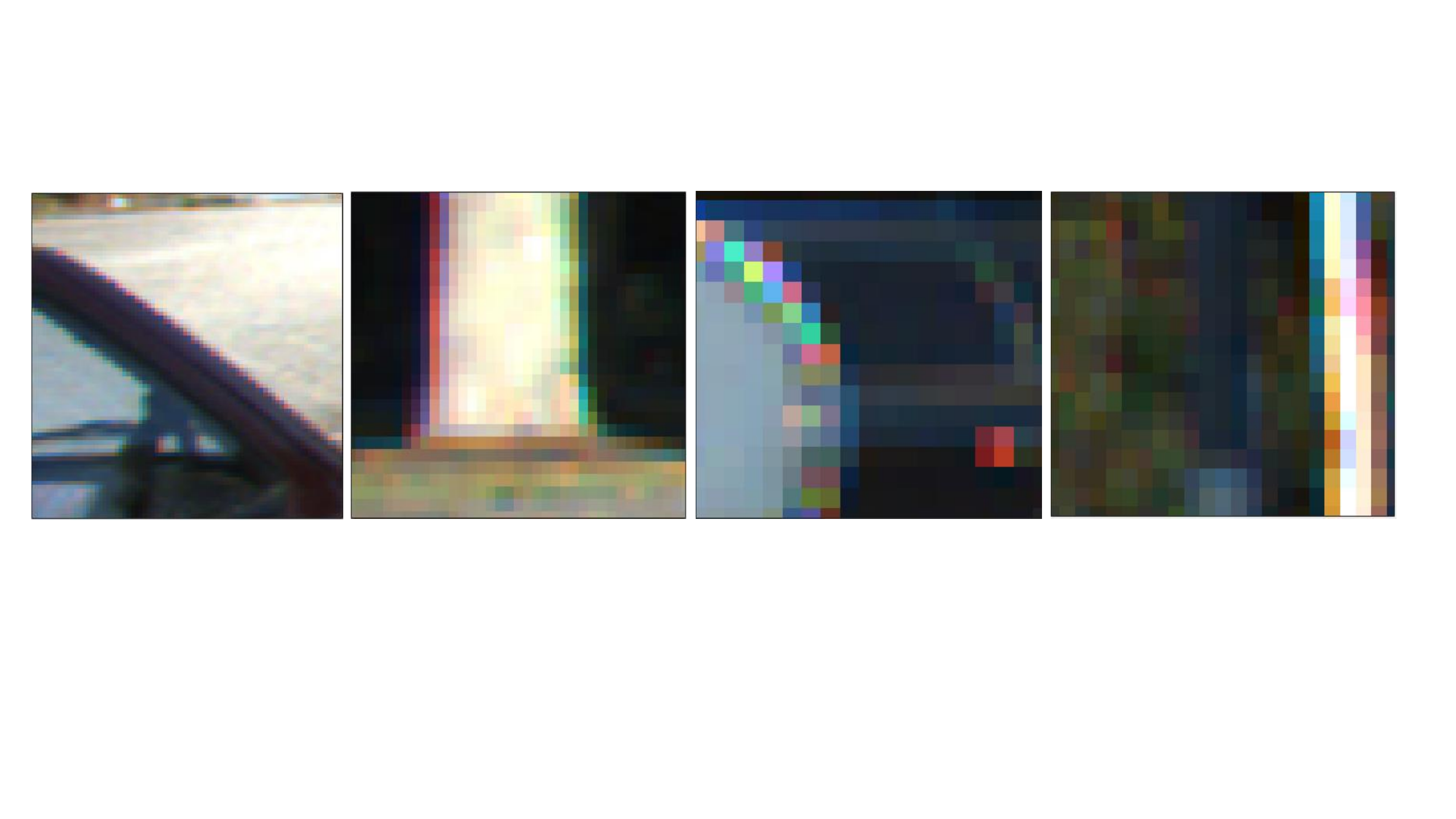}} & 
        \sidesubfloat{\hspace{0mm}\includegraphics[trim=20 120 20 120,clip, width=0.248\linewidth]{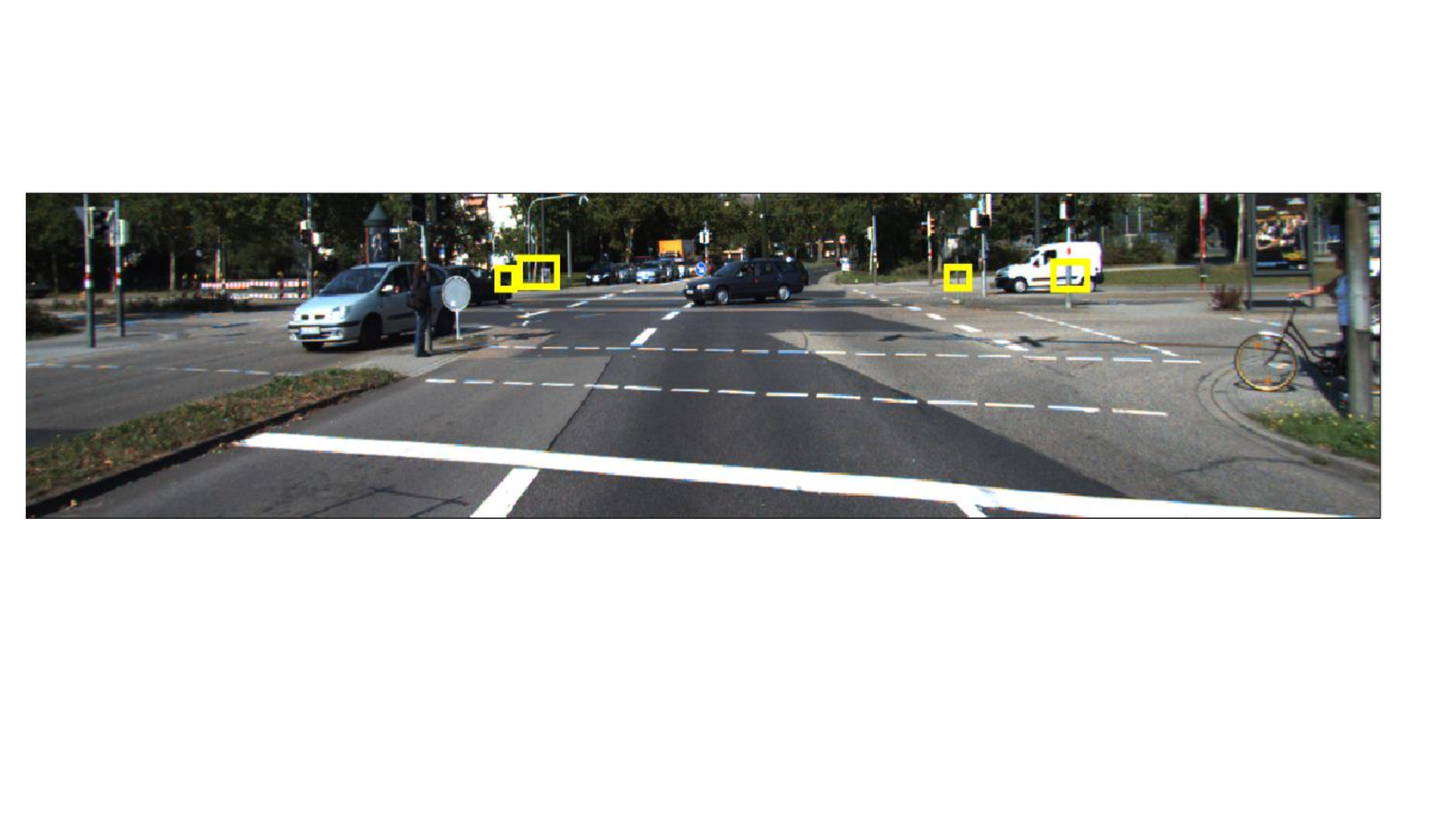}}  & \sidesubfloat{\hspace{0mm}\includegraphics[trim=20 120 20 120,clip, width=0.248\linewidth]{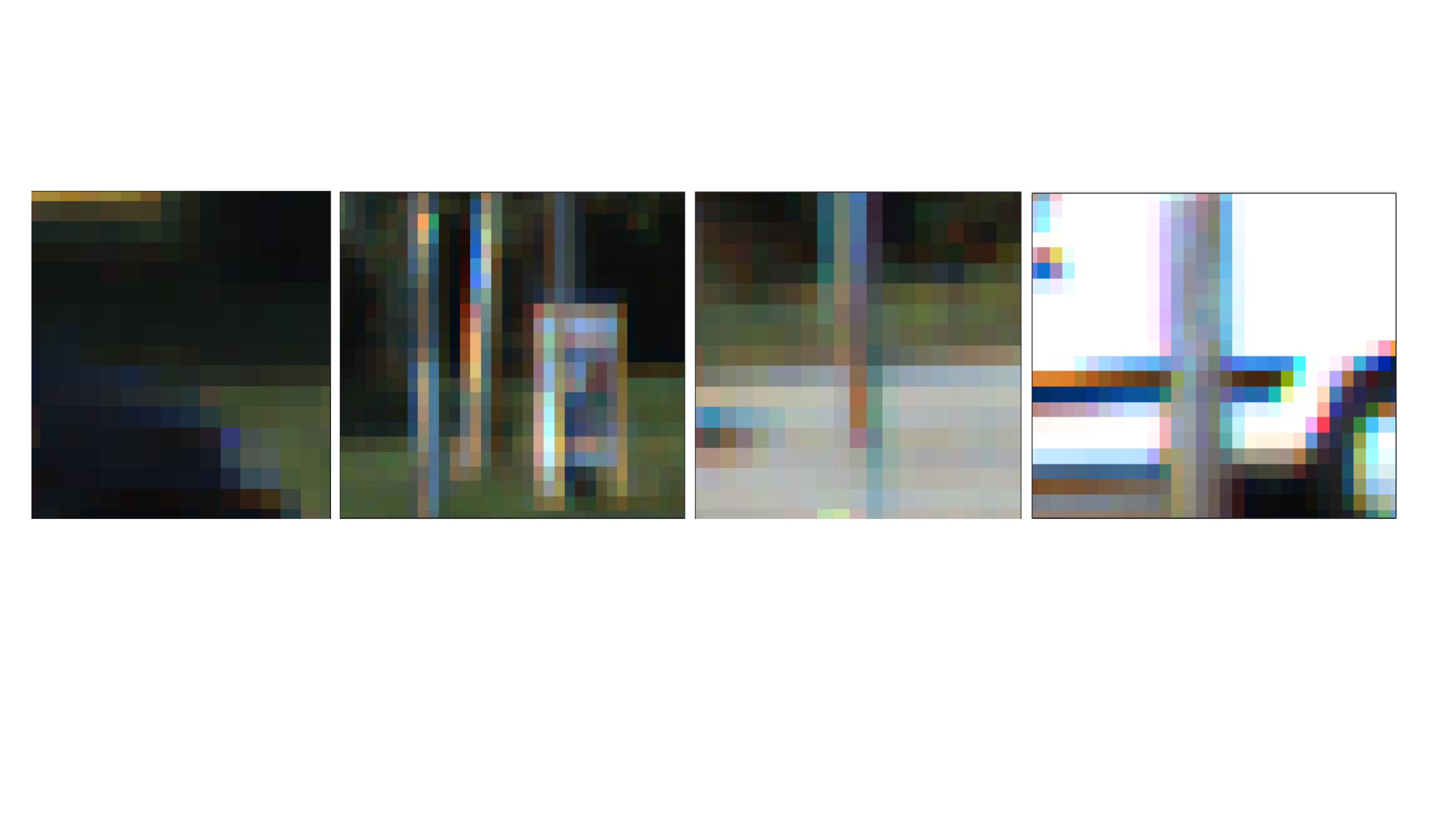}} \vspace{-3mm} \\ 
        \addtocounter{subfigure}{-3}
         \sidesubfloat[]{\includegraphics[trim=20 120 20 120,clip, width=0.248\linewidth]{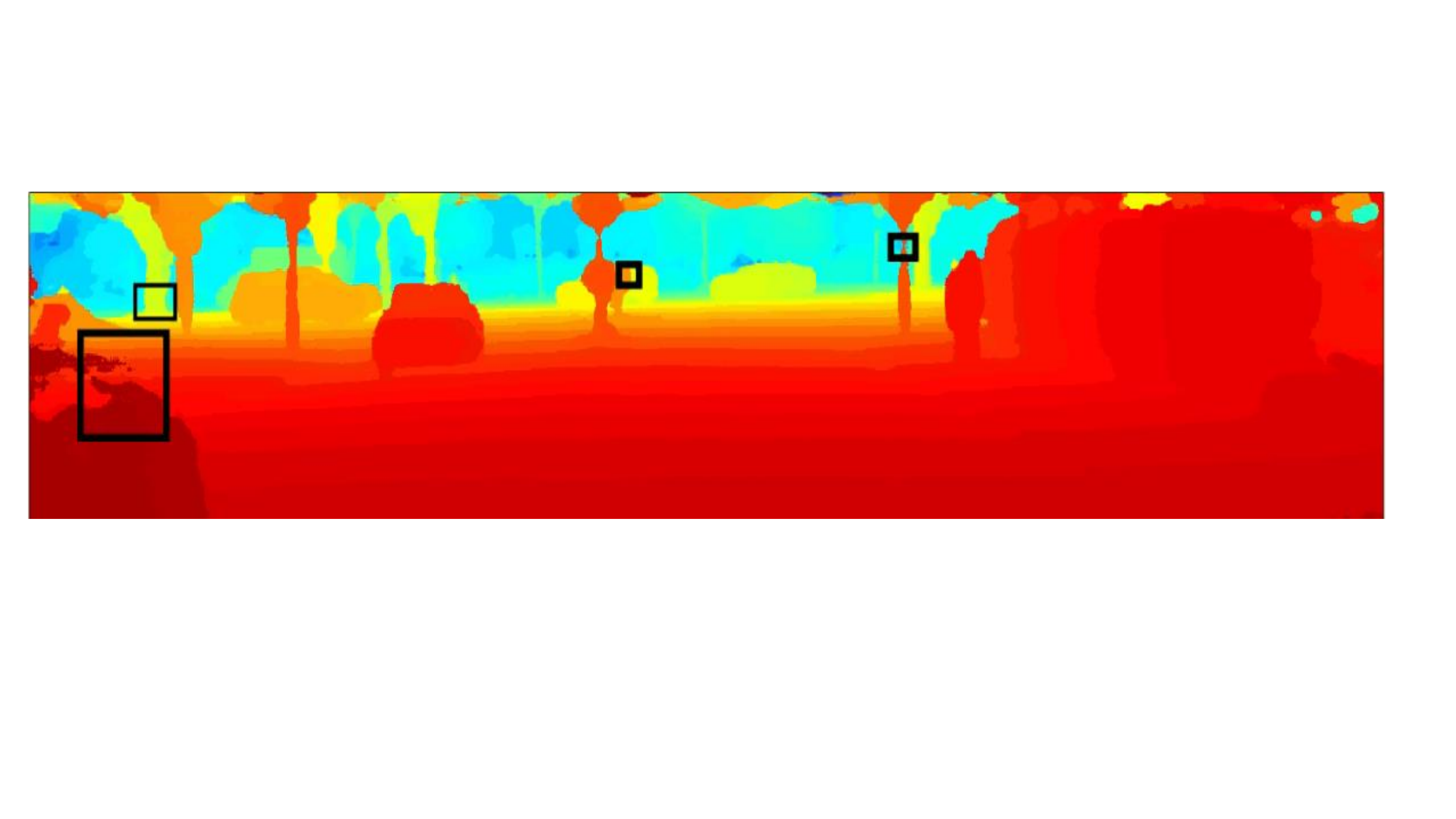}} & \sidesubfloat{\hspace{0mm}\includegraphics[trim=20 120 20 120,clip, width=0.248\linewidth]{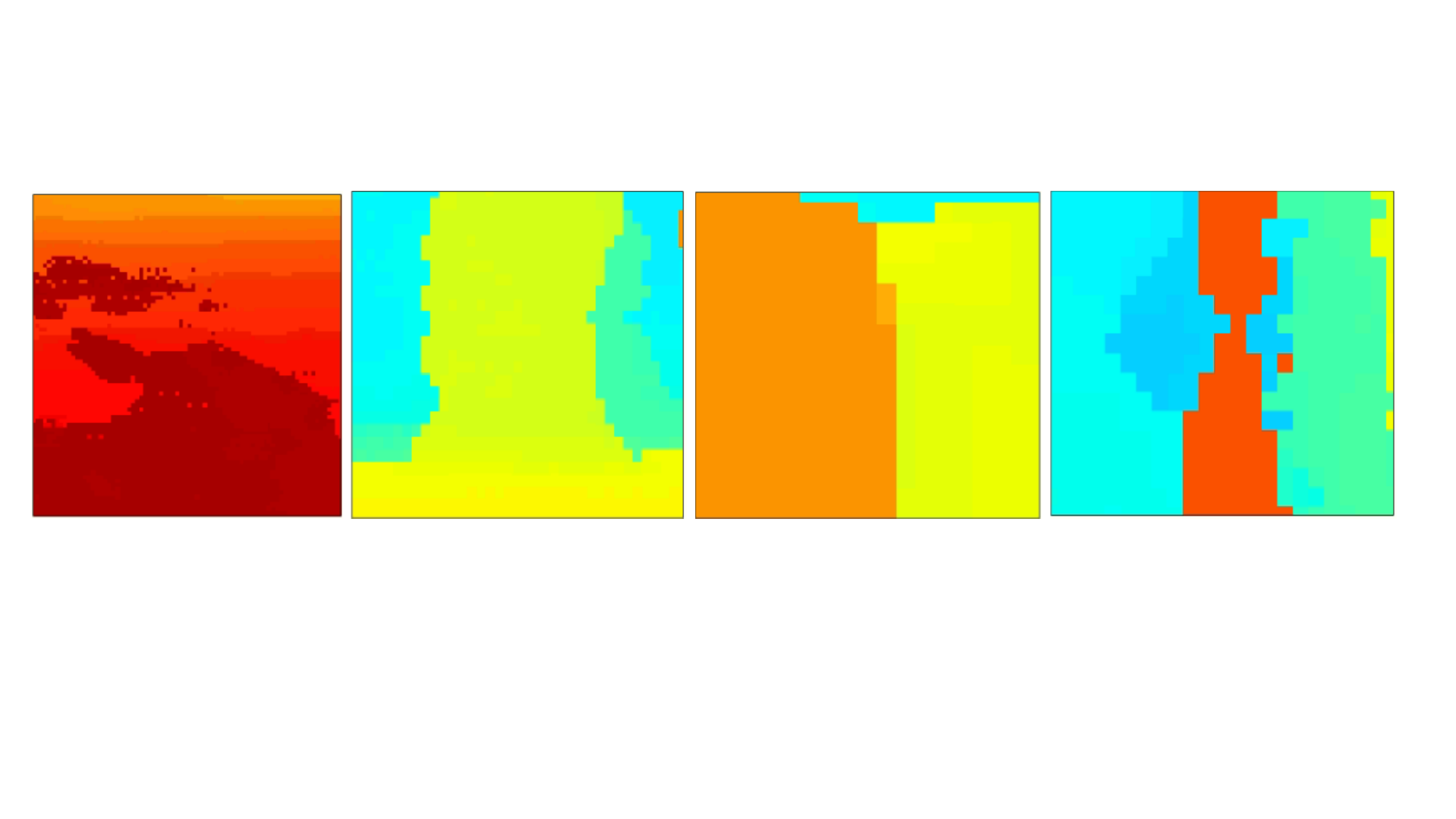}} & 
        \sidesubfloat{\hspace{0mm}\includegraphics[trim=20 120 20 120,clip, width=0.248\linewidth]{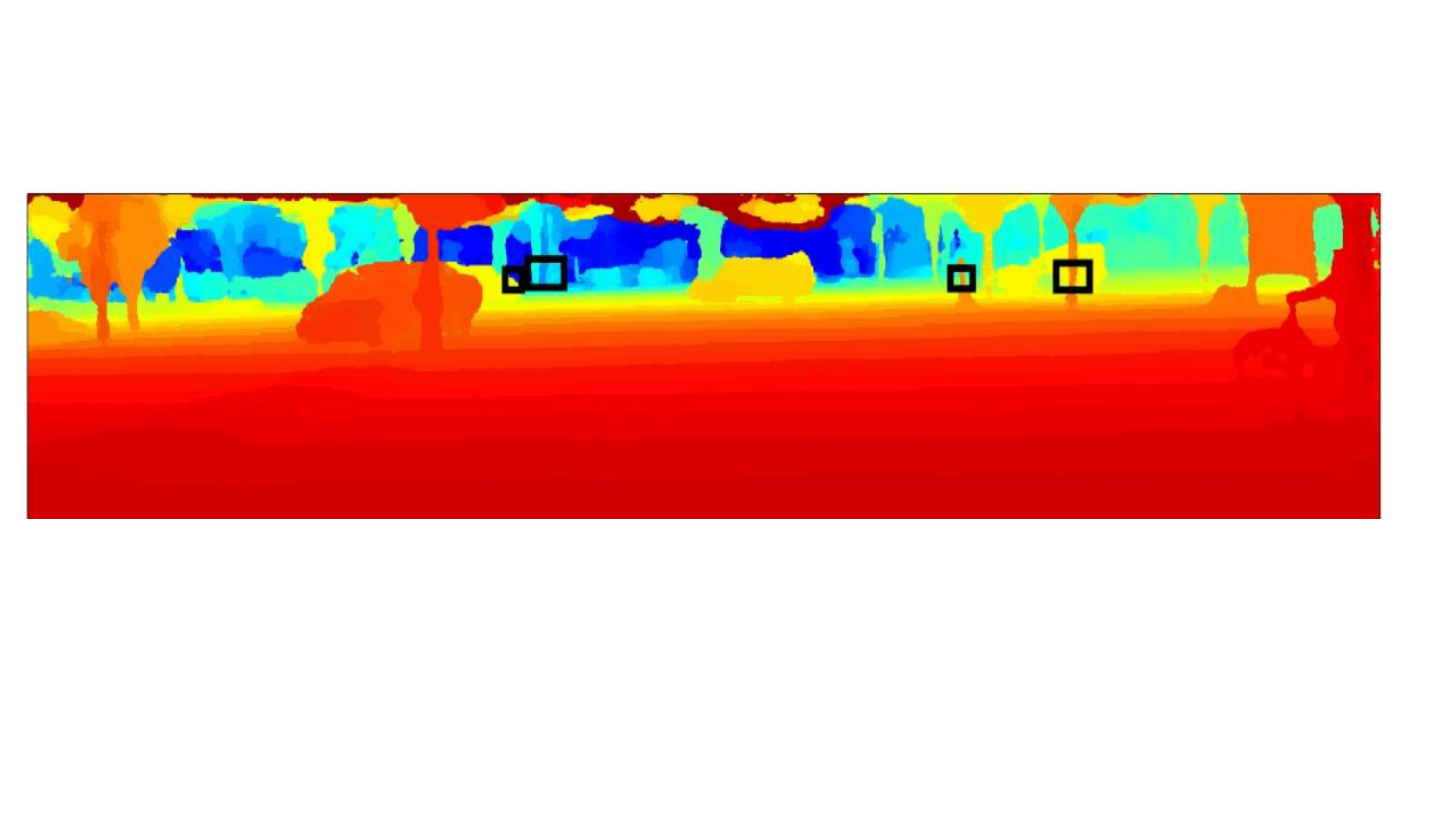}} & \sidesubfloat{\hspace{0mm}\includegraphics[trim=20 120 20 120,clip, width=0.248\linewidth]{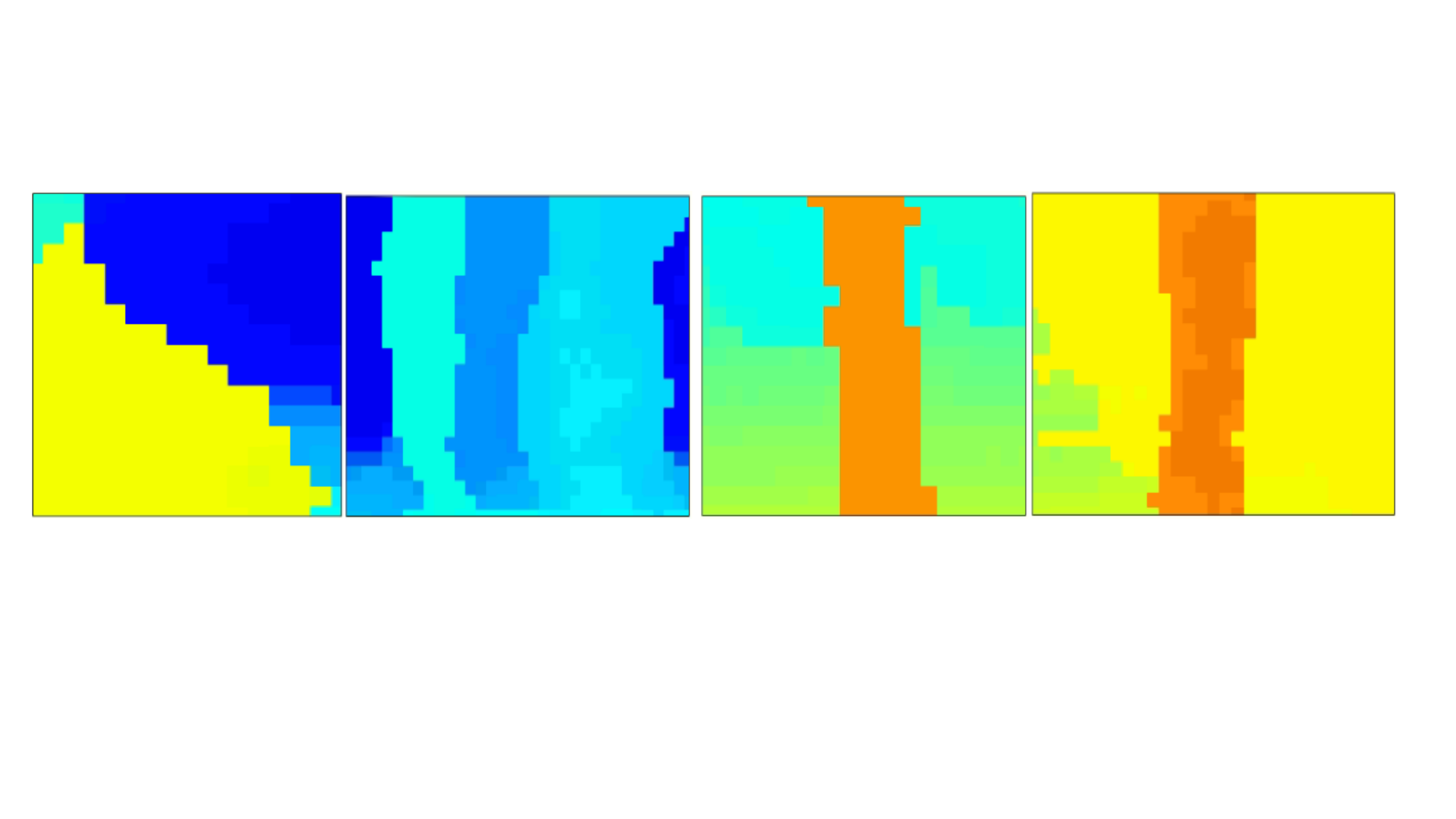}} \vspace{-3mm}\\ 
        \addtocounter{subfigure}{-3}
        \sidesubfloat[]{\includegraphics[trim=20 120 20 120,clip, width=0.248\linewidth]{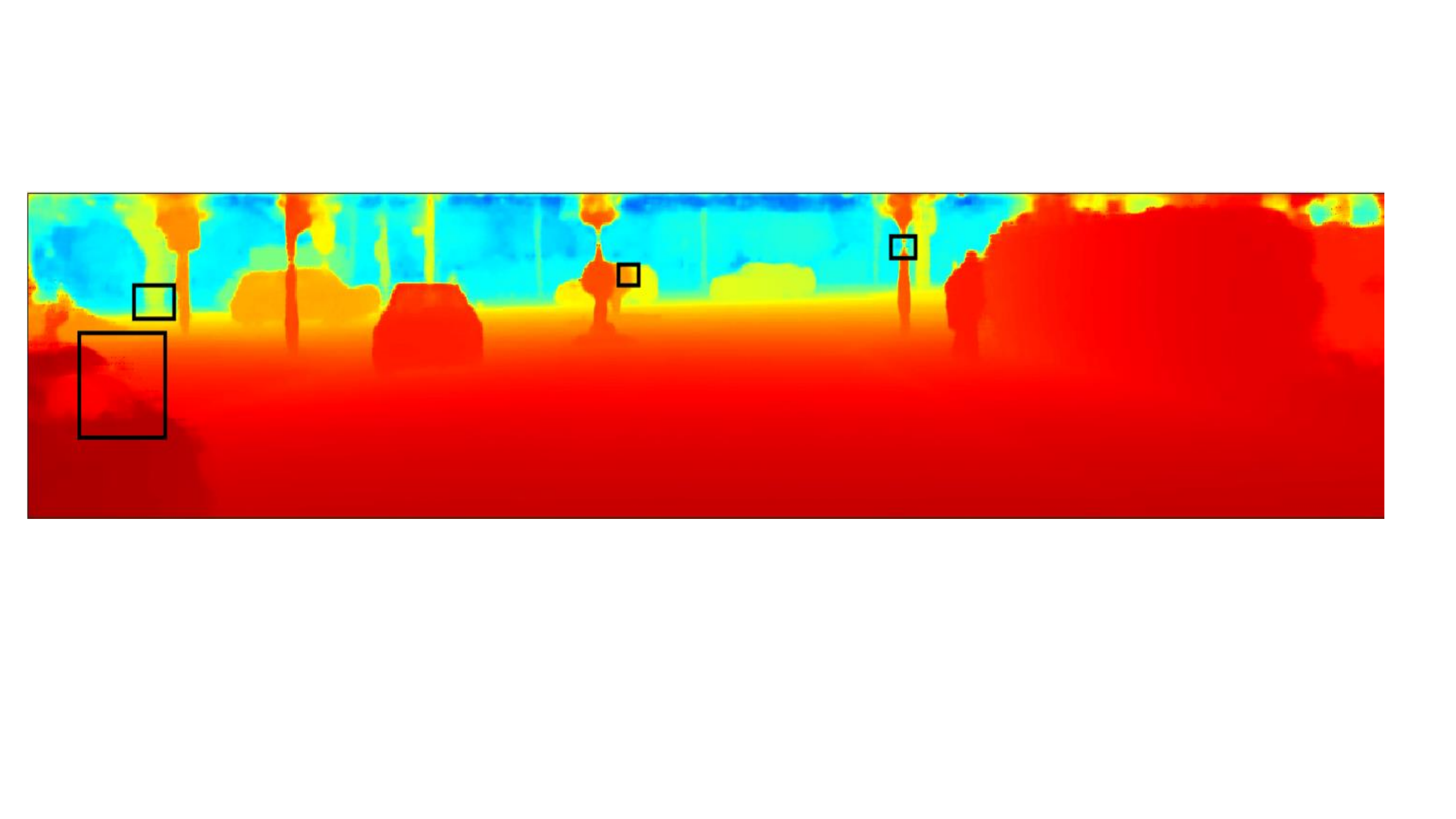}} & \sidesubfloat{\hspace{0mm}\includegraphics[trim=20 120 20 120,clip, width=0.248\linewidth]{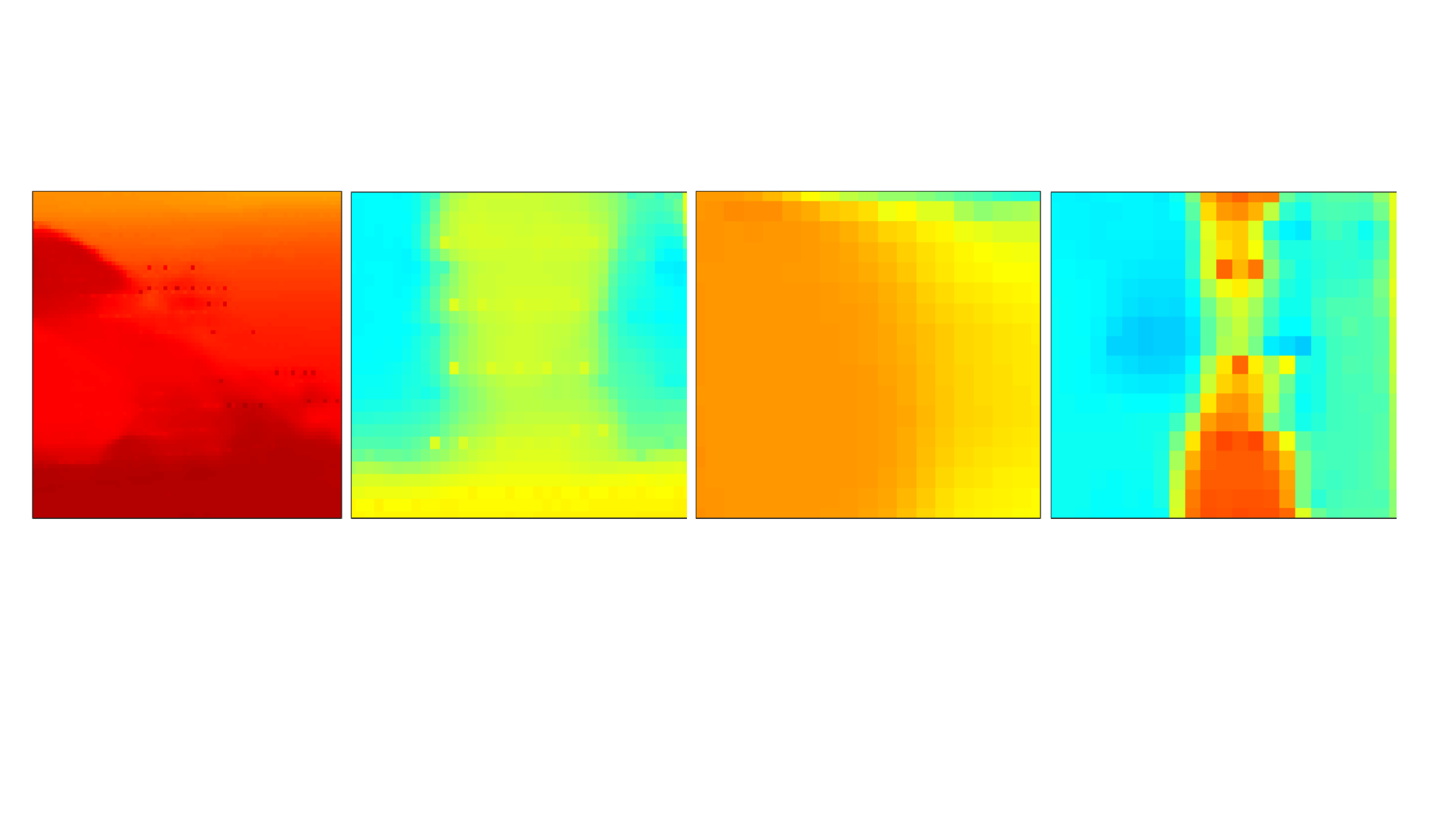}} & 
        \sidesubfloat{\hspace{0mm}\includegraphics[trim=20 120 20 120,clip, width=0.248\linewidth]{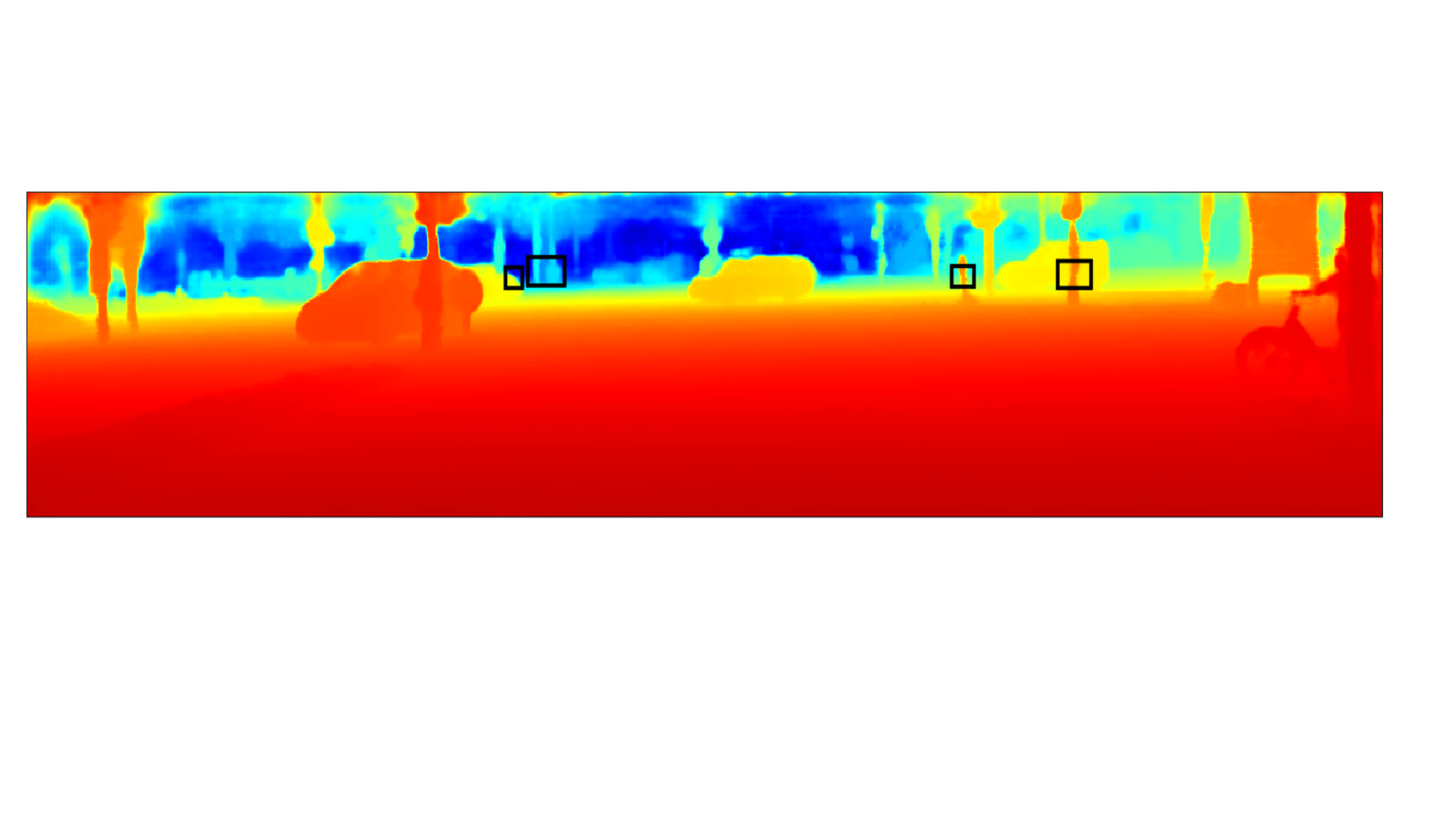}} & \sidesubfloat{\hspace{0mm}\includegraphics[trim=20 120 20 120,clip, width=0.248\linewidth]{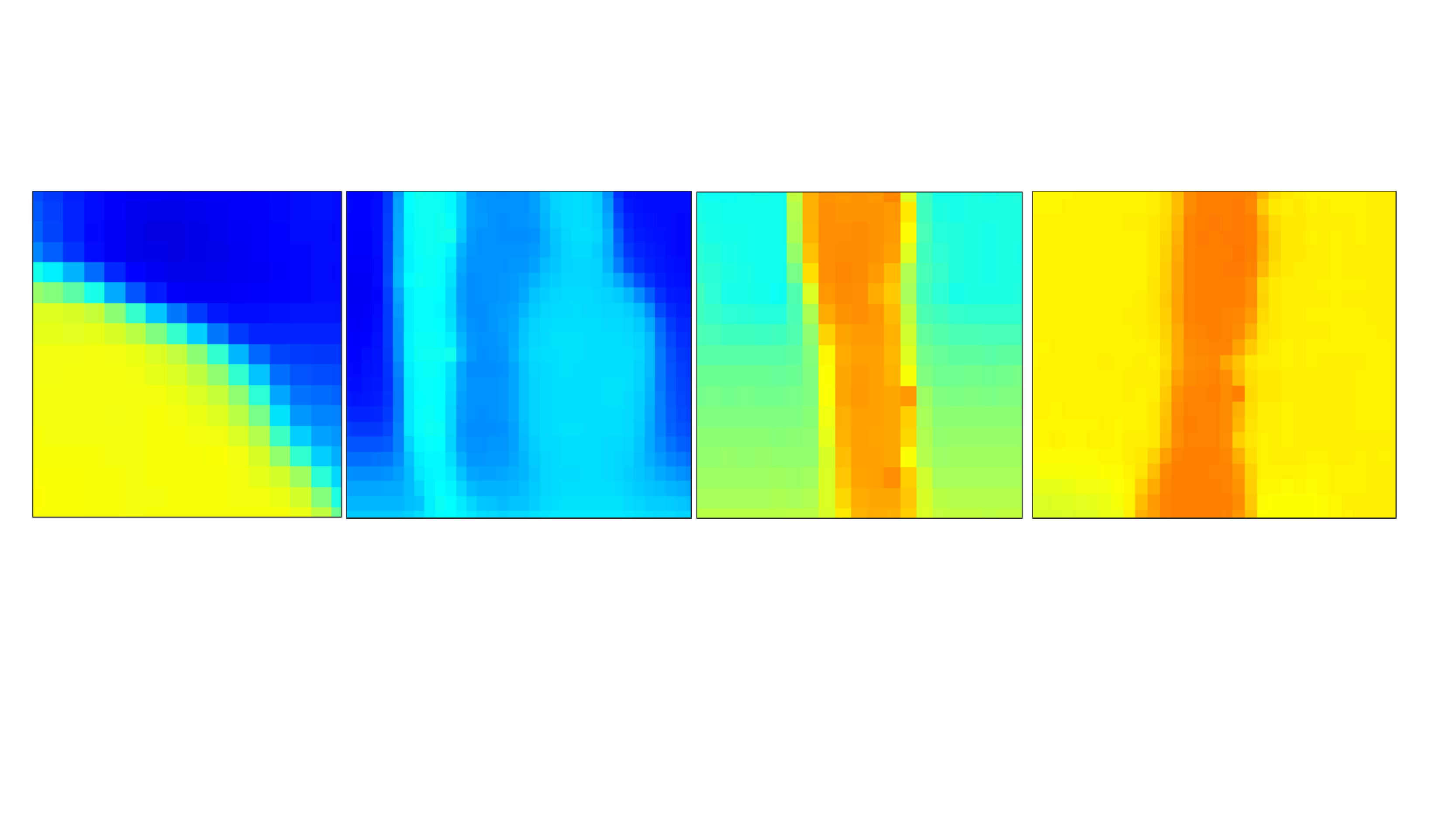}}\vspace{-3mm} \\
        \addtocounter{subfigure}{-3}
         \sidesubfloat[]{\includegraphics[trim=20 120 20 120,clip, width=0.248\linewidth]{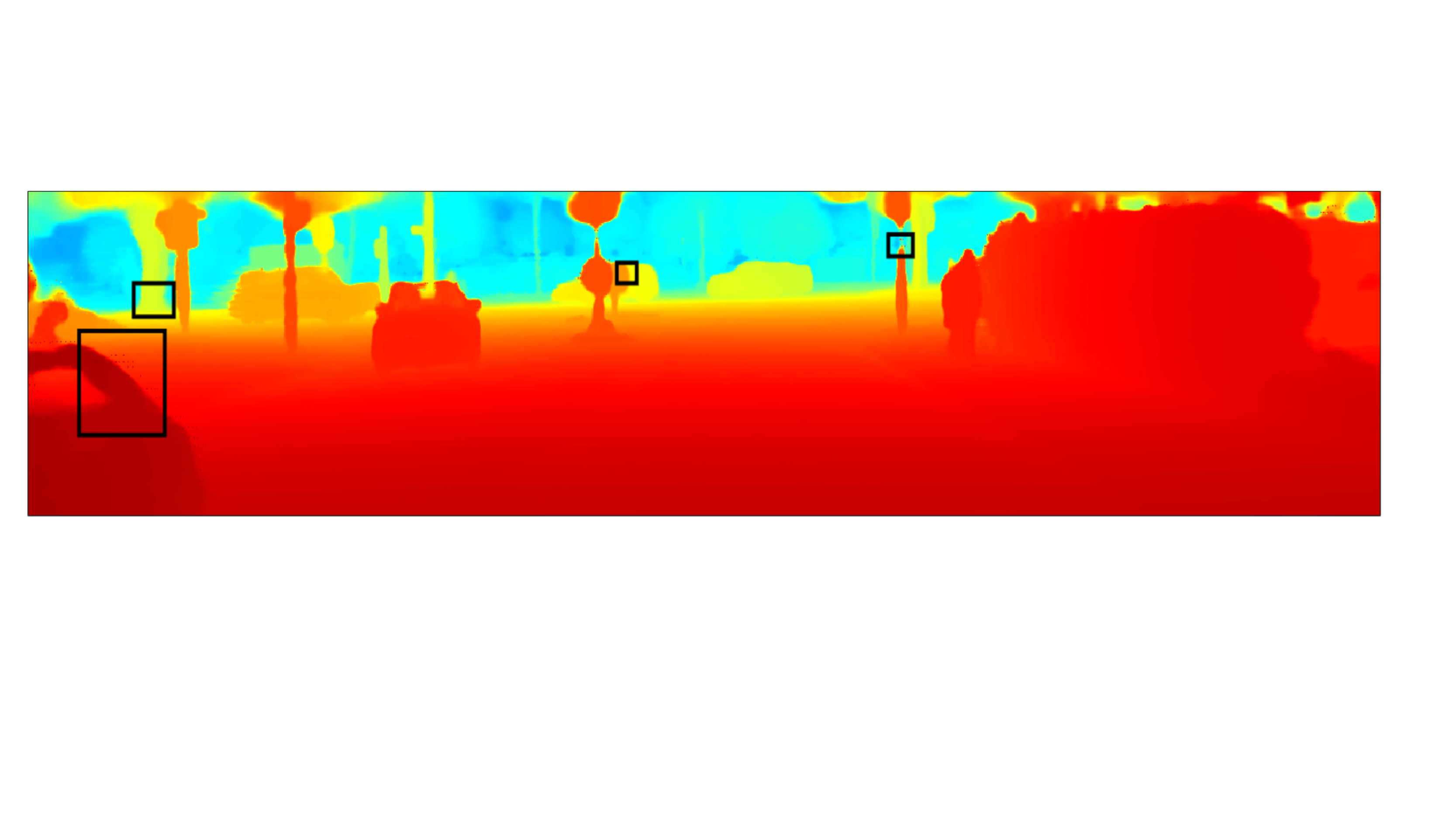}} & \sidesubfloat{\hspace{0mm}\includegraphics[trim=20 120 20 120,clip, width=0.248\linewidth]{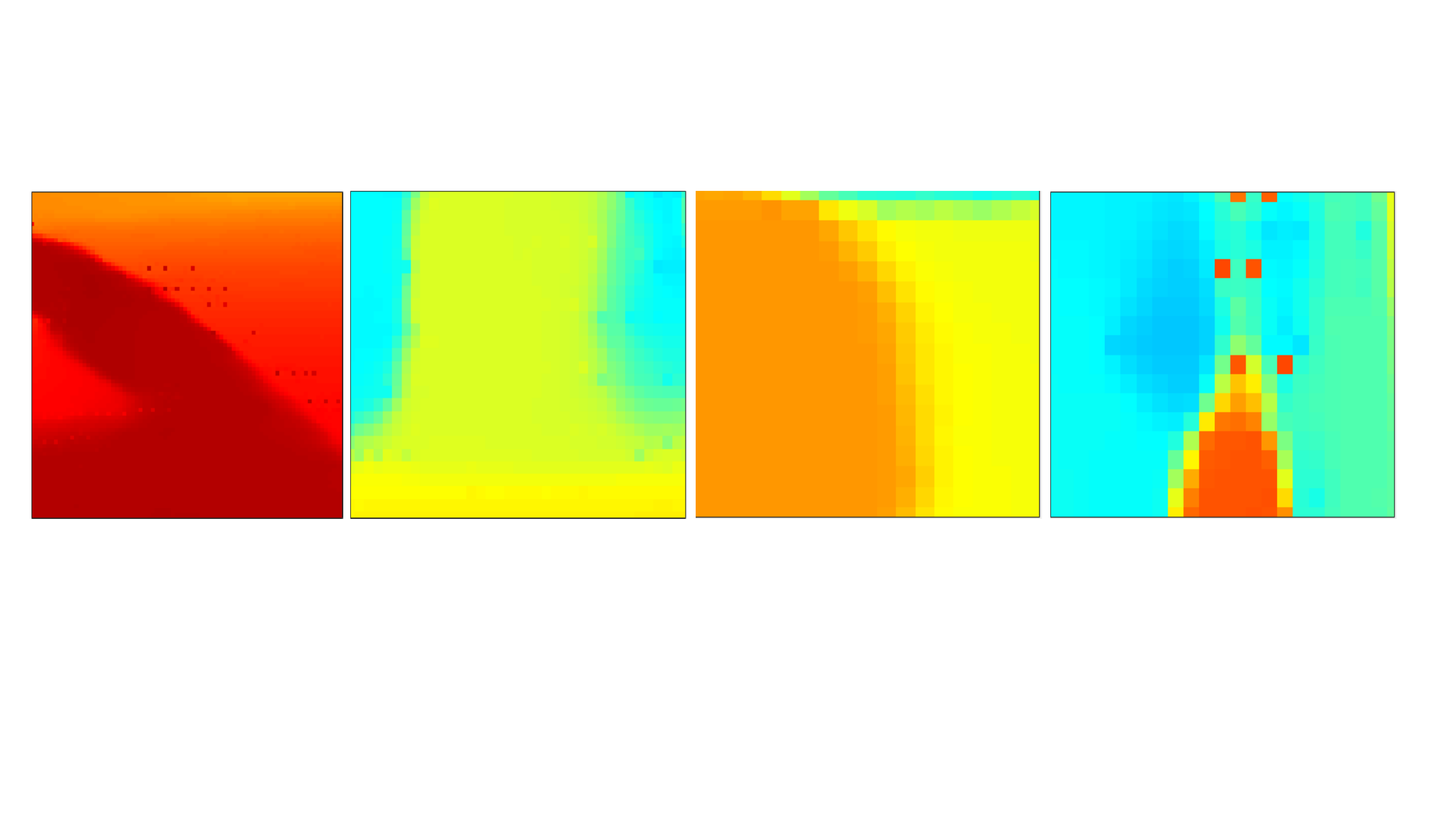}} & 
        \sidesubfloat{\hspace{0mm}\includegraphics[trim=20 120 20 120,clip, width=0.248\linewidth]{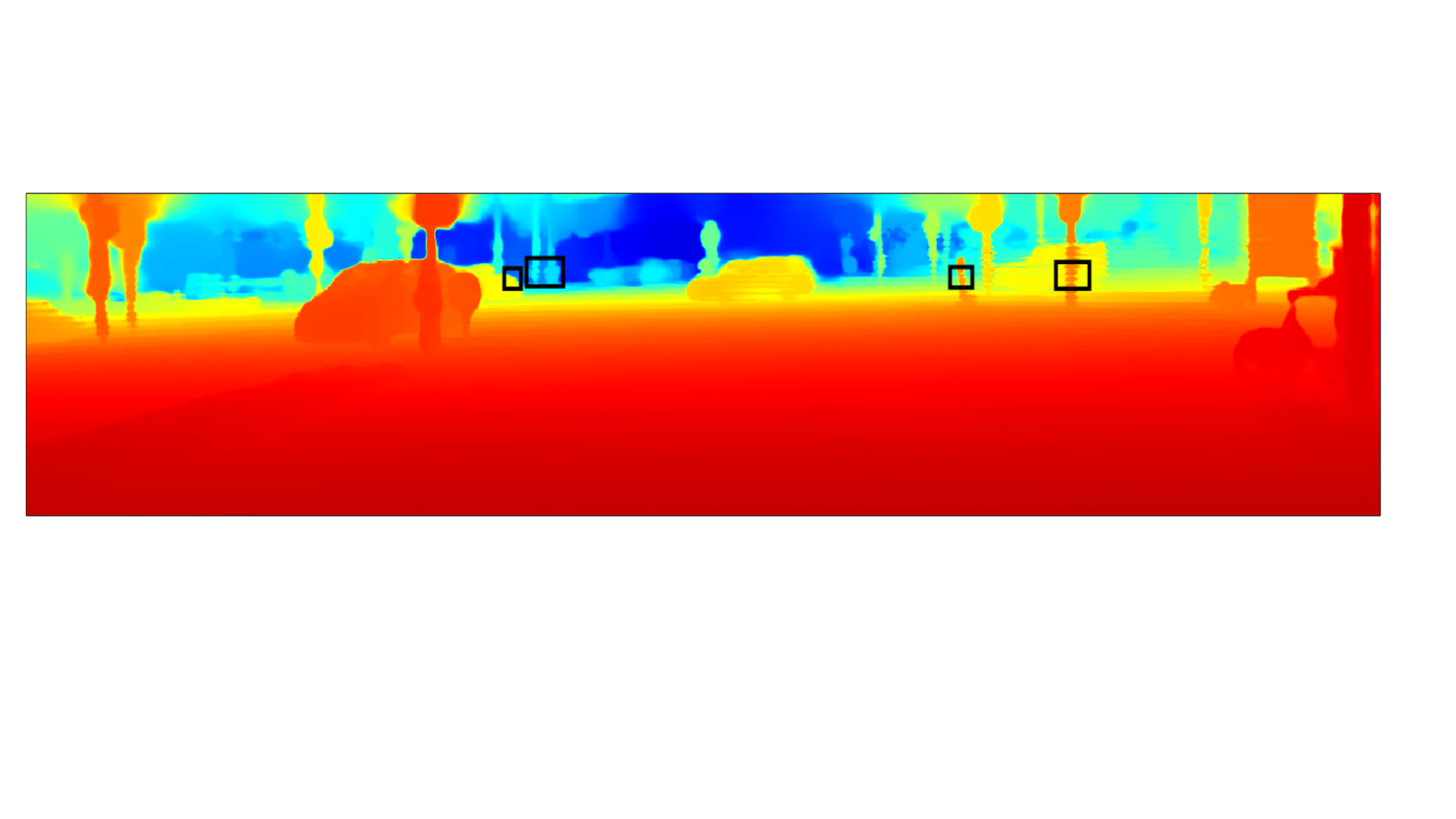}} & \sidesubfloat{\hspace{0mm}\includegraphics[trim=20 120 20 120,clip, width=0.248\linewidth]{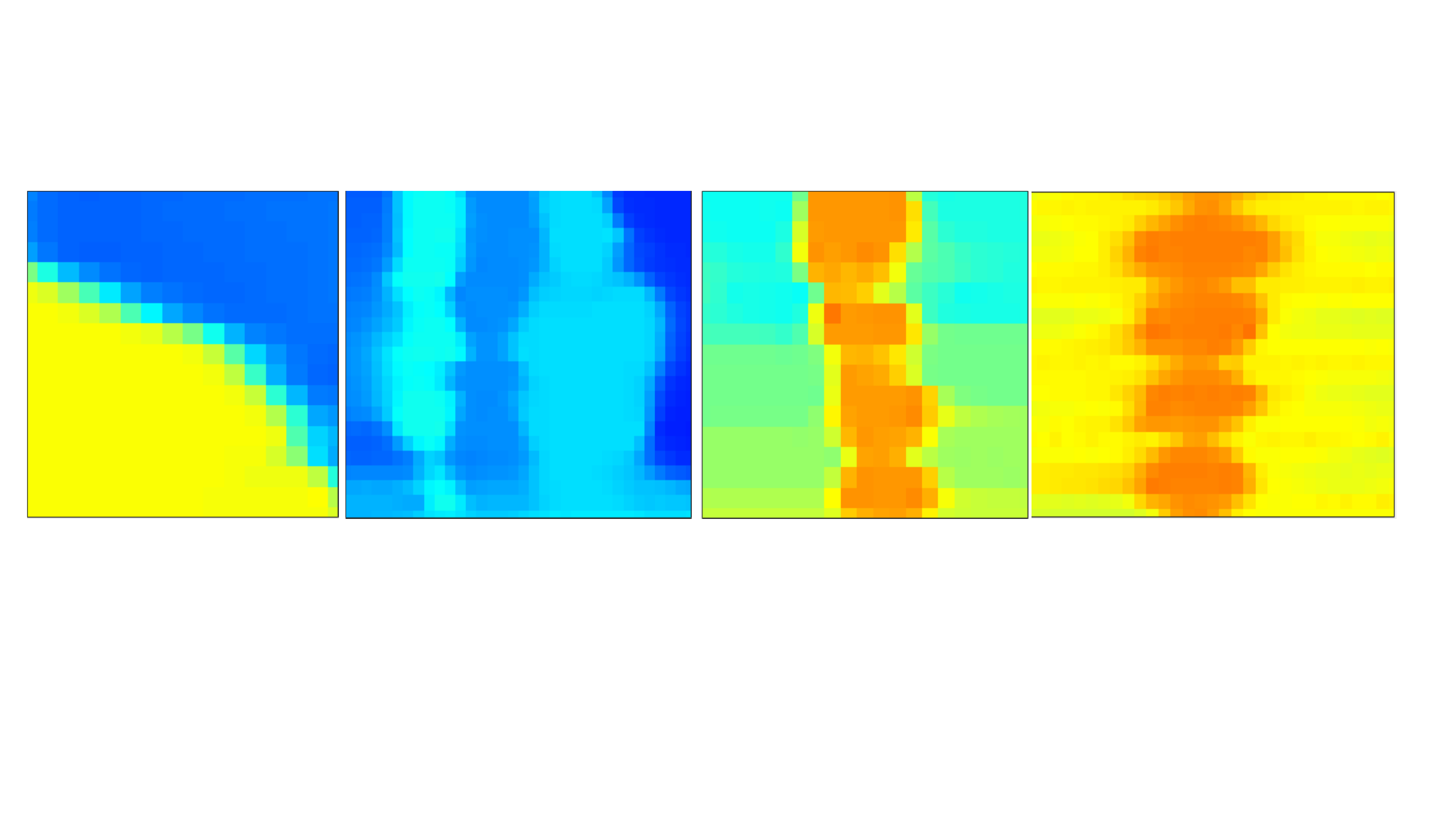}}\vspace{-3mm} \\
        \addtocounter{subfigure}{-3}
         \sidesubfloat[]{\hspace{0mm}\includegraphics[trim=10 120 0 120,clip, width=0.248\linewidth]{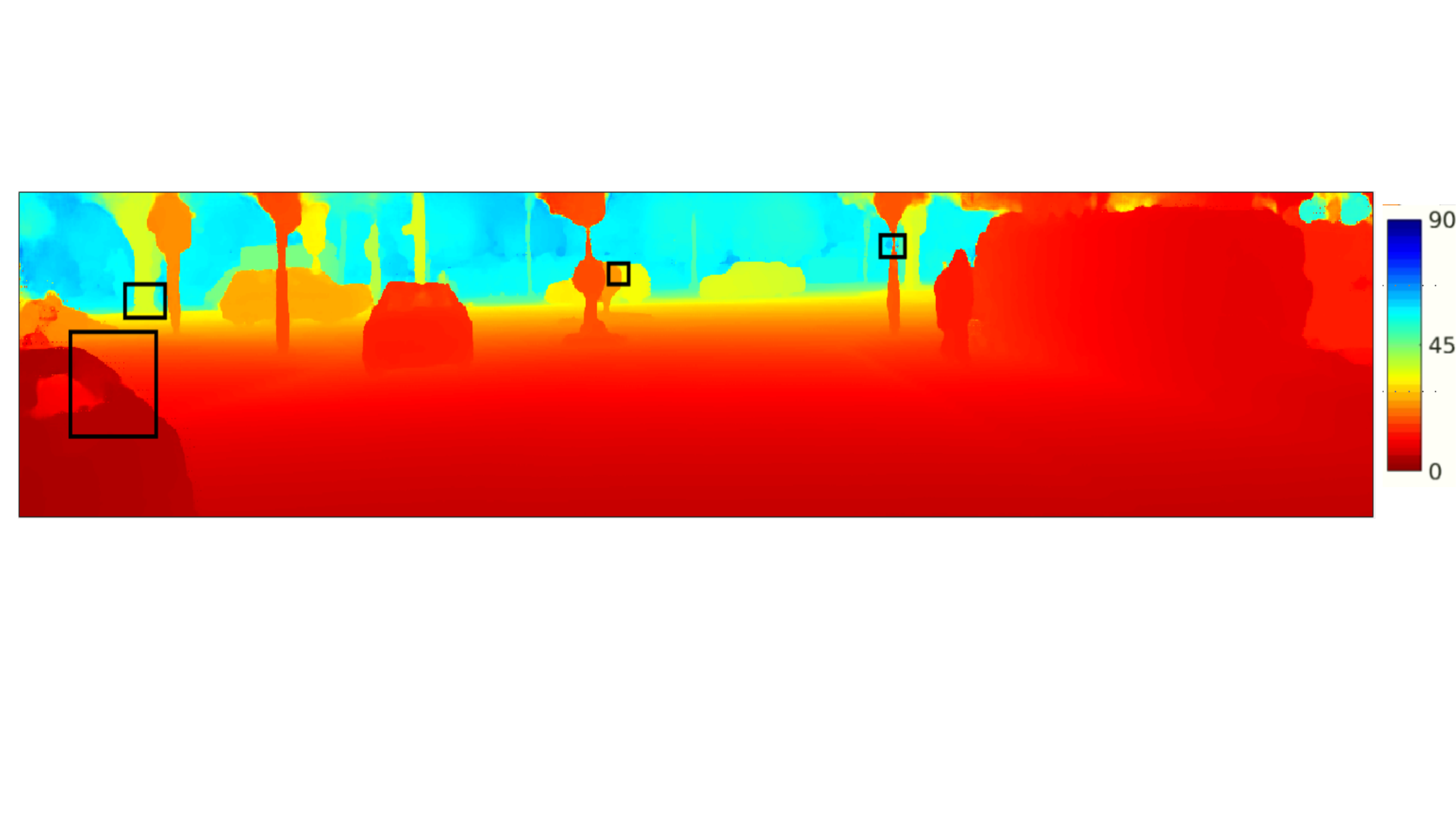}} & \sidesubfloat{\hspace{0mm}\includegraphics[trim=20 120 20 120,clip, width=0.248\linewidth]{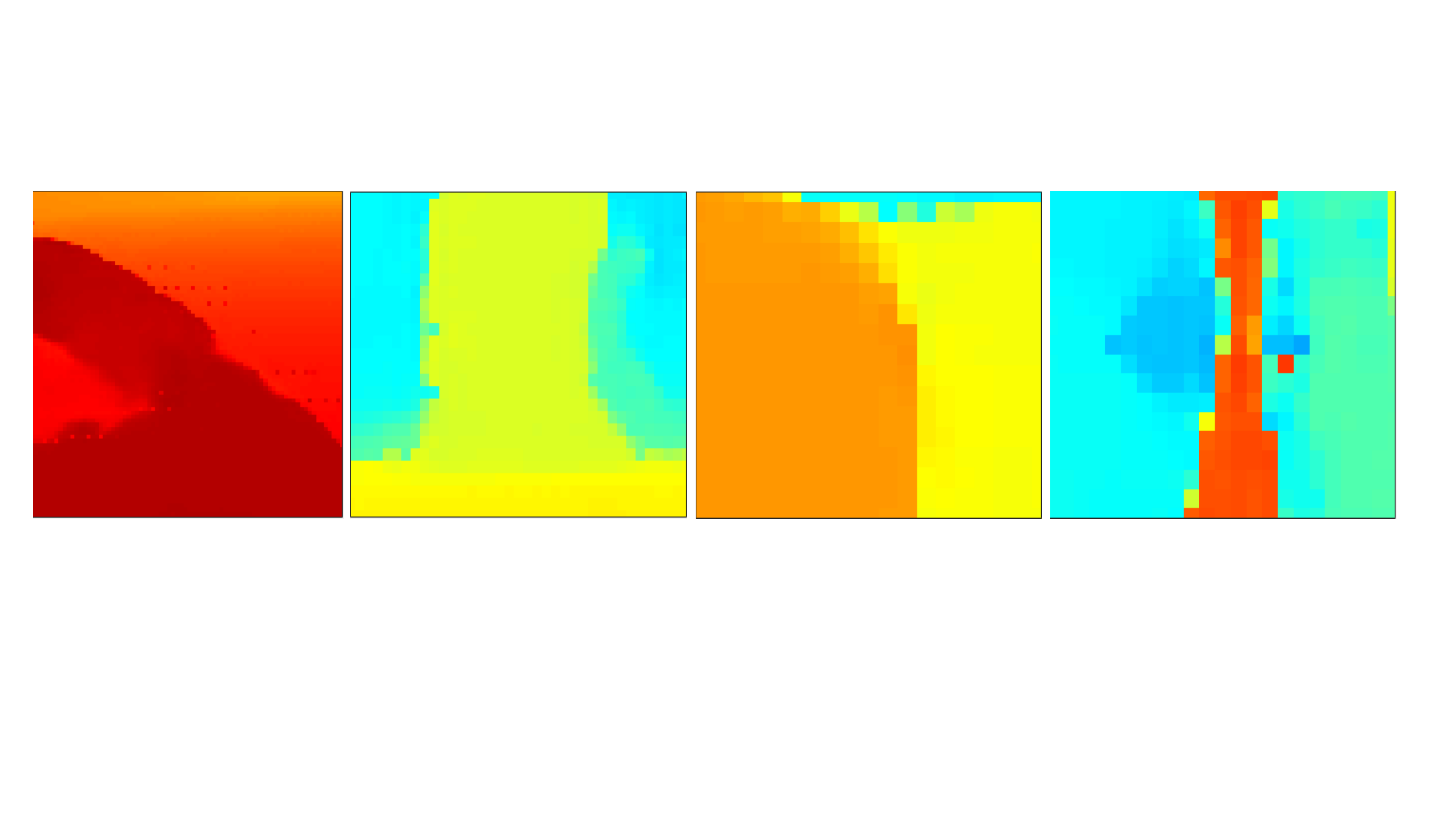}} & 
        \sidesubfloat{\hspace{0mm}\includegraphics[trim=20 120 20 120,clip, width=0.248\linewidth]{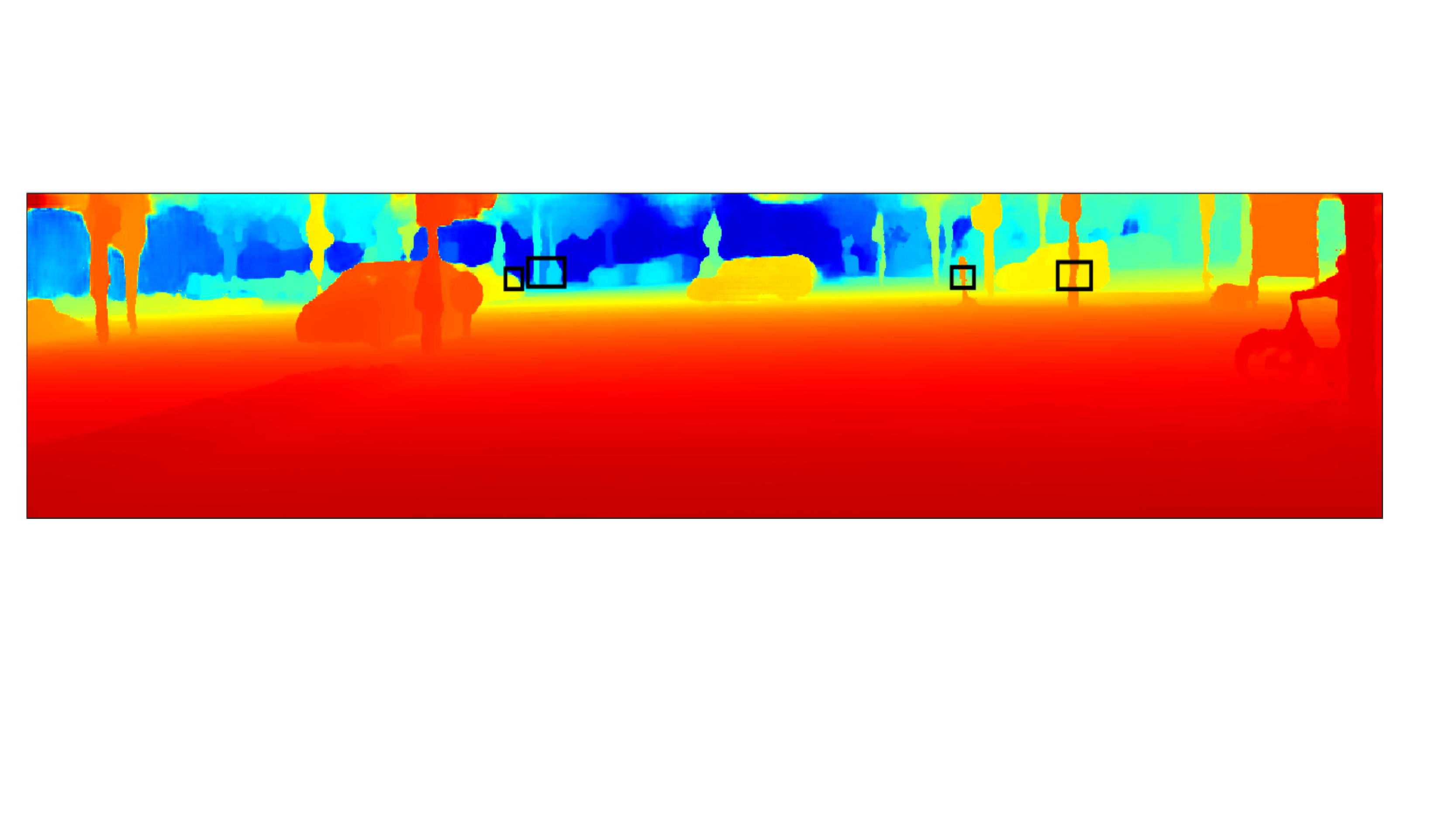}} & \sidesubfloat{\hspace{0mm}\includegraphics[trim=20 120 20 120,clip, width=0.248\linewidth]{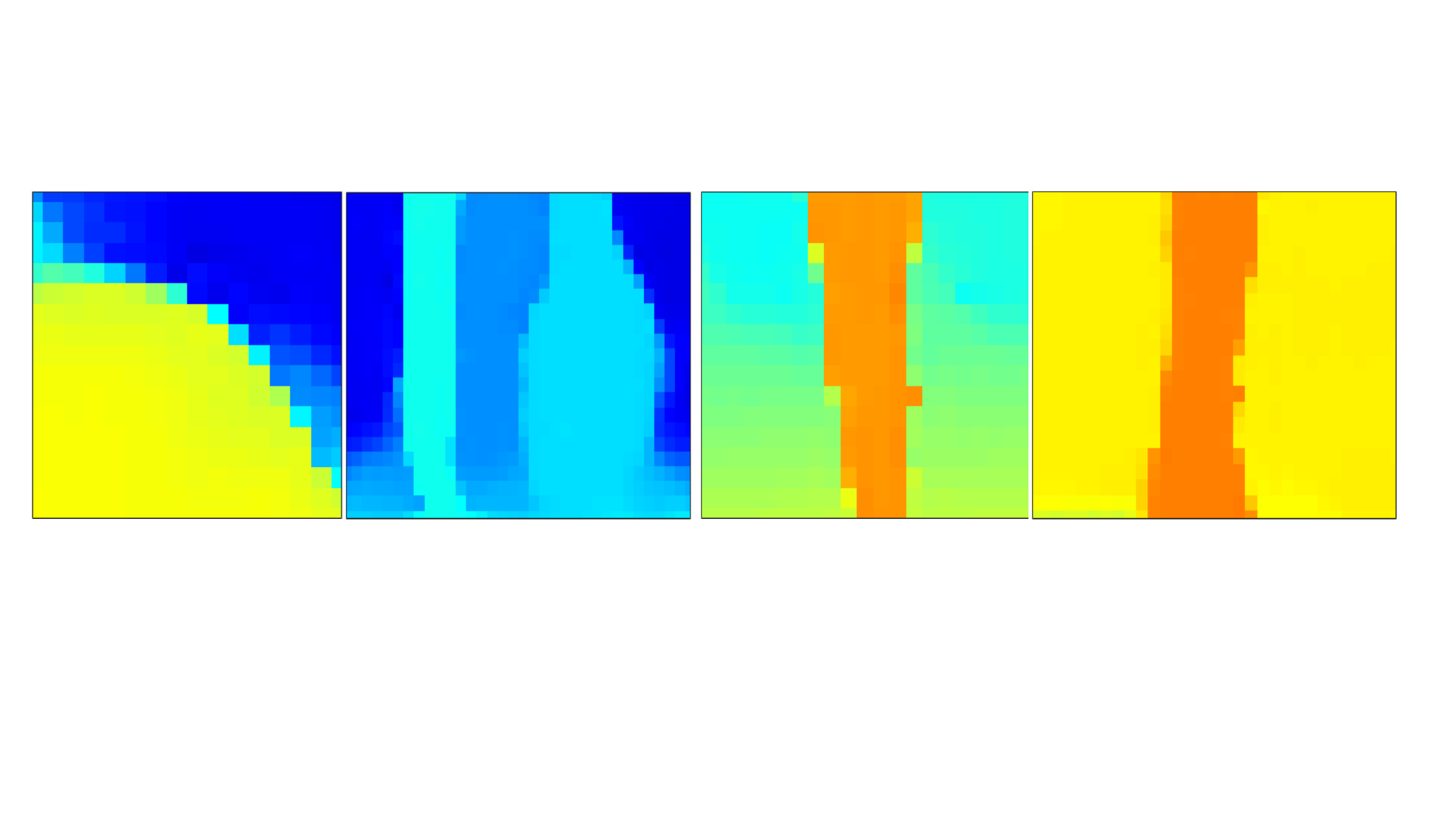}}\vspace{-5mm} \\
    \end{tabular}}
    \caption{\small Comparison of our method with SoTA methods with whole and zoom in views (a) showing Color Images (b) DC~\cite{imran2019depth}, (c) MultiStack~\cite{Li_2020_WACV} (d) NLSPN~\cite{park2020non} and our method (e). Four different regions of the image from two different instants are selected to show depth quality from diverse areas.}
    \label{fig:qual_results}
\end{figure*}

\section{Experimental Results}
\Paragraph{Dataset}
We evaluate the proposed algorithm on the standard KITTI Depth Completion dataset~\cite{Geiger2012CVPR}, a real-world outdoor scene, NYU2, with indoor scenes \cite{Silberman:ECCV12}, and Virtual KITTI \cite{cabon2020virtual}, a synthetic dataset with photo-realistic images and dense ground-truth depth.
KITTI depth is created by aggregating LiDAR scans from $11$ consecutive frames into one, producing a semi-dense ground truth (GT) with $30$\% annotated depth pixels.  
The sparsity of GT makes depth estimation more challenging.
Note that we do not require any synthetic depth data for pre-training as used by~\cite{yang2019dense,qiu2019deeplidar} to improve performance. 
The dataset consists of $85$K, $1$K, and $1$K samples for training, validation, and testing respectively. 
Although the training set has different image sizes, the test and validation sets are cropped to a uniform size of $352\times1,216$.  

Although created in a real world scenario, the semi-dense GT produced by Uhrig {\it et al.}~\cite{uhrig2017sparsity} has far fewer depth points on object boundaries (see Fig.~\ref{fig:dep_smearing} (a)), and is susceptible to outliers. 
As we claim our method works well on boundaries, we also evaluate on VKITTI 2.0, a synthetic dataset with clean and dense GT depth at depth discontinuities. The VKITTI $2.0$, created by the Unity game engine, contains $5$ different camera locations ($15^o$ left, $15^o$ right, $30^o$ left, $30^o$ right, clone) in addition to $5$ different driving sequences. Additionally, there are stereo image pairs for each camera location. For training and testing, we only use the clone (forward facing camera) with stereo image pairs. For VKITTI training, $2$k training images were created from driving sequences $01$, $02$, $06$, and $018$ respectively. For testing, we use sequence $020$ at the left stereo camera, and choose every other frames, with total $420$ images.
We subsample the dense GT depth in azimuth-elevation space to simulate LiDAR-like pattern as sparse inputs. 
Further, we create the pseudo GT following~\cite{uhrig2017sparsity} to study the effects of outlier noise on training and evaluation. More details are shared in the supplementary. 

To show the generalizibility of our method, we also evaluate on NYU-Depth v2 dataset~\cite{Silberman:ECCV12}, which consists of RGB and depth images obtained from Kinect in $464$  scenes. We use the official split of data, where $249$ scenes are used for training and we sample $50$K images out of the training similar to~\cite{qiu2019deeplidar, park2020non}.   For  testing,  the  standard  labelled  set  of $654$ images is used.  The original image size is first downsampled to half, and then center-cropped, producing a network input dimension of $304\times208$. Unlike~\cite{park2020non}, we use the {\it same} loss function for all the datasets.

\Paragraph{Metrics}
The standard metrics used by KITTI include RMSE, MAE, iMAE and iRMSE. 
Since RMSE is used as the preferred metric for depth completion, most SoTA methods on the KITTI leaderboard use MSE as their primary loss. 
We also include tMAE and tRMSE metrics proposed in~\cite{imran2019depth} since it can discount outlier depth pixels ({\it i.e.}, floating depth pixels around boundary regions) and give a better evaluation of depth pixels at and within object boundaries.

\subsection{Results}

\Paragraph{Quantitative Results} Tab.~\ref{tab:kitti_results} compares the performance on KITTI's test/validation sets, with a $64$-row LiDAR and color image as input. We list the SoTA methods with performance quoted from their papers. The inference times are calculated on a single GPU of GTX $1080$~Ti.
The method~\cite{park2020non} with lowest RMSE achieves this at the expense of inference time.
We outperform the SoTA methods in other metrics including MAE, and iMAE. 
The exception is RMSE, by which the methods are ranked in the KITTI leaderboard.
That leads us to investigate in which areas are our method perform better and worse, which we examine next.

\Paragraph{Qualitative Results} Fig.~\ref{fig:qual_results} shows our depth estimation quality compared to baselines. 
We choose three best SoTA methods: MultiStack~\cite{Li_2020_WACV}, NLSPN~\cite{park2020non}, and DC~\cite{imran2019depth}. 
Different local regions including poles, trees, cars, and traffic signs, illustrate the depth quality of close- and long-range depth pixels. 
The zoomed-in view shows the substantial improvement of our depth map over SoTA, especially along sharp object boundaries. 
\cite{Li_2020_WACV} has a more blurred estimation around boundaries leading to mixed depth pixels and holes within objects, such as on the traffic poles and van. 
Although~\cite{park2020non} has reduced mixed depths and more tighter boundary, depth mixing still exists (blurriness at object boundaries), additionally it suffers from jagged boundary edges and streaking artifacts. 
\begin{figure}[t!]
\vspace{-3mm}
\captionsetup{font=small}
\begin{center}
          \sidesubfloat[]{\includegraphics[trim=45 270 45 90,clip,width=0.96\linewidth]{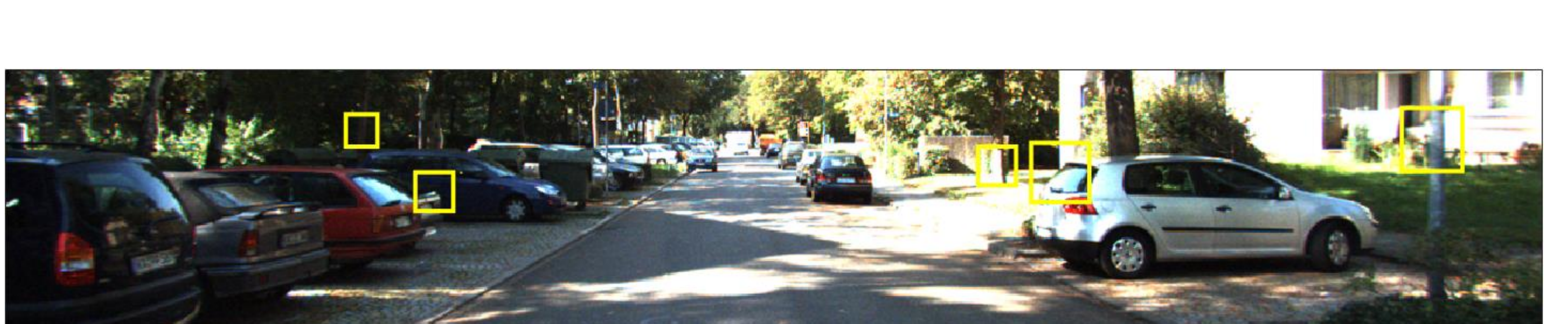}} \vspace{-3mm}\\
         \sidesubfloat[]{\hspace{0.4mm}\includegraphics[trim=38 250 40 112,clip,width=0.96\linewidth]{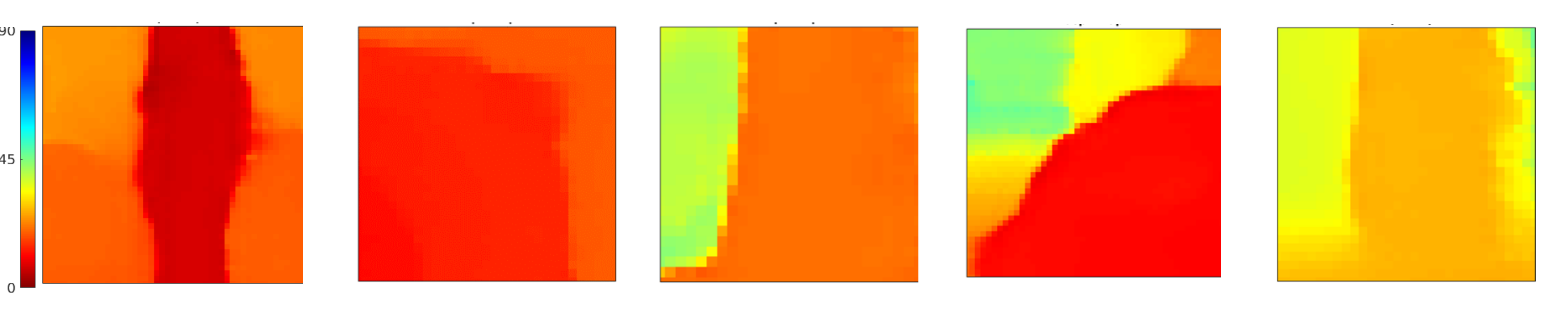}} \vspace{-3mm}\\
         \sidesubfloat[]{\includegraphics[trim=32 250 40 120,clip,width=0.96\linewidth]{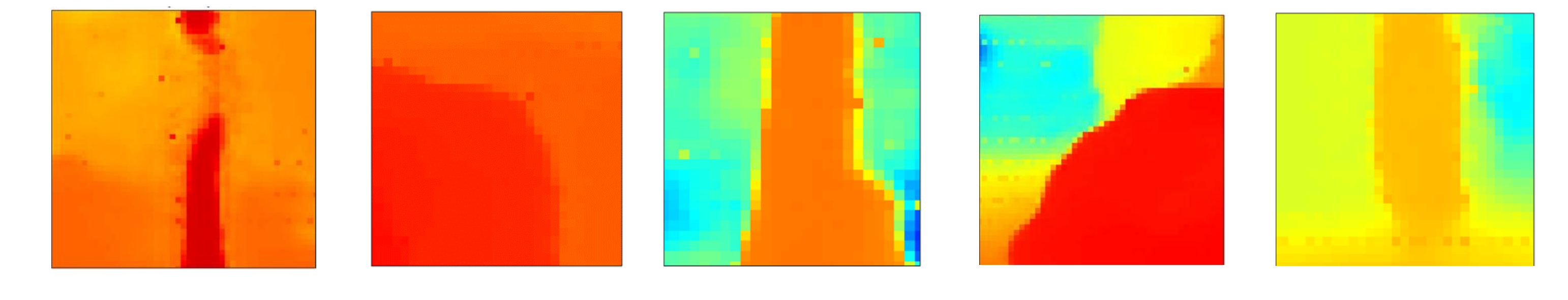}} \vspace{-3mm}\\
          \sidesubfloat[]{\hspace{-1mm} \includegraphics[trim=32 250 40 120,clip,width=0.96\linewidth]{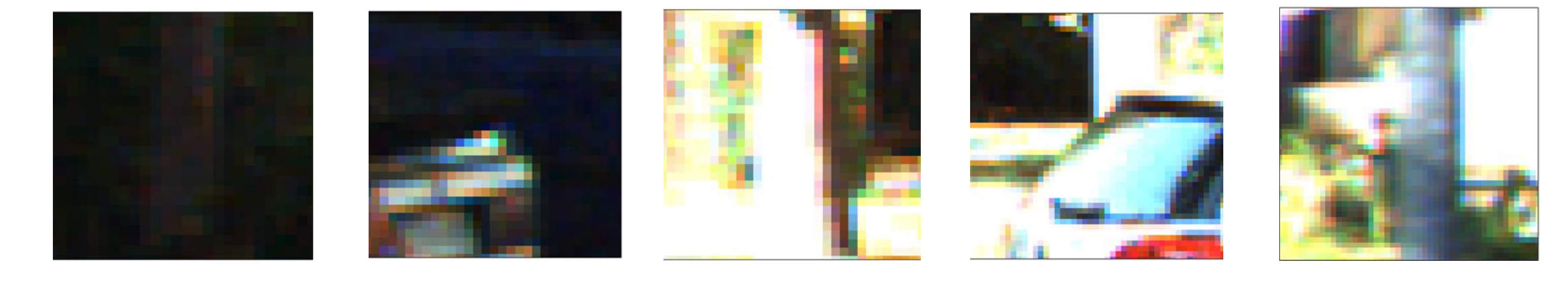}} \vspace{-3mm}\\
        \sidesubfloat[]{\includegraphics[trim=32 250 40 120,clip,width=0.96\linewidth]{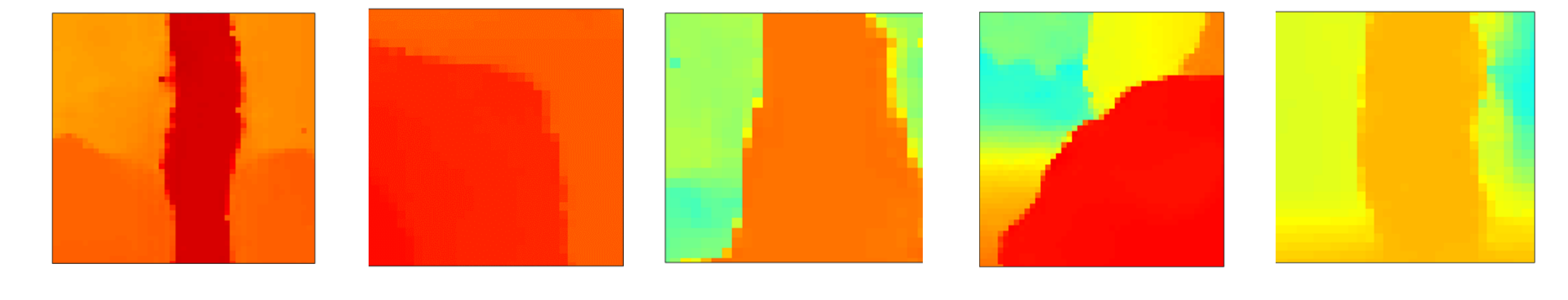}} \vspace{-3mm}\\
          \sidesubfloat[]{\hspace{0.5mm}\includegraphics[trim=32 250 40 120,clip,width=0.96\linewidth]{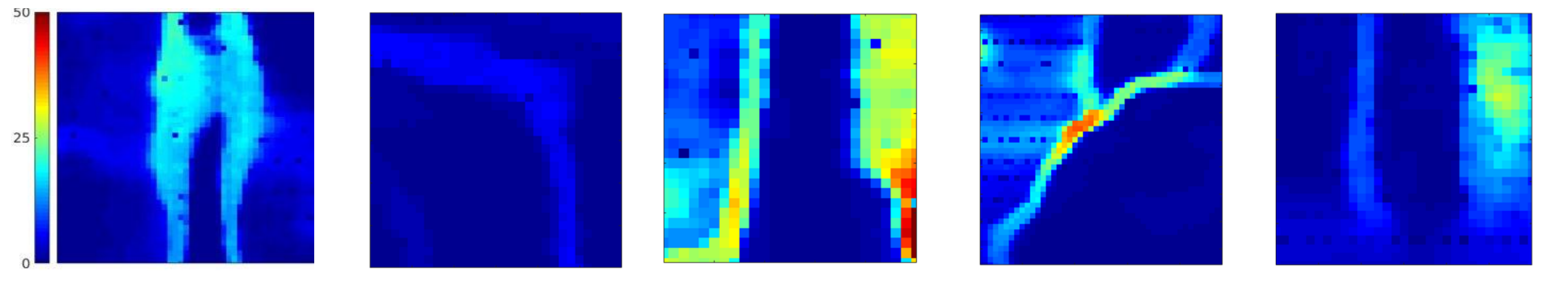}}\vspace{-5mm}
    \caption{\small Input image (a), its zoom-in views (d), our estimation on foreground depth (b), background depth (c), fused depth (e), and the depth difference between foreground and background depth (f). \vspace{-3mm} 
    }
    \label{fig:indep_anal}
    \end{center}
\end{figure}

\Paragraph{Qualitative Parsing}
Fig.~\ref{fig:indep_anal} offers a more detailed analysis of our method by showing different estimation at foreground, background depths and fused depth respectively. We choose five zoom-in views from diverse objects, {\it e.g.}, tree, poles, car, and even pixels at far-away depth pixels. 
It shows that our fused depth estimator can learn to choose foreground and background regions well, resulting in a clear shape estimation of objects.  We note that it is biased to choose the foreground surface as ambiguity increases, {\it e.g.}, relatively large depth gap between foreground and background surfaces (see depth difference in Fig.~\ref{fig:indep_anal} (f)). 
This can be explained by the fact that there are more supervision at the close-up region than the far-away region on account of uneven distribution of GT depth pixels. 

\Paragraph{Quantitative Results on NYU2}: Results on NYU2 are shown in Tab.~\ref{tab:nyu2res}, based on its standard metrics. We are currently ranked the \textit{second} in all standard metrics. Note that compared to NLSPN~\cite{park2020non}, ours is $10\times$ faster in inference on KITTI. The results also show that TWISE is equally generalizable to indoor scenes. 

\begin{table}[t!]
\captionsetup{font=small}
    \centering
    \scalebox{0.8}{
    \begin{tabular}{|c|c|c|c|c|c|}
    \hline
       Method & \makecell{RMSE (m)} & REL & $\delta_{1.25}$ & $\delta_{1.25}^2$ & $\delta_{1.25}^3$  \\
       \hline
       DC-3co \cite{imran2019depth} & $0.118$ & $0.013$ & $99.4$ & \boldsymbol{$99.9$} & \boldsymbol{$100.0$} \\
       \hline
       DeepLidar \cite{qiu2019deeplidar} & $0.115$ & $0.022$ & $99.3$ & \boldsymbol{$99.9$} & \boldsymbol{$100.0$} \\
       \hline
       DepthNormal \cite{xu2019depth} & $0.112$ & $0.018$ & $99.5$ & \boldsymbol{$99.9$} & \boldsymbol{$100.0$} \\
       \hline
        GNN \cite{xiong2020sparse} & $0.106$ & $0.016$ & \boldsymbol{$99.6$} & \boldsymbol{$99.9$} & \boldsymbol{$100.0$} \\
        \hline
        TWISE & $0.097$ & $0.013$ & \boldsymbol{$99.6$} & \boldsymbol{$99.9$} & \boldsymbol{$100.0$} \\
        \hline
        NLSPN \cite{park2020non} & \boldsymbol{$0.092$} & \boldsymbol{$0.012$} & \boldsymbol{$99.6$} & \boldsymbol{$99.9$} & \boldsymbol{$100.0$} \\
        \hline
    \end{tabular}}\vspace{-2mm}
    \caption{\small{Depth completion results on NYU2~\cite{Silberman:ECCV12}.}\vspace{-2mm}}
    \label{tab:nyu2res}
\end{table}

\subsection{Ablation Studies}

In this section, we conduct extensive ablation studies to investigate the effect of different parameters of our proposed loss. 
We train with $1/6$ data ($\sim$$12$K training samples) due to resource constraints, and maintain this protocol for all ablations unless otherwise noted. 

\Paragraph{Effect of Loss Functions}
We show that performance of our loss function is network agnostic.  
Tab.~\ref{tab:effect_of_loss} refers to different loss functions typically used in SoTA depth estimation works. 
Although $L_2$ is a widely used loss for estimating depth~\cite{ma2019self, Li_2020_WACV, chen2018estimating}, $L_1$ loss~\cite{mal2018sparse}, Huber loss~\cite{chen2019learning}, $L_1+L_2$~\cite{park2020non} are some of the widely used losses for depth completion. 
We compare our TWISE loss with all others, including the CE loss~\cite{imran2019depth}. 
Top performances on MAE and TMAE show the positive side effect of our loss addressing the smearing problem at the boundary. 
We particularly note that TWISE performs better than a standard $L_1$ loss on both the backbone networks, leading to believe that TWISE offers more benefit than a mere trade-off between MAE and RMSE.

\begin{table}[t!]
\captionsetup{font=small}
    \centering
    \scalebox{0.6}{
    \begin{tabular}{|c|c|c|c|c|c|c|c|c|}
    \hline
         &  \multicolumn{4}{c|}{Res-18 \cite{ma2019self}} & \multicolumn{4}{c|}{MultiStack \cite{Li_2020_WACV}} \\
         \hline
        Loss & MAE & RMSE & TMAE & TRMSE & MAE & RMSE & TMAE &TRMSE \\ \hline
        $L_1$~\cite{mal2018sparse} & $282.6$ & $110.6$ & $181.8$ & $295.6$ & $211.0$ & $950.0$ & $138.6$ & $246.0$ \\ \hline
        $L_2$~\cite{ma2019self} & $341.2$ & $987.8$ & $244.6$ & $349.5$ & $247.4$ & $880.0$ & $170.3$ & $285.0$ \\ \hline
        $L_2$+$L_1$~\cite{park2020non} & $298.8$ & \boldsymbol{$972.2$} & $206.5$ & $316.7$ & $231.8$ & $\boldsymbol{887.5}$ & $156.9$ & $271.2$ \\ \hline
        Huber~\cite{chen2019learning} & $288.6$ & $1039.6$ & $198.0$ & $302.1$ & $222.6$ & $927.1$ & $153.9$ & $256.0$ \\ \hline
        CE~\cite{imran2019depth} & $279.1$ & $1125.1$ & $184.3$ & \boldsymbol{$239.1$} & -- & -- & -- & --\\ \hline
        TWISE & \boldsymbol{$275.5$} & $1045.1$ & \boldsymbol{$181.1$} & $294.0$ & \boldsymbol{$201.3$} & $927.6$ & \boldsymbol{$134.1$} & \boldsymbol{$240.1$} \\ \hline
    \end{tabular}}
    \vspace{-3mm}
   \caption{\footnotesize Effect of different loss functions. Compared to single channel losses, CE requires $80$ channel, while TWISE requires 3 channel. \vspace{-2mm}}
   \label{tab:effect_of_loss}
\end{table}

\begin{table}[t!]
\captionsetup{font=small}

    \hfill
    
    \centering
      \scalebox{0.76}{
    \begin{tabular}{|c|c|c|c|c|}
    \hline
        Options & MAE & RMSE & TMAE & TRMSE \\ \hline
        $\hat{d_t} = \hat{d_1}$ ($\sigma=1$) & $306.9$ & $1109.9$ & $204.4$ & $314.8$ \\ \hline
        $\hat{d_t} = \hat{d_2}$ ($\sigma=0$) & $295.4$ & $1092.9$ & $193.9$ & $306.1$ \\ \hline
      $\hat{d_t}=0.5*(\hat{d_1} + \hat{d_2})$ ($\sigma=0.5$) & $220.7$ & \boldsymbol{$854.8$} & $148.2$ & $262.4$ \\ \hline
        $\hat{d_t}=\hat{d_1}/\hat{d_2}|\sigma>0.5$ & $261.0$ & $1008.0$ & $180.4$ & $287.9$ \\
        \hline
        No color & $222.4$ & $1067.5$ & $139.2$ & $247.8$ \\
        \hline
        $\hat{d_t}= \sigma \hat{d_1} + (1 - \sigma)\hat{d_2}$ & \boldsymbol{$193.4$} & $879.4$ & \boldsymbol{$131.1$} & \boldsymbol{$236.0$} \\
        \hline
    \end{tabular}}
\caption[]{\small Effect of learned $\sigma$ in TWISE, evaluated by our best model.} 
\vspace{-3mm}
\label{tab:effect_sigma}

\end{table}

\begin{table}[t!]
\captionsetup{font=small}
    \centering
    \scalebox{0.8}{
    \begin{tabular}{|c|c|c|c|c|}
    \hline
        $\gamma$ & MAE & RMSE & TMAE & TRMSE \\ \hline
        $ 1.0$ & $223.1$ & $950.1$ & $145.8$ & $257.0$ \\ \hline
        $ 1.5$ & $207.8$ & $947.9$ & $138.1$ & $245.1$ \\ \hline
        $ 2.0$ & \boldsymbol{$201.3$} & $927.6$ & \boldsymbol{$134.1$} & \boldsymbol{$240.1$} \\ \hline
        $ 2.5$ & $204.4$ & $932.5$ & $136.1$ & $242.5$ \\ \hline
        $ 5.0$ & $207.1$ & $923.4$ & $138.7$ & $246.1$ \\ \hline
        $10$ & $216.1$ & \boldsymbol{$922.8$} & $146.7$ & $255.4$ \\ \hline
    \end{tabular}}
    \vspace{1mm}
    \caption{\small Effect of $\gamma$ on depth completion performance.} \vspace{-3mm}
    \label{tab:effect_of_3chanloss}
\end{table}

\begin{figure*}[h!]
\captionsetup{font=small}
    \centering
    \RawFloats
    \parbox{0.3\textwidth}{
\scalebox{0.62}{
    \centering
    \begin{tabular}{|c|c|c|c|c|c|c|c|c|c|}
       \hline
    \multicolumn{2}{|c|}{Supervision} & \multicolumn{4}{|c|}{Noisy Semi-Dense GT} & \multicolumn{4}{|c|}{Clean GT} \\ \hline
     Backbone &  Method & MAE & RMSE & TMAE & TRMSE &
                 MAE & RMSE & TMAE & TRMSE \\ \hline
    \multirow{4}{*}{\begin{tabular}{@{}c@{}}MultiStack\\ \cite{Li_2020_WACV}\end{tabular}} & $L_1$ & $8.79$ & $49.9$ & $7.09$ & $16.02$ & $14.43$ & $130.62$ & $6.16$ & $18.28$ \\
    & $L_2$ & $10.40$ & $45.35$ & $8.61$ & $17.89$ & $17.75$ & $127.14$ & $8.45$ & $20.18$ \\
    & $L_1+L_2$ & $9.42$ & \boldsymbol{$44.90$} & $8.23$ & $16.82$ &
                   $15.45$ & \boldsymbol{$126.20$} & $7.14$ & $19.30$ \\
       & TWISE & \boldsymbol{$7.98$} & $47.5$ & \boldsymbol{$6.25$} & \boldsymbol{$15.35$} &
                  \boldsymbol{$12.71$} & $126.4$ & \boldsymbol{$5.22$} & \boldsymbol{$16.67$} \\
                  \hline
    \multirow{4}{*}{\begin{tabular}{@{}c@{}}ResNet-18\\ \cite{ma2019self}\end{tabular}}    & CE & $10.50$ & $58.67$ & $8.64$ & \boldsymbol{$16.57$} &
                $19.03$ & $155.24$ & \boldsymbol{$8.26$} & $18.29$ \\
       & $L_1$ & $12.95$ & $62.97$ & $14.68$ & $19.25$ &
                $23.52$ & $147.42$ & $12.51$ & $29.33$ \\
        & $L_2$ & $17.45$ & $50.48$ & $17.48$ & $21.26$ &
                $27.48$ & $133.21$ & $15.57$ & $36.73$ \\
        & $L_1+L_2$ & $14.21$ & \boldsymbol{$48.25$} & $15.80$ & $20.10$ &
                   $25.40$ & \boldsymbol{$132.6$} & $14.35$ & $32.47$ \\
        & TWISE & \boldsymbol{$10.24$} & $52.37$ & \boldsymbol{$8.42$} & $16.77$ &
                   \boldsymbol{$18.88$} & $132.94$ & $9.45$ & \boldsymbol{$18.17$} \\
                   \hline
    \end{tabular}
    }\vspace{-2.5mm}
    \captionsetup{labelformat=empty}
    \caption{\hspace{55mm}(a)}}
    \hspace{5.6mm}
    \parbox{0.29\textwidth}{
    \scalebox{0.95}{
        \hbox{\hspace{8.04em}}
         \includegraphics[trim=4 2 18 17,clip, width=0.24\textwidth, right]{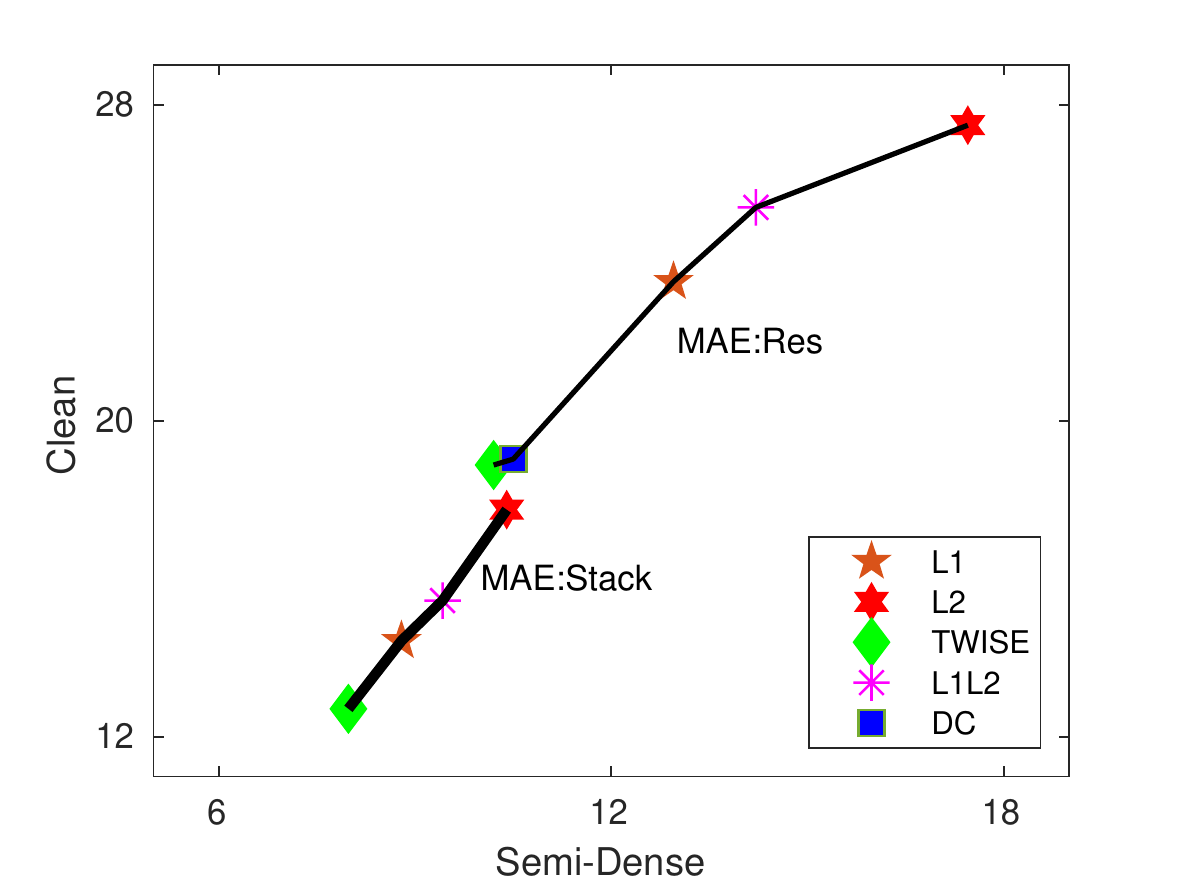}}\vspace{-2.5mm}
    \captionsetup{labelformat=empty}
    \caption{\hbox{\hspace{17.5em}}(b)}
}
 \hspace{9mm}
 \parbox{0.29\textwidth}{
    \scalebox{0.95}{
        \hbox{\hspace{1.4em}}
         \includegraphics[trim=2 2 20 17,clip, width=0.24\textwidth, right]{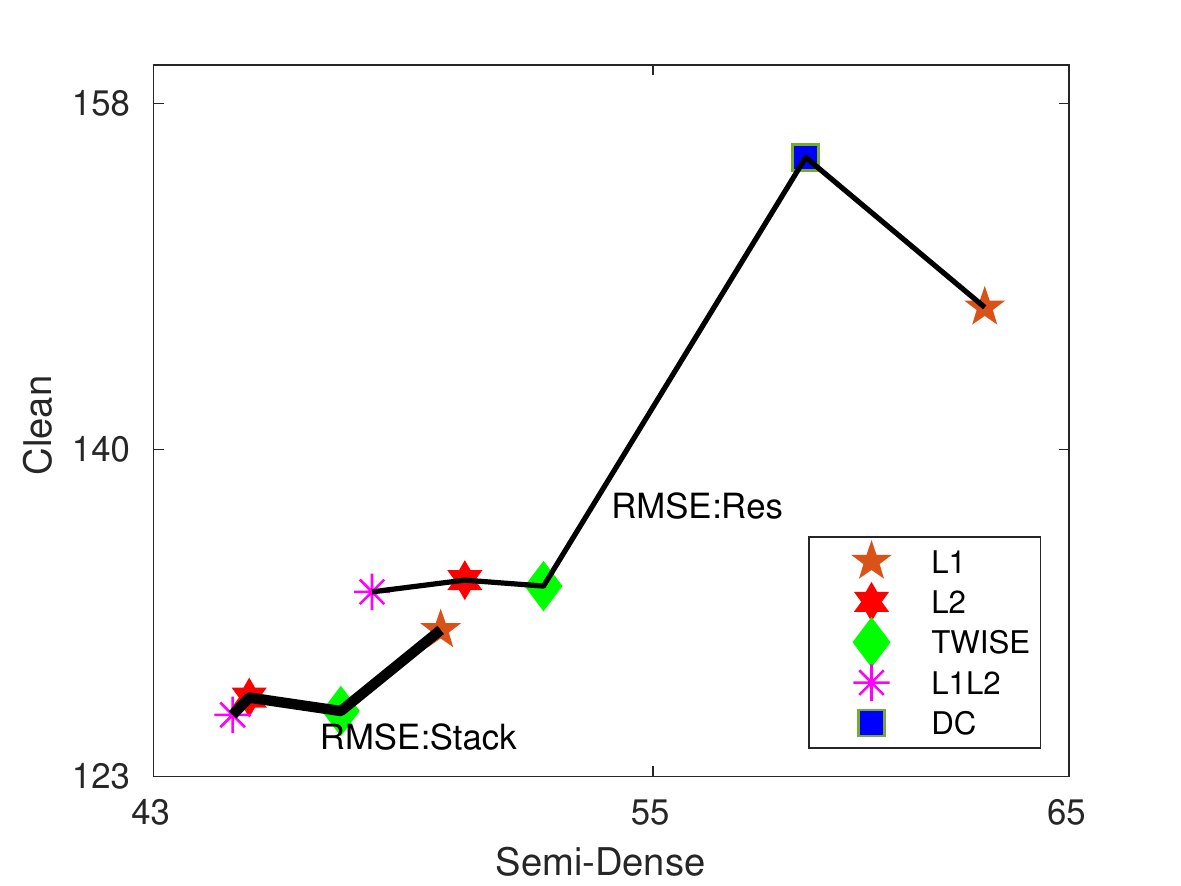}}\vspace{-2.5mm}
    \captionsetup{labelformat=empty}
    \caption{\hbox{\hspace{4.2em}}(c)}
}
\vspace{-2mm}
\caption{\small (a) Results on Virtual KITTI experiments trained on clean GT and synthesized semi-dense respectively (units in cm). (b) MAE and (c) RMSE curves of scatter plots (Semi-Dense vs Clean GT) for different loss functions (colored symbols) and two backbone networks (MultiStack~\cite{Li_2020_WACV} and ResNet-18~\cite{ma2019self}). Methods trained with the same backbone network are connected. }\vspace{-1mm}
\label{fig:vkitti_results}
\end{figure*}

\Paragraph{Effect of $\sigma$ on Estimated Surfaces}
Another interesting evaluation is the importance of learned $\sigma$ on different estimated surfaces. 
In Tab.~\ref{tab:effect_sigma}, we evaluate estimated depths for different combinations of $\sigma$ and compare individually its depth completion metrics. 
The performance is evaluated on our best model in Tab.~\ref{tab:kitti_results}, except for the row with ``no color'', where we train without color input on the same network of our best model. 
From Tab.~\ref{tab:effect_sigma}, foreground and background depth surface estimates, as usual, have higher error metric, since they are individually a biased estimate of depth. 
If we fix $\sigma$ at $0.5$, we see it is possible to achieve decent performance on MAE and RMSE on account of averaging (interpolation) between the two surfaces. We make a binary choice between foreground and background surface if $\sigma >0.5$ and the results are worse than averaging. 
In addition, we see $\sigma$ does not learn effectively without color input.
So high-resolution imagery helps to learn effective $\sigma$ and resolve ambiguities at the boundaries.

\Paragraph{Effect of $\gamma$ on Performance}
Since $\gamma$ impacts the separation of foreground and background surfaces, we perform an ablation to assess its impact on TWISE. Tab.~\ref{tab:effect_of_3chanloss} shows depth completion performance with several $\gamma$ values.
With $\gamma = 1$, the loss is equivalent to MAE. As $\gamma$ increases, the gap between foreground and background surface increases. At small $\gamma$ values, the interpolation benefits, thus leading to lower MAE, TMAE, TRMSE, since it is easier to interpolate between two nearby surfaces; however, in the meantime extrapolation suffers, thus leading to higher RMSE. 
At larger $\gamma$, the slope between two surfaces increase, and interpolation becomes harder. 
We choose $\gamma=2.0$ in our experiment as a compromise between interpolation and extrapolation.

\Paragraph{Effect of Sparsity on Depth Performance}
We also ran an extensive ablation study on generalization of SoTA methods due to sparsity. Sparsity is created by subsampling LiDAR-points in azimuth-elevation space to simulate LiDAR-like structured patterns. All the SoTA methods compared have been retrained using the author provided code with variable sparse input patterns. Tab.~\ref{tab:eff_sparsity} shows that TWISE has better generalization and exhibits significantly less errors in all the metrics compared to SoTA methods. With more sparsity, TWISE is able to beat the RMSE metrics of methods supervised by standard losses. Particularly interesting is the fact that TWISE can be used for monocular depth estimation with no sparse depth input.

\begin{table}[t!]
\captionsetup{font=small}
    \centering
    \scalebox{0.78}{
    \begin{tabular}{|c|c|c|c|c|c|}
       \hline
        Sparsity & Method & MAE & RMSE & TMAE & TRMSE  \\ \hline
        \multirow{4}{*}{$64$R} & DC \cite{imran2019depth} & $279.1$ & $1125.1$ & $183.1$ & $292.3$ \\ 
        & MultiStack~\cite{Li_2020_WACV} & $229.4$ & $889.7$ & $156.8$ & $265.0$ \\ 
        & NLSPN~\cite{park2020non} & $219.1$ & \boldsymbol{$868.0$} & $147.7$ & $263.4$ \\
        & TWISE & \boldsymbol{$201.3$} & $927.6$ & \boldsymbol{$134.1$} & \boldsymbol{$240.1$} \\
        \hline 
        \multirow{4}{*}{$32$R} & DC & $392.7$ & $1456.2$ & $232.1$ & $350.7$ \\ 
        & MultiStack & $439.2$ & $1288.8$ & $275.4$ & $402.3$ \\ 
        & NLSPN & $392.4$ & \boldsymbol{$1229.2$} & $248.2$ & $373.8$ \\ 
        & TWISE & \boldsymbol{$327.9$} & $1242.6$ & \boldsymbol{$204.9$} & \boldsymbol{$324.3$} \\
        \hline 
        \multirow{4}{*}{$16$R} & DC & $477.7$ & $1777.3$ & $259.5$ & $382.9$ \\ 
        & MultiStack & $528.4$ & $1504.3$ & $308.6$ & $439.5$ \\ 
        & NLSPN & $497.1$ & $1483.1$ & $286.8$ & $419.2$ \\ 
        & TWISE & \boldsymbol{$414.0$} & \boldsymbol{$1481.1$} & \boldsymbol{$237.3$} & \boldsymbol{$365.1$} \\
        \hline
        \multirow{4}{*}{$8$R} & DC & $634.7$ & $2311.9$ & $288.5$ & $420.6$ \\ 
        & MultiStack & $672.58$ & $1841.6$ & $353.2$ & $486.8$ \\ 
        & NLSPN & $669.05$ & $1869.5$ & $340.3$ & $475.2$ \\ 
        & TWISE & \boldsymbol{$532.1$} & \boldsymbol{$1782.5$} & \boldsymbol{$275.6$} & \boldsymbol{$409.4$} \\
        \hline
        \multirow{4}{*}{RGB} & DC & $2423.8$ & $4433.6$ & $715.4$ & $797.2$ \\ 
        & MultiStack & $2070.4$ & $4185.1$ & $635.7$ & $735.4$ \\ 
        & NLSPN & $2192.9$ & $4362.35$ & $646.0$ & $743.6$ \\ 
        & TWISE & \boldsymbol{$1964.1$} & \boldsymbol{$4078.8$} & \boldsymbol{$612.0$} & \boldsymbol{$716.5$} \\ 
        \hline
    \end{tabular}}
    \vspace{-3mm}
    \caption{\small Row sparsity impact on SoTA depth completion methods.\vspace{-5mm}}
    \label{tab:eff_sparsity}
\end{table}

\Paragraph{Synthetic Experiments with VKITTI}
Using both semi-dense GT and clean GT of VKITTI, we ran experiments on different loss functions using two different backbone networks. The conclusion is drawn by training and evaluation on noisy semi-dense and clean GT respectively. The results are shown in Fig.~\ref{fig:vkitti_results} (a).   
Several inferences can be drawn from the scatter plot of Fig.~\ref{fig:vkitti_results} (b) and (c). Firstly, the MAE score is smooth and monotonic as opposed RMSE which zigzags. This implies that given a MAE score on semi-dense, we are able to predict its score on the clean dataset as well. Additionally, the ranking of the methods in both the datasets is the same for MAE but not RMSE. As a result, we can conclude that MAE is a superior metric to RMSE for comparing and ranking depth completion methods.


Secondly, TWISE is more than a trade-off between MAE and RMSE. One of the objective of TWISE is to improve depth points at discontinuity regions. But KITTI semi-dense GT lacks dense ground-truth depth points, and contains more outliers in the boundary regions owing to methodology adopted in creating the GT. In presence of outliers, RMSE in TWISE suffers the most, but when clean GT can be provided,  RMSE in TWISE performs as well as those methods with the $L_2$ loss.

\section{Conclusion}
In this paper we propose TWISE, a new twin-surface representation and estimation method for depth images.  Our proposed asymmetric loss functions, \ALE{} and \RALE{}, bias these twin surface estimates towards the foreground and background at pixels with depth ambiguity.  A third channel of our output fuses these estimates to achieve a single surface estimate.  This solution simplifies the task of learning depth discontinuities, and as a result better maintains step-wise depth discontinuities across boundaries, and generates SOTA depth estimates. We also compared the robustness of MAE and RMSE as metrics for ranking depth completion methods and our analysis suggests that MAE is a superior metric in presence of noisy GT datasets.
In future, we would like to improve our estimates at far-away depth pixels where learning suffers due to sparsity of ground-truth pixels. 


{\small
\bibliographystyle{ieee_fullname}
\bibliography{abbrev, egbib}
}

\end{document}


\title{Depth Completion with Twin Surface Extrapolation at Occlusion Boundaries: Supplementary Material}

\author{Saif Imran \hspace{1cm} Xiaoming Liu \hspace{1cm} Daniel Morris\\
Michigan State University\\
{\tt\small \{imransai, liuxm, dmorris\}@msu.edu}
}

\maketitle
\thispagestyle{empty}
\begin{abstract}
In the supplementary section, we provide additional insights of our results with SoTA methods, show evidences of boundary outliers on KITTI semi-dense ground-truth and its effect on depth completion performance, and discuss our data generation process in KITTI and Virtual KITTI used for our ablation study in the main paper.
\end{abstract}

\section{Relative Error Maps}

It is worthwhile to examine where our method has lower errors in comparison with majority of the SoTA methods which use MSE. For this purpose, we choose the MultiStack method~\cite{Li_2020_WACV} for comparison.
we calculate the difference of error maps of  Absolute Error, $A(i)$, and Squared Error, $S(i)$, of two methods respectively to show the gains of our method over MultiStack \cite{Li_2020_WACV}. The error differences are calculated by the following equation: 
\begin{equation}
   A(i) = |\hat{d}_M(i) - d_t(i)| - |\hat{d}_{T}(i) - d_t(i)|,
\end{equation}
\begin{equation}
   S(i) = |\hat{d}_M(i) - d_t(i)|^2 - |\hat{d}_T(i) - d_t(i)|^2,
\end{equation}
where $\hat{d}_M$ and $\hat{d}_T$ are depth estimates of MultiStack \cite{Li_2020_WACV} and TWISE respectively. $A(i)$ and $S(i)$ are Absolute Error Difference and Squared Error Difference of pixel $i$ on two competing methods respectively. For a particular pixel, when $A(i)$ and $S(i)$ is ($+$)ve, TWISE is performing better then MultiStack and vice-versa for ($-$)ve values. We note that the errors are evaluated only where there are valid ground-truth pixels.

\begin{figure}[t!]
    \centering
    \begin{tabular}{@{}c@{\hskip1pt}c}
        {\footnotesize ($a$)} & \includegraphics[trim=70 250 95 100,clip,width=0.96\linewidth]{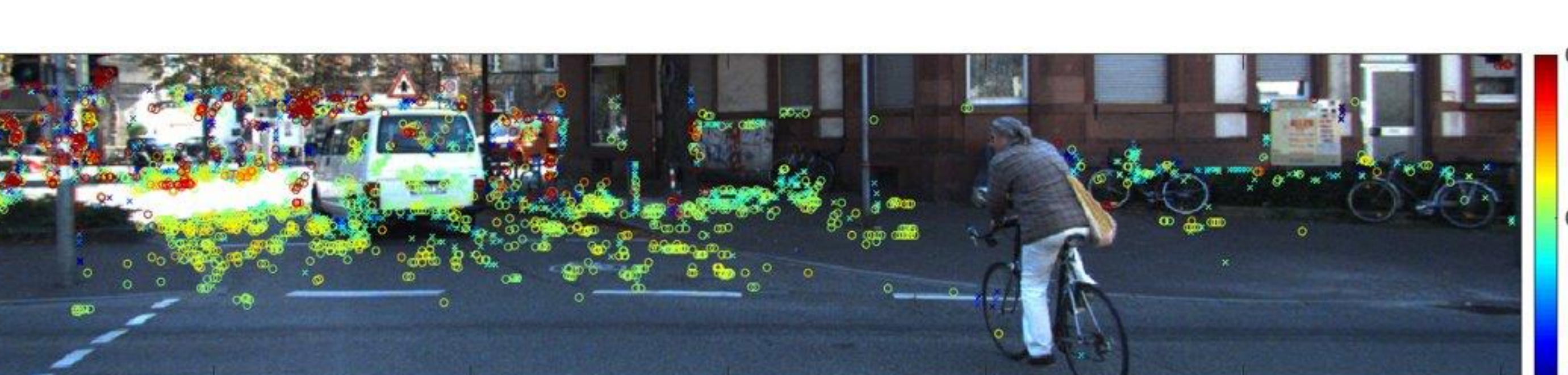} \vspace{-1mm} \\
        {\footnotesize ($b$)} & \includegraphics[trim=72 250 95 100,clip,width=0.96\linewidth]{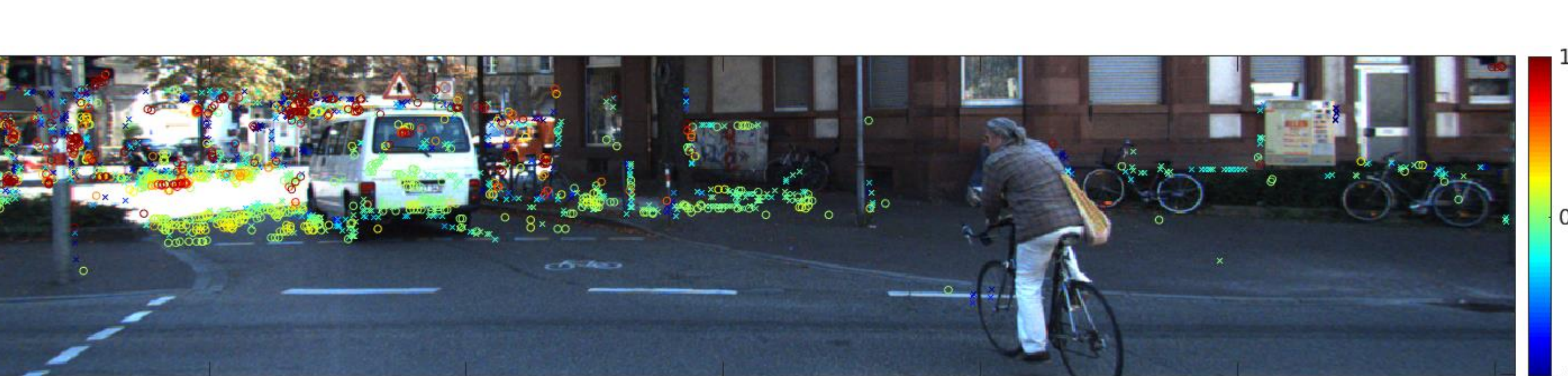} \vspace{-1mm} \\
    \end{tabular}
    \caption{\small Difference of TWISE vs MultiStack~\cite{Li_2020_WACV} in ($a$) Absolute Error (AE) and ($b$) Squared Error (SE) respectively. The red indicates the most gain of ours over \cite{Li_2020_WACV}, marked by 'o'; while the blue is vice-versa, marked by 'x'. Zoom in for details. \vspace{-2mm}}
    \label{fig:relerr_map}
\end{figure}

As shown in Fig.~\ref{fig:relerr_map}, our method wins in substantially more pixels than losing. Errors in our method often comes from few pixels at boundary regions, when a FG depth is erroneously chosen over a BG depth/vice versa;  we term them as outliers {\it e.g.}, see depth error at the traffic sign pixels, edge of tree-trunk etc close to/at the boundary. 
These outliers with large depth errors are strongly weighted by the RMSE metric, leading to our worse performance on that metric. 

To further our analysis, we do a statistical evaluation on $200$ samples of the validation set (chosen every $5$ samples from KITTI's $1,000$ validation set) to confirm that TWISE has better depth estimate on most pixels compared to MultiStack \cite{Li_2020_WACV} except for few erroneous pixels (outliers) at boundaries (see Fig. \ref{fig:relerr_map}). 

\begin{figure}[h]
    \centering
    \begin{tabular}{@{}c@{\hskip1pt}c}
         \includegraphics[trim=0 2 25 5,clip, width=0.5\linewidth]{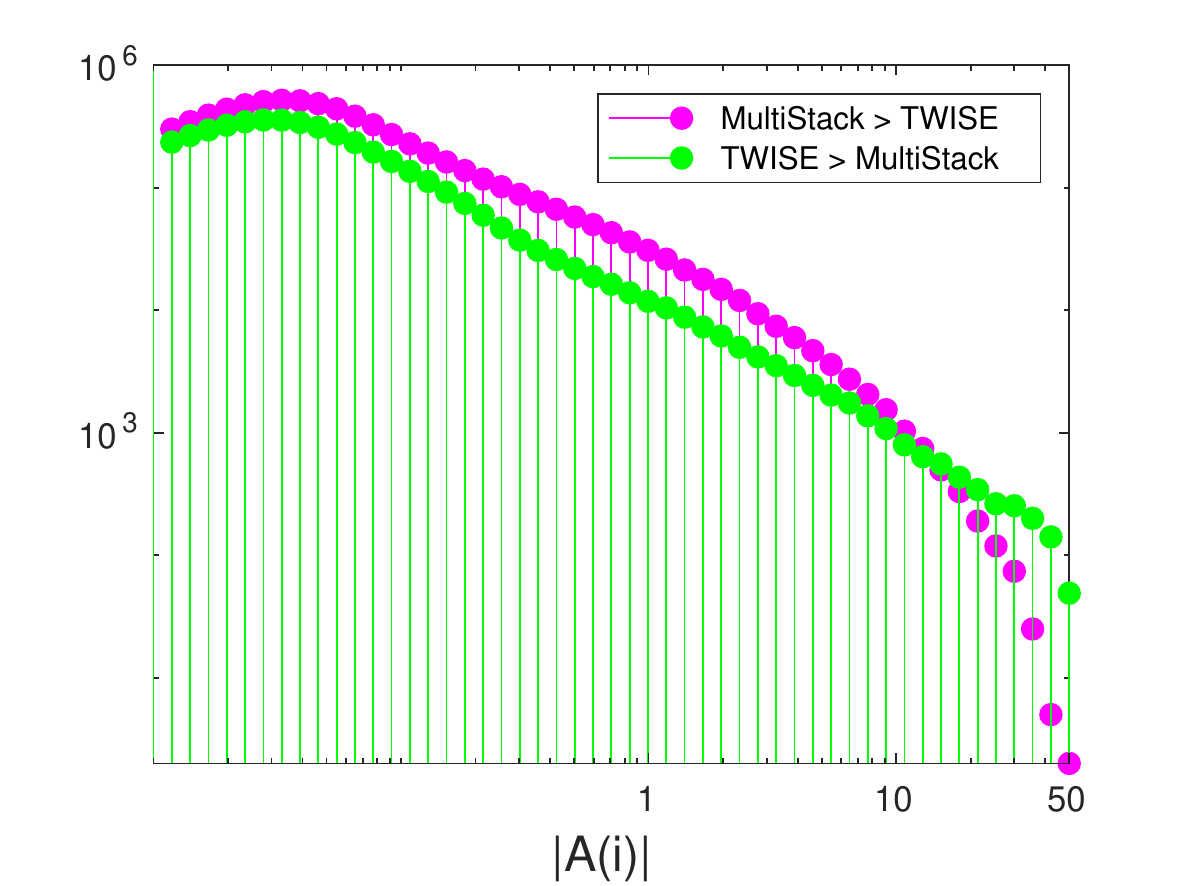}  &  
         \includegraphics[trim=0 2 25 5,clip, width=0.5\linewidth]{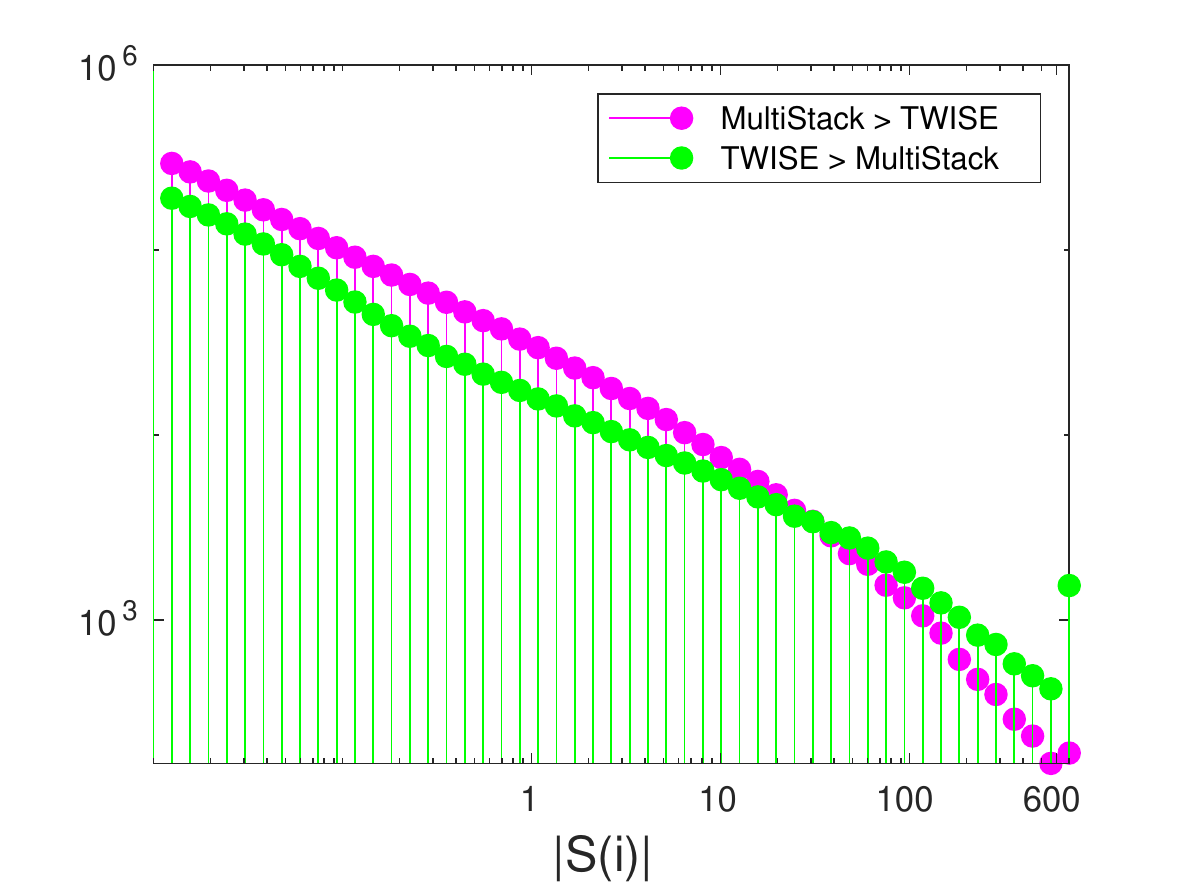} \\
         ($a$) & ($b$) \\
    \end{tabular}
    \caption{($a$) Magenta is a histogram of absolute error differences $A(i)$ for $A(i)>0$ (where MultiStack errors > TWISE errors) and green is a histogram of $|A(i)|$ for $A(i)<0$ (where TWISE errors > MultiStack errors).  ($b$) Corresponding histograms for squared pixel error differences $S(i)$. } 
    \label{fig:relerror_stats}
\end{figure}

We do a histogram binning of $A(i)$ for pixels where $A(i) > 0$ (Multistack > TWISE is equivalent to performance gain of TWISE over MultiStack) and of $|A(i)|$ for pixels where  $A(i) < 0$ (TWISE > MultiStack is equivalent to performance gain of MultiStack over TWISE). There histograms are plotted together in Fig. \ref{fig:relerror_stats}($a$).  Analogous histograms are plotted for the squared error difference, $S(i)$, in Fig. \ref{fig:relerror_stats}($b$).  These histograms show that TWISE has less error than Multi-Stack \cite{Li_2020_WACV} for most pixels ($\sim 2.70*10^6$) compared to just ($\sim 6,100$) pixels where Multi-stack bests TWISE.  The average image in this set has $13,500$ pixels where TWISE is better versus $31$ pixels where MultiStack is better.

\begin{figure}[h!]
    \centering
        \includegraphics[trim=70 20 70 20,clip, width=0.87\linewidth]{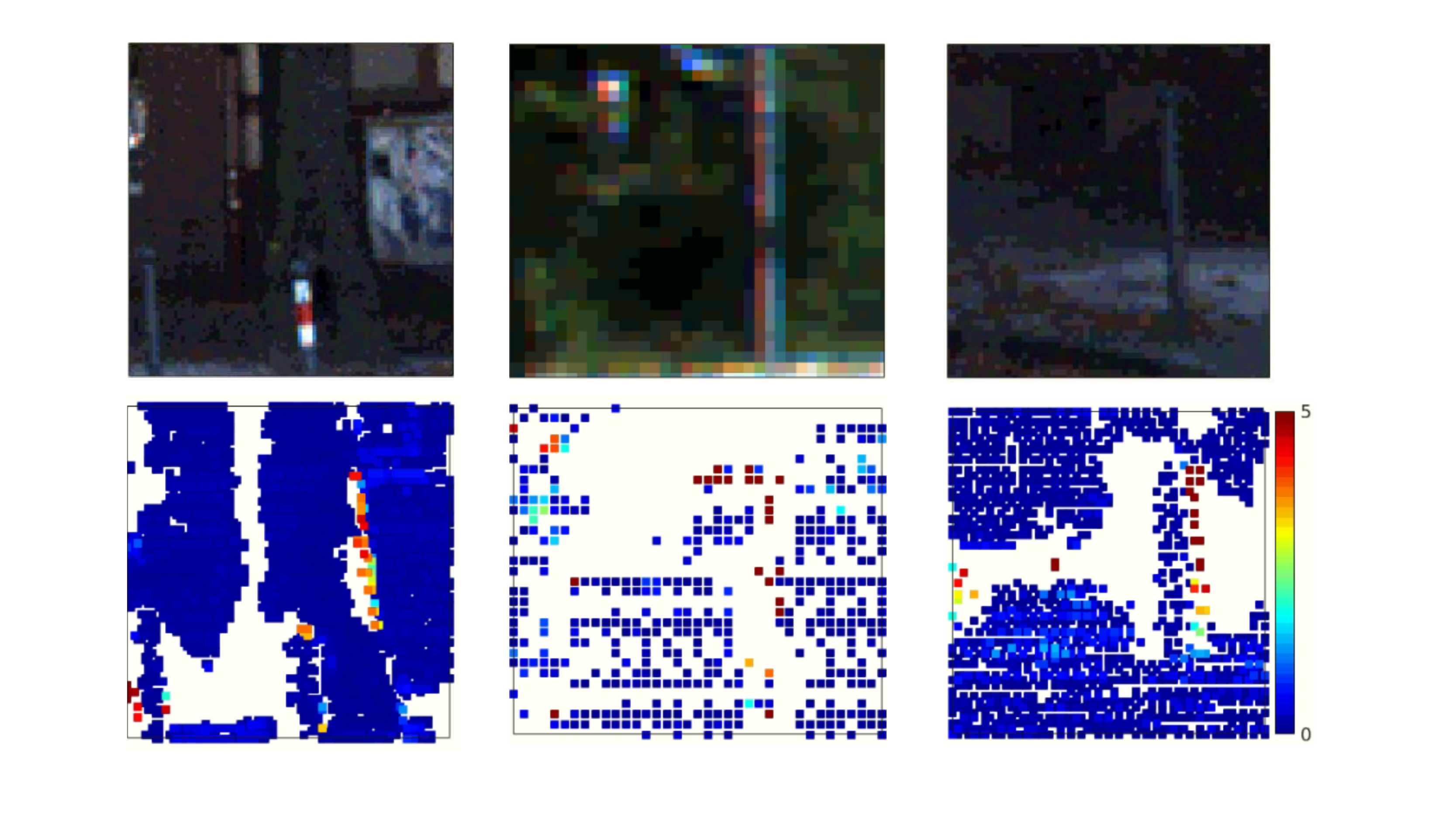}  
    \vspace{-2mm}
    \caption{Color images (top) and depth error maps in $0-5$m (bottom).}
    \label{fig:rep_errors}
\end{figure}

The reason for large RMSE errors in TWISE is believed to be caused by the outliers (erroneous FG/BG depth selection by TWISE) closer to object boundaries. The outliers are penalized heavily by RMSE metric as opposed to floating depth pixels estimated by MultiStack; as a result, our depth estimate suffers in that metric. As representative examples in Fig.~\ref{fig:rep_errors},  the error maps show depth errors around the boundary, and missing thin objects like poles. The reasoning can be further enhanced by the Tab. \ref{tab:err_bound}. In this analysis, we leverage GT semantics provided by KITTI semantic segmentation dataset. In $140$ images, FG objects are poles, boundaries, traffic signs, vehicle, person and the rest as background. For each image, we label all pixels where whose distances to object boundareis are less than $3$ pixels as edge pixels and the remaining as inside object pixels. Tab. \ref{tab:err_bound} validates substantial larger errors are around boundary.

\begin{table}[t!]
    \centering
    \scalebox{0.85}{
    \begin{tabular}{|c|c|c|c|c|}
    \hline
        Area & MAE & RMSE & TMAE & TRMSE \\
         \hline
         Inside Object & $196.1$ & $752.3$ & $138.6$ & $327.3$ \\
        Edge Pixels & $731.6$ & $2396.9$ & $304.4$ & $454.6$ \\
        Whole Image & $215.1$ & $880.9$ & $144.6$ & $254.3$ \\
        \hline
    \end{tabular}}\vspace{2mm}
    \caption{Error metrics for different image regions on TWISE.}
    \label{tab:err_bound}
\end{table}


\begin{figure*}[h]
    \centering
    \begin{tabular}{@{}c@{\hskip1pt}c@{}c@{\hskip1pt}c}
      {\footnotesize ($a$)} & \includegraphics[trim=5 100 20 10,clip, width=0.45\linewidth]{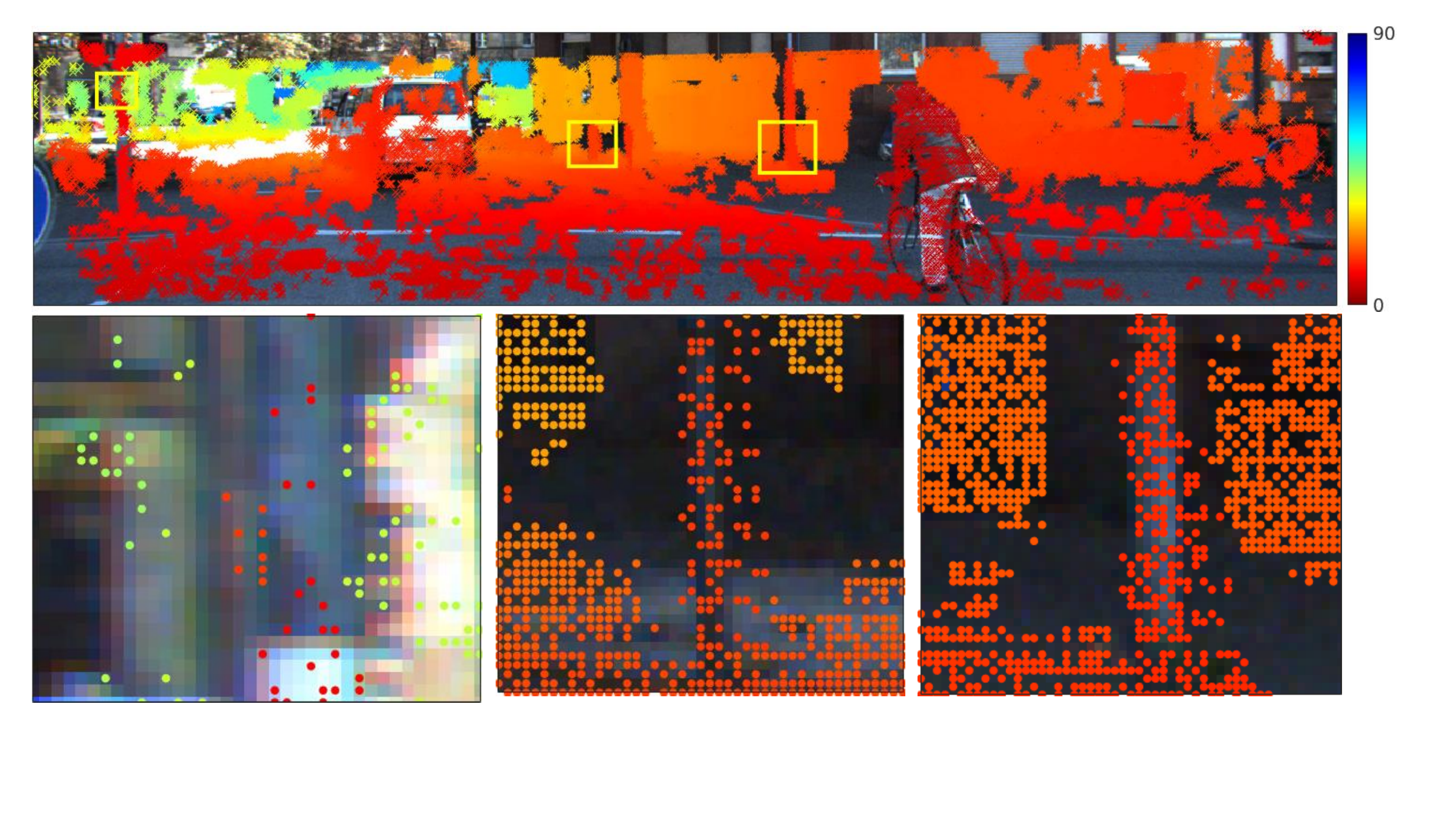} & {\footnotesize ($b$)} & 
      \includegraphics[trim=5 100 20 10,clip, width=0.45\linewidth]{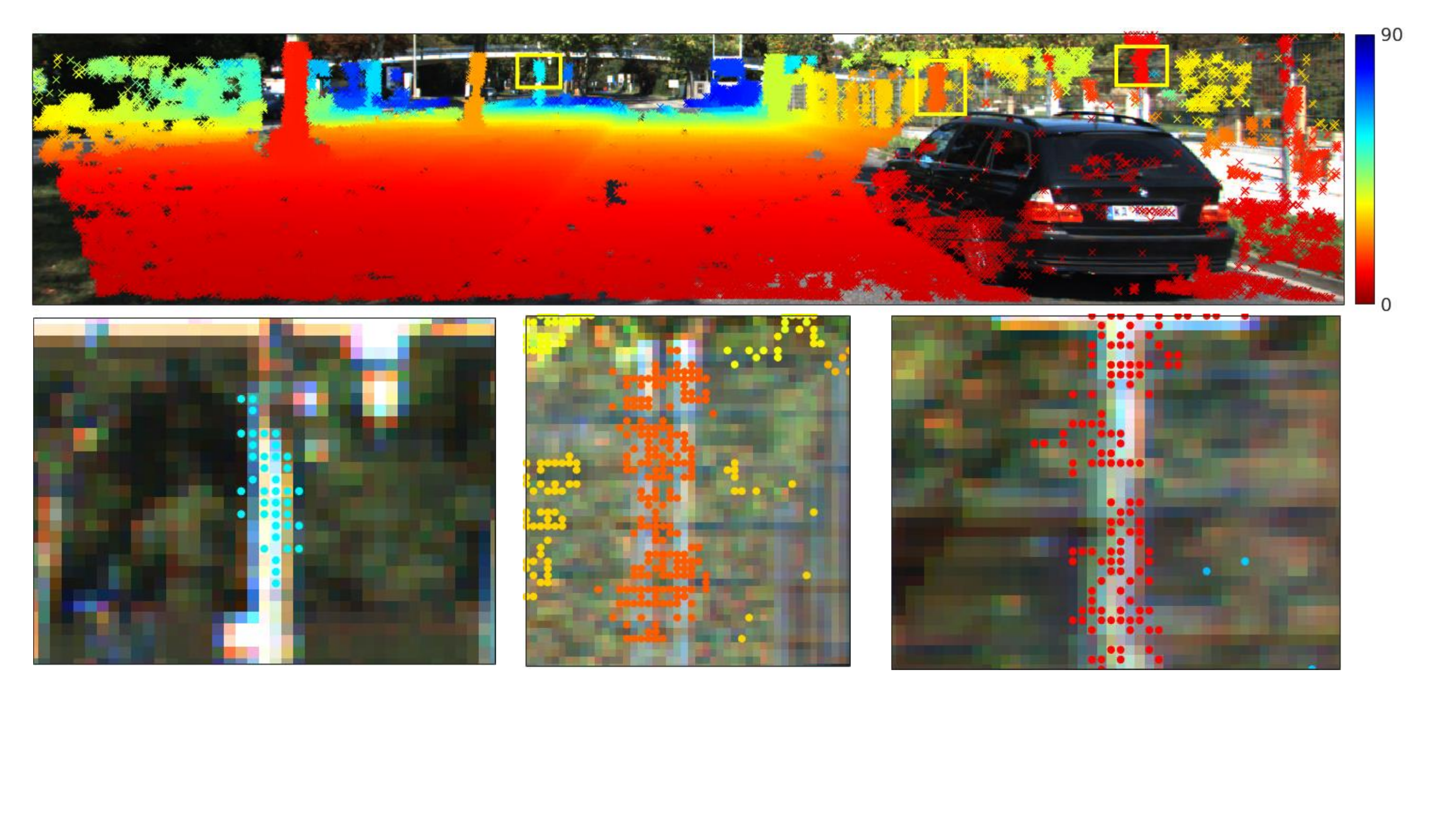}
    \end{tabular}
    \caption{Semi-dense GT depths overlaid on color images. Zoom-in views show foreground/background depths are incorrectly spread (dilated/constricted) across boundaries of poles, traffic signs etc. visible in color images. }
    \label{fig:kitti_outliers}
\end{figure*}

While outliers can be caused by wrong estimation of foreground/background depth, another important source of outliers is incorrect labelling of ground-truth depths in KITTI. As a result, loss functions that are more sensitive to outliers (i.e. MSE loss) can be negatively influenced by the presence of noise. We highlight the noisy ground-truth labels in KITTI in the next section.

\section{Outlier Errors and Analysis on KITTI Semi-Dense GT}
In this section we show some evidence of outliers (noisy ground-truth depth) on boundaries of objects in KITTI's semi-dense GT.

Uhrig \cite{uhrig2017sparsity} proposed an approach \cite{uhrig2017sparsity} to generate large-scale semi-dense GT data ($85k$ training images) on realistic outdoor scenes suitable for neural network training. Although the approach is scalable on any dataset, it creates noisy ground-truth depth. Uhrig's \cite{uhrig2017sparsity} analysis shows that the semi-dense GT has larger errors on dynamic objects and large-range pixels. Additionally, we show that it also contains incorrect depth labels on some boundaries of objects. In both ($a$) and ($b$) of Fig.~\ref{fig:kitti_outliers}, we show zoomed in views of how foreground and background depths that are incorrectly spread across the boundaries of the poles, traffic signs, trees etc. of color images.

Our analysis shows that the outliers in the semi-dense GT are caused by a variety of reasons; 
\begin{itemize}[leftmargin=*,labelindent=4mm,labelsep=6.3mm]
    \item Noisy rotation $R$, and translation $t$ obtained from the IMU sensor
    \item Timing synchronization between camera trigger and time taken to spin one LiDAR revolution
    \item Consistency Check on Stereo-Global Matching algorithm which introduce boundary artifacts
    \item Accumulation of LiDAR points from dynamic objects.
\end{itemize}

In order to evaluate the depth quality of semi-dense GT, Uhrig~\cite{uhrig2017sparsity} used the manually cleaned training set of 2015 KITTI stereo benchmark as reference data. The depth evaluation is done in pixel units. We realize that it is equally important to evaluate the semi-dense ground-truth depths in metric units to notice the \emph{effect of boundary outliers} on semi-dense ground-truth depth metric performance. We translate the error in pixel units to error in metric units in Tab.~\ref{tab:disp2dep}, by converting the ground-truth disparity to depth using KITTI's provided intrinsics. It shows the noisy semi-dense ground-truth depths suffering from boundary noise and dynamic objects can also have significant errors in metric units. It is also a possible indication that lowering the RMSE error in semi-dense GT might result in learning the noise inherent in semi-dense ground-truth.

\begin{savenotes}
\begin{table}[t!]
    \centering
    \begin{tabular}{|c|c|c|c|c|}
        \hline
        \makecell{MAE \\ (in pixel)} & \makecell{RMSE \\ (in pixel)} & \makecell{KITTI \\Outliers*} & \makecell{MAE \\ (in cm)} & \makecell{RMSE \\ (in cm)}  \\
        \hline
        $0.35$ & $0.84$ & $0.31$ & $38.6$ & $94.1$ \\
        \hline
    \end{tabular}\vspace{2mm}
    \caption{Relation between Disparity Error and Depth Error in metric units (cm). Note that KITTI Outliers are defined by: $>3$ pix disparity error and $5\%$ error.}
    \label{tab:disp2dep}
\end{table}
\end{savenotes}

\begin{figure}[t!]
\centering
\begin{tabular}{@{}c@{\hskip1pt}c}
    {\footnotesize($a$)} & \includegraphics[trim=50 160 50 125,clip, width=0.9\linewidth]{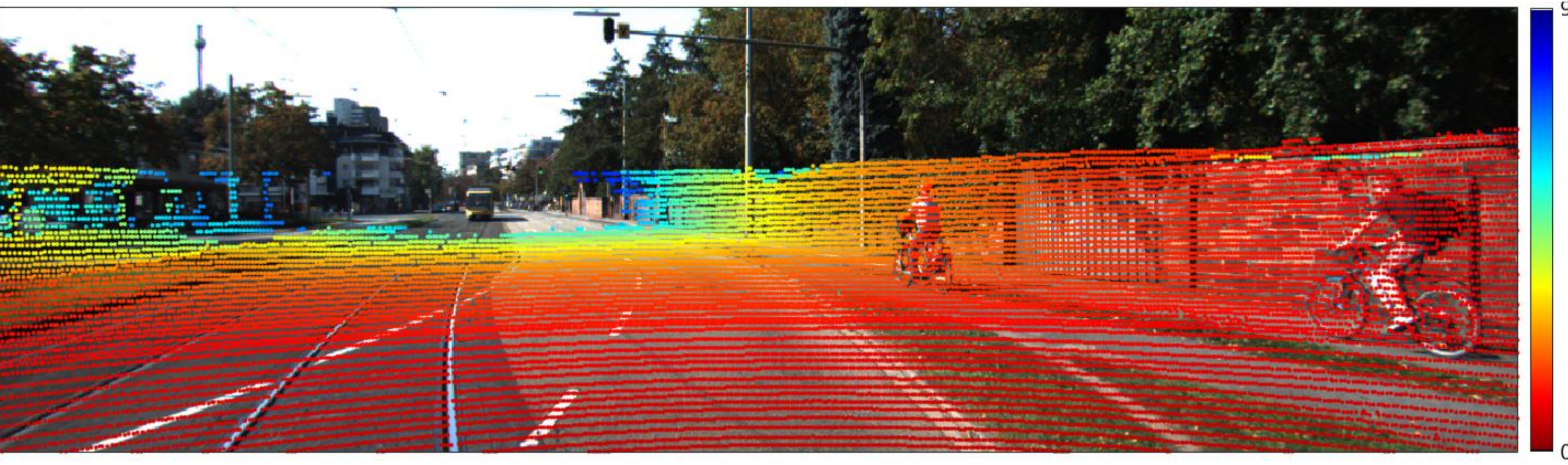}  \\
     {\footnotesize($b$)} & \includegraphics[trim=50 160 50 125,clip, width=0.9\linewidth]{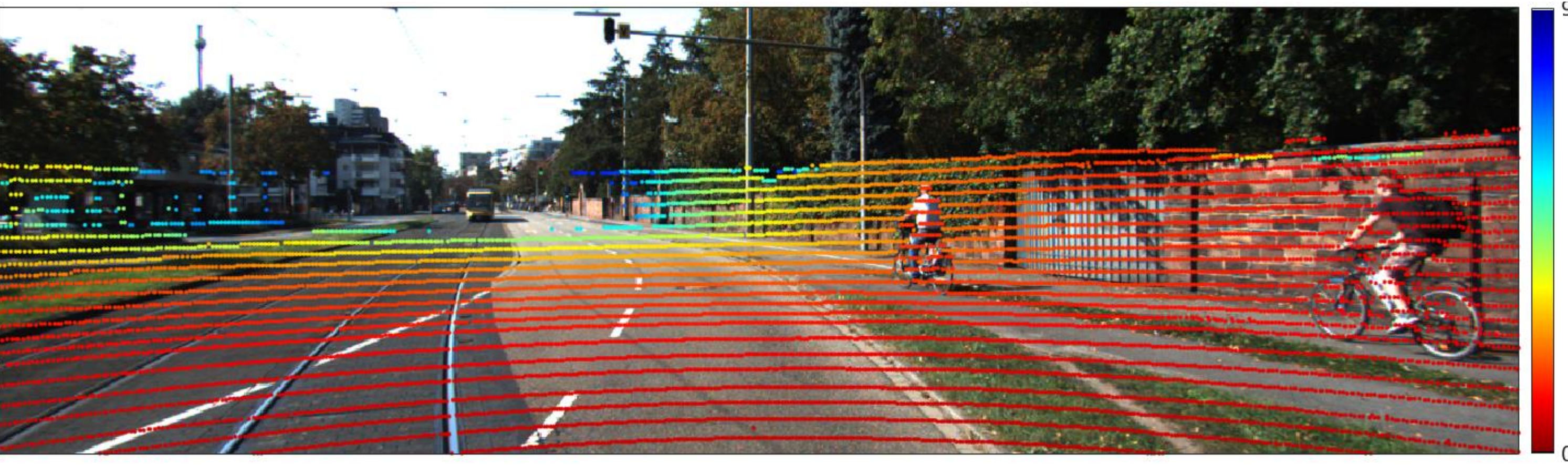} \\
     {\footnotesize($c$)} & \includegraphics[trim=50 160 50 125,clip, width=0.9\linewidth]{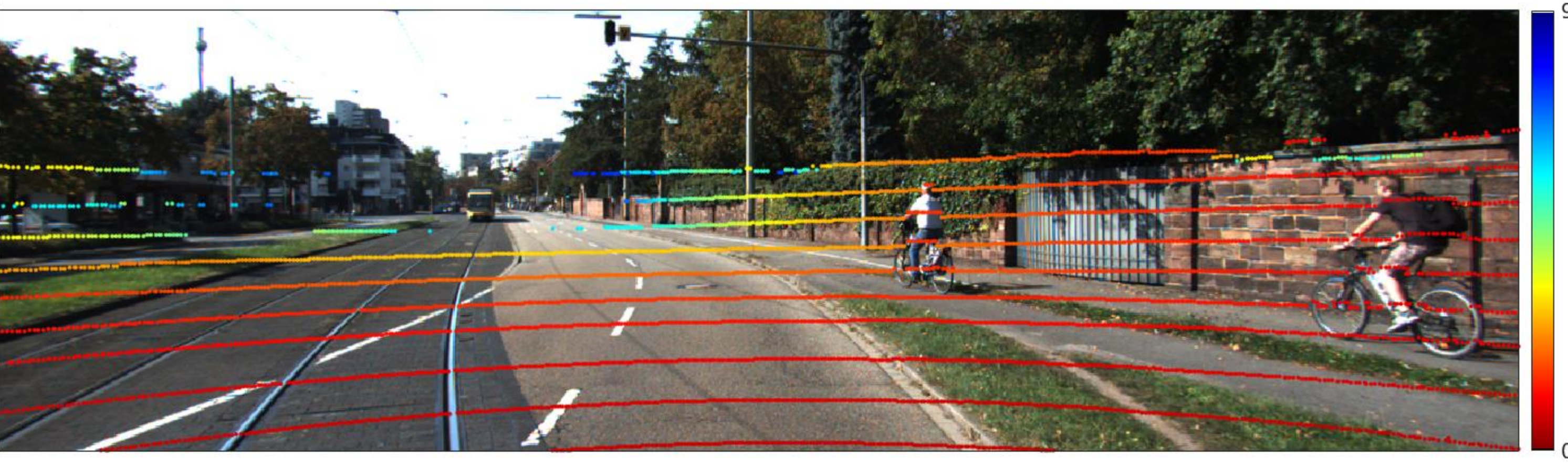} \\
     {\footnotesize($d$)} & \includegraphics[trim=50 160 50 125,clip, width=0.9\linewidth]{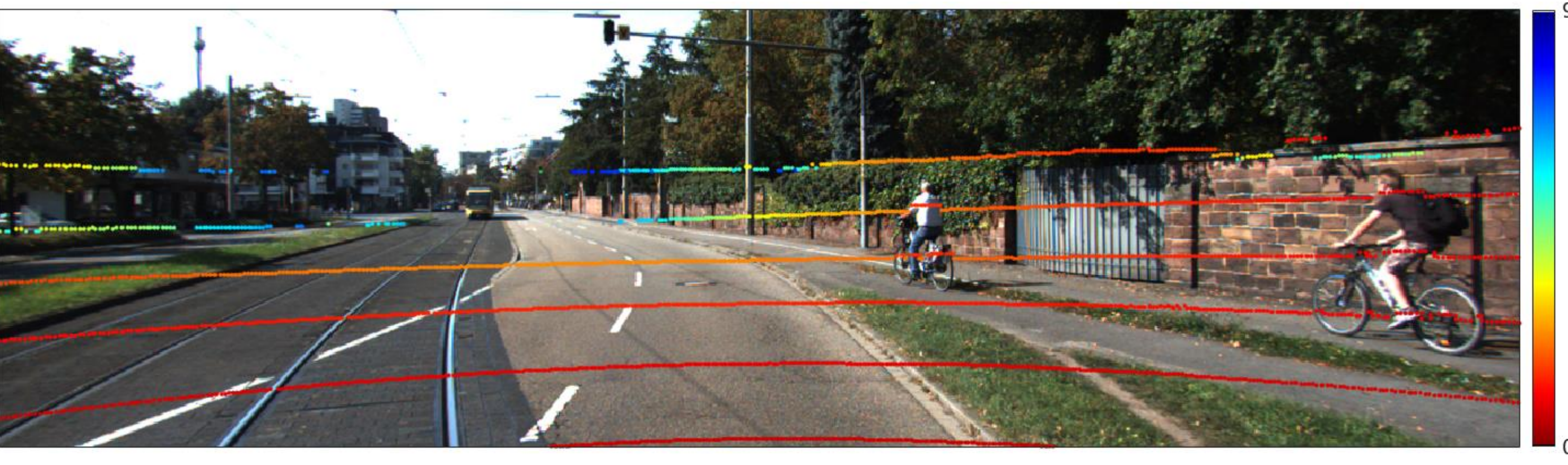} \\
\end{tabular}
\caption{KITTI sparse patterns of ($a$) $64R$, ($b$) $32R$, ($c$) $16R$, and ($d$) $8R$ subsampled LiDAR respectively overlaid on a color image.}
\label{fig:kitti_sparsepattern}
\end{figure}

\begin{figure*}[t!]
    \centering
    \begin{tabular}{@{}c@{\hskip1pt}c@{}c@{\hskip1pt}c}
        {\footnotesize($a$)} & \includegraphics[trim=50 170 50 125,clip, width=0.45\linewidth]{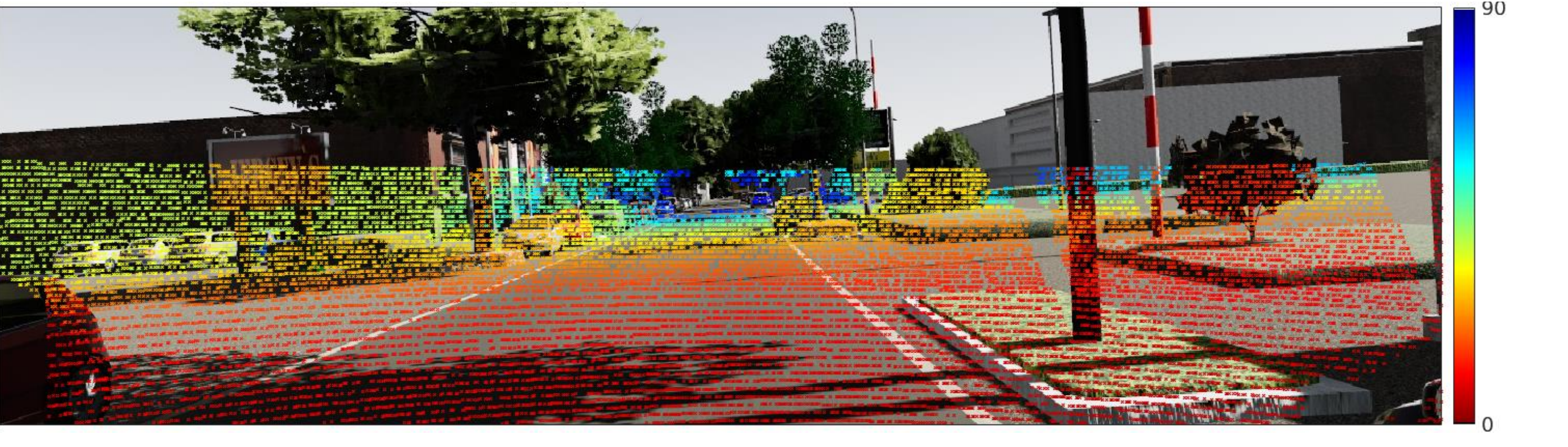} &
        {\footnotesize($b$)} & \includegraphics[trim=50 170 50 125,clip, width=0.45\linewidth]{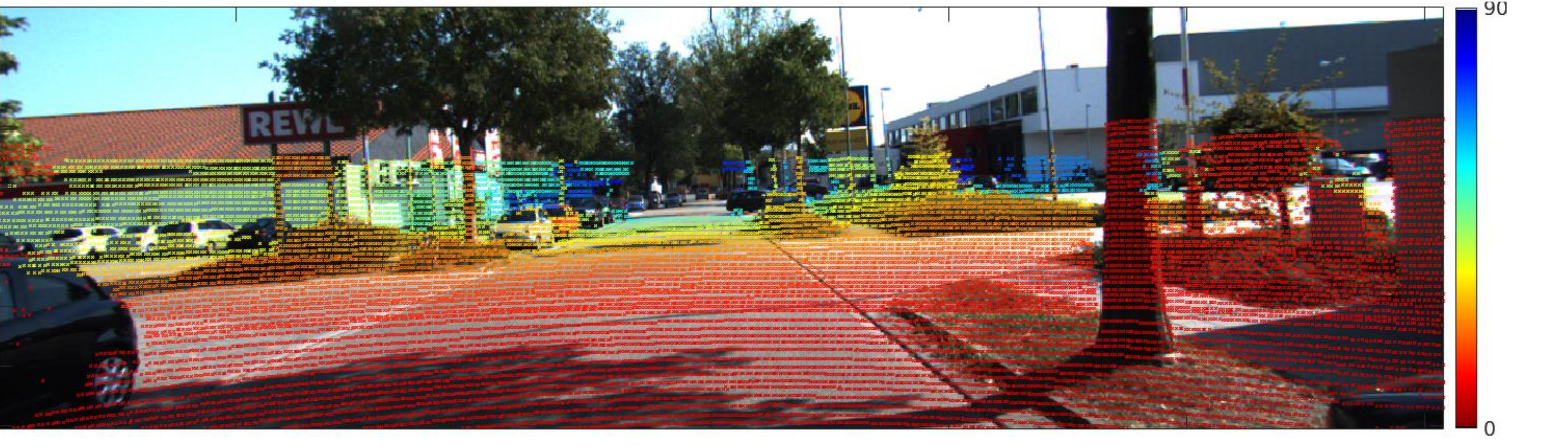} \\
          {\footnotesize($c$)} & \includegraphics[trim=50 170 50 125,clip, width=0.45\linewidth]{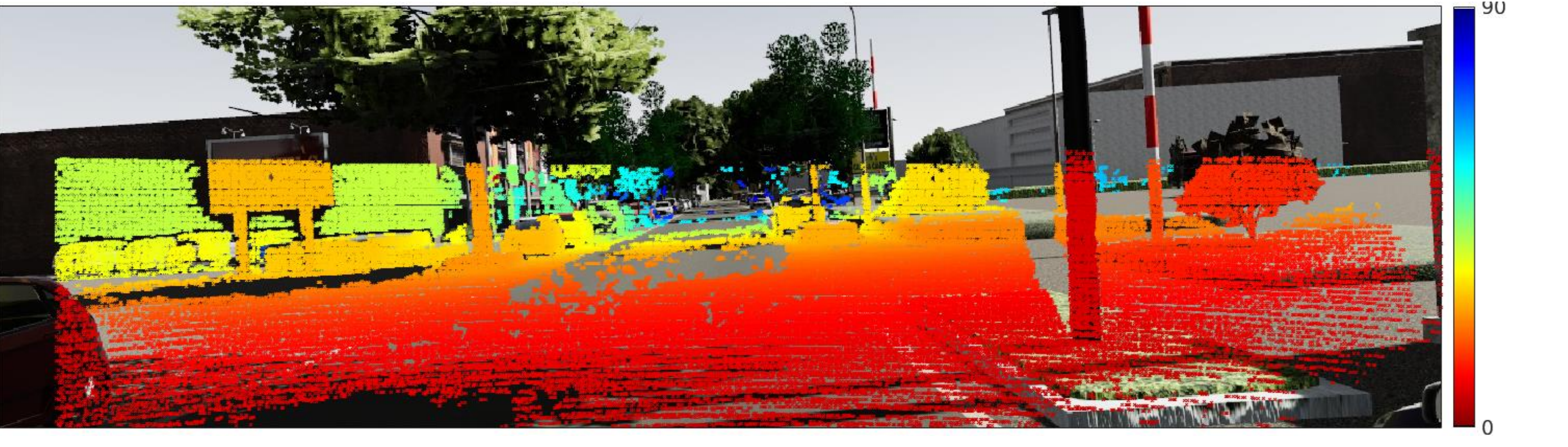} &
        {\footnotesize($d$)} & \includegraphics[trim=50 170 50 125,clip, width=0.45\linewidth]{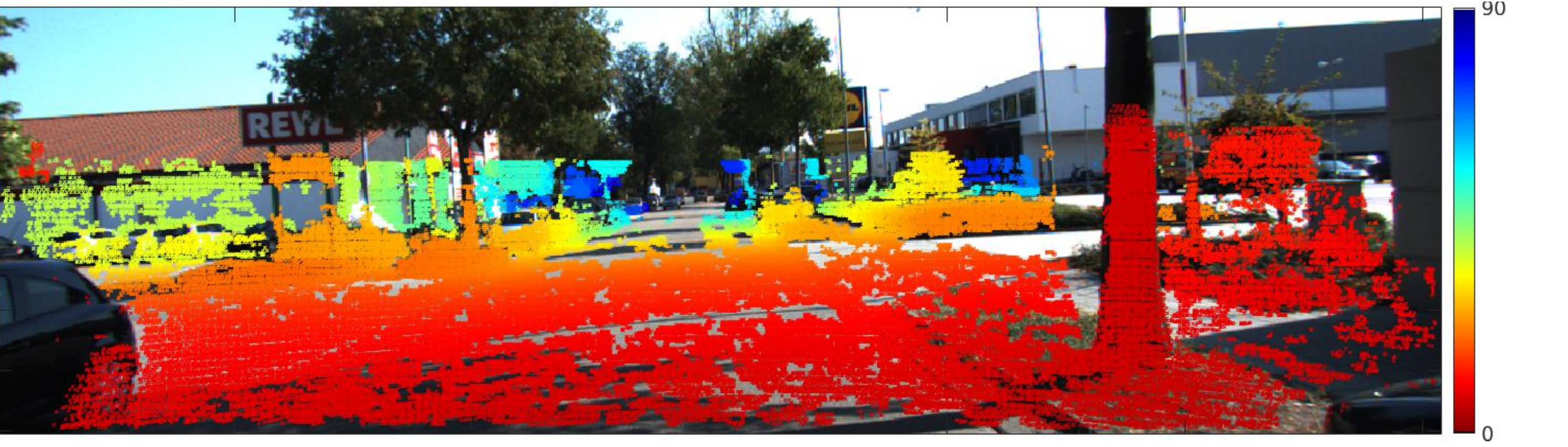} \\
        {\footnotesize($e$)} & \includegraphics[trim=50 170 50 125,clip, width=0.45\linewidth]{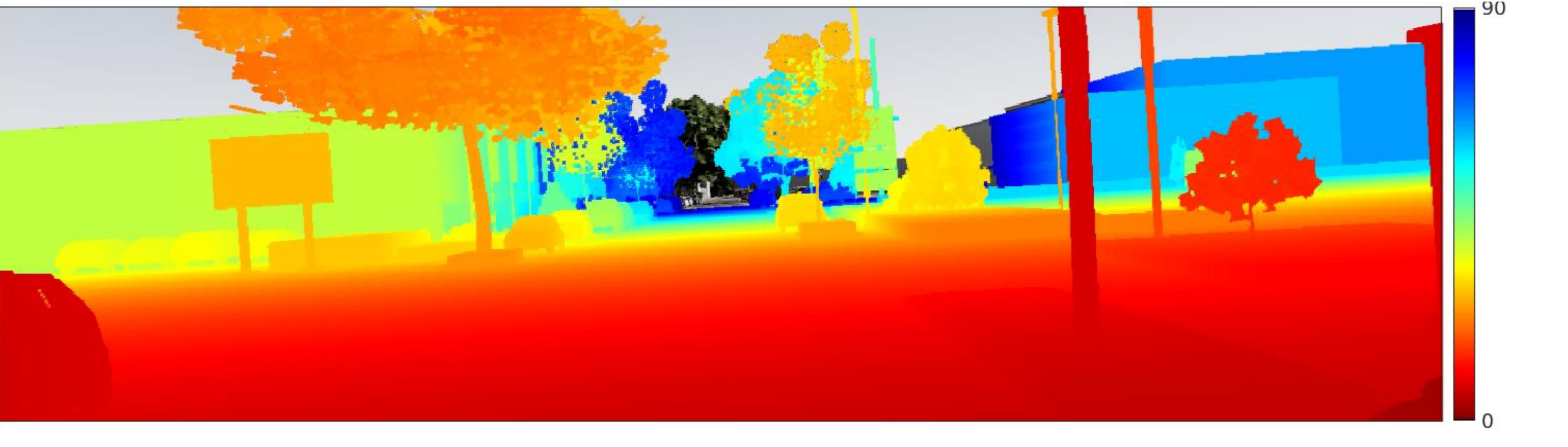} & & 
    \end{tabular}
    \caption{Visual examples of ($a$) sparse depth, ($c$) semi-dense depth and ($e$) dense depth of virtual KITTI. ($b$) and ($d$) shows sparse depth and semi-dense GT of KITTI respectively (shown for comparison with VKITTI data).}
    \label{fig:vkitti_okitti}
\end{figure*}

\begin{figure}[t]
    \centering
    \begin{tabular}{c}
        \includegraphics[trim=120 200 110 120,clip, width=0.9\linewidth]{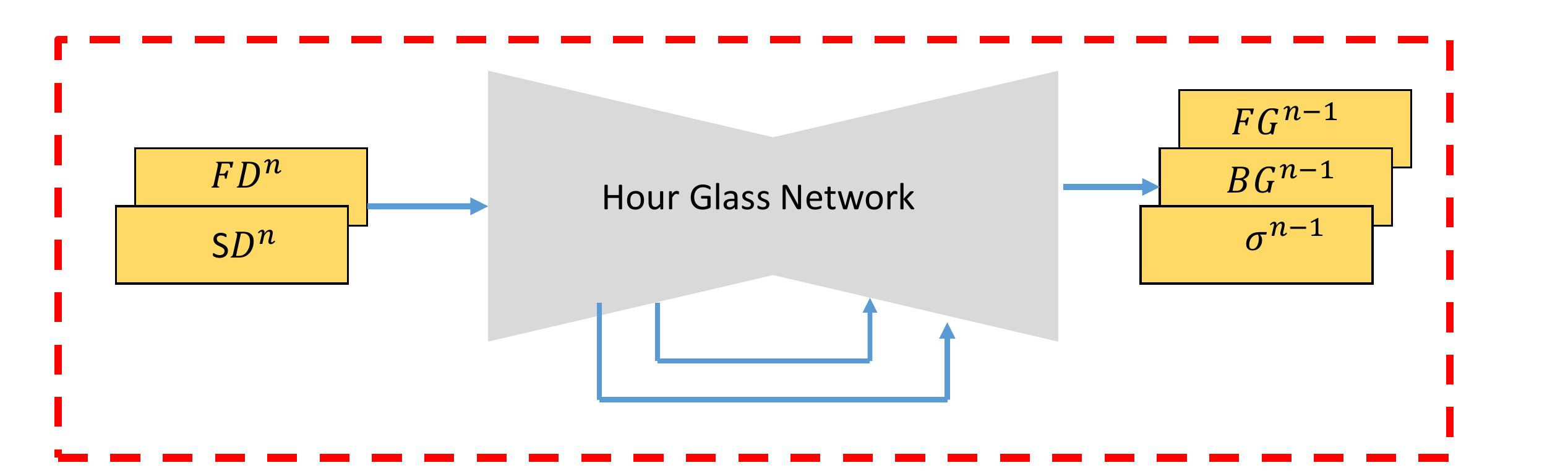} \\
    \end{tabular}
    \caption{Incorporating 3-channel at the output of the Hour-glass network used in \cite{Li_2020_WACV}. $SD^n$ and $FD^n$ are the sparse inputs and fused depth obtained from $FG^n$, $BG^n$, and $\sigma^n$ at multi resolution scale $n$ respectively.}
    \label{fig:net_arch}
\end{figure}

\section{Sparse Patterns in KITTI}
In the main paper, we show the improved generalizability of TWISE over other SoTA methods in terms of sparsity. In this section, we explain how sparsity is created from $64R$ LiDAR in KITTI. Ma {\em et al.}~\cite{ma2019self} reported improved performance with uniform subsampling from KITTI's ground-truth data. But in real scenarios, sparse sensors such as LiDAR often generate non-uniform, structured patterns.  We simulate lower resolution LiDARs by subsampling $32$R, $16$R, $8$R rows from $64$R LiDAR (depth acquisition sensor used by KITTI).  The different sparse patterns can be seen in Fig. \ref{fig:kitti_sparsepattern}. We subsample the points based on selecting a subset of evenly spaced rows of $64$R raw data provided by KITTI (split based on the azimuth angle in the LiDAR space) and then projecting the points into the image. 

\section{Network Architecture}
In the main paper, we mentioned that we used the network of Li {\it et al.}~\cite{Li_2020_WACV} as a backbone network for TWISE. The only modification we made are at the last layer of the network, where we used three channels representing $d_1$ (foreground estimate), $d_2$ (background estimate), and $\sigma$ (see Fig.~\ref{fig:net_arch}). We repeat this strategy in the hourglass networks in all the three multi-resolution levels. Please see \cite{Li_2020_WACV} for more details of the network.

\section{Additional VKITTI Resutls}

\subsection{VKITTI Results on Different Weathers}

The high-resolution color features is an important cue for FG/BG selection in TWISE. We also analyze the effect of different weather conditions that can deteriorate high-resolution boundary cues from color in Tab. \ref{tab:vkitti_weather}. In this study, we found that model trained on 'clone' set is evaluated on different weather conditions in VKITTI.T The performance is largely maintained, with minor degradations in fog and rain. 
\begin{table}[t!]
    \centering
    \begin{tabular}{|@{}c@{}|c|c|c|c|}
    \hline
        RGB Mode & MAE & RMSE & TMAE & TRMSE \\
        \hline
        Clone & $12.71$ & $126.40$ & $5.22$ & $16.67$
        \\
        \hline
         Morning & $12.99$ & $130.90$ & $5.17$ & $16.60$ 
        \\
        \hline
        Fog & $13.19$ & $131.97$ & $5.15$ & $16.74$ 
        \\
        \hline
        Sunset & $12.77$ & $129.50$ & $5.10$ & $16.50$\\
        \hline
        Rainy & $13.08$ & $132.09$ & $5.17$ & $16.67$ \\
        \hline
        OverCast & $12.48$ & $126.82$ & $5.08$ & $16.47$ 
        \\
        \hline
    \end{tabular}\vspace{2mm}
    \caption{VKITTI Results on different weather conditions}
    \label{tab:vkitti_weather}
\end{table}
It shows although the low-quality RGB (low contrast, shadows, fog, rain etc) might create ambiguity and the blending coefficient fail to correctly select FG/BG, it is possible to detect boundary information using sufficient training examples.

\subsection{Creating Semi-Dense and Sparse Depth from Dense VKITTI GT}

In the main paper, we performed an ablation study on Virtual KITTI \cite{cabon2020virtual} (VKITTI) using semi-dense and sparse samples created from dense VKITTI depth maps. We created semi-dense VKITTI to simulate outlier noise similar to that existing in real KITTI dataset. In this section, we discuss the data generation process in detail and show some visual examples of how the sparse depth/semi-dense compares with sparse/semi-dense gt of KITTI dataset in Fig.~\ref{fig:vkitti_okitti}.

The dense ground-truth depth maps from VKITTI contains accurate depth on object discontinuities. Using this as a reference, we subsampled the ground-truth depth maps. Instead of uniformly subsampling the GT depth, we subsample the LiDAR in the azimuth-elevation coordinates to make the input sparse depth resemble structured patterns found in original LiDAR (see ($a$) and ($b$) of Fig. \ref{fig:vkitti_okitti}). The subsampled depth from the left camera is then projected to the right camera, and vice versa to simulate LiDAR points projected onto images in real-world scenes. For supervision, GT depth beyond $90$m are suppressed to simulate LiDAR points with no returns (see ($e$) of Fig.~\ref{fig:vkitti_okitti}). 
In addition to supervision using clean ground-truth present, we also perform supervision on Semi-Dense GT of VKITTI (Fig.~$10$ of the main paper) created by simulating outliers existing in original KITTI dataset \cite{uhrig2017sparsity}. In the KITTI dataset,  semi-dense GT is created by accumulating LiDAR points from $+/-5$ frames from the reference frame. 
We follow the similar procedure as followed by~\cite{uhrig2017sparsity} when creating semi-dense GT. Additionally, we add Gaussian noise to model noisy $R$, $t$ from the IMU sensor to simulate noisy semi-dense GT. Refer to Fig.~\ref{fig:vkitti_okitti} for a comparison between semi-dense VKITTI and semi-dense KITTI (see ($c$) and ($d$) of Fig. \ref{fig:vkitti_okitti}).

\begin{table}[t!]
    \centering
    \scalebox{0.70}{
    \begin{tabular}{|c|c|c|c|c|c|}
    \hline
        Dataset & $R$, $t$ & Outliers\% & Pix. Coverage\% & MAE (cm) & RMSE (cm)  \\
        \hline
        \multirow{1}{*}{KITTI} & IMU & $4.4$ & $16$ & $38.6$ & $94.1$ \\ 
        \hline
        \multirow{2}{*}{VKITTI} & Clean $R$, $t$ & $3.0$ & $20$ & $19.3$ & $128.96$ \\
        & Noisy $R$, $t$ & $4.1$ & $18$ & $29.3$ & $145.18$ \\ 
        \hline
    \end{tabular}}
    \caption{Comparison of VKITTI semi-dense errors with KITTI semi-dense GT errors. Higher errors in RMSE in the VKITTI dataset is due to dense depth pixels at far-away points, contrary to KITTI's stereo benchmark data which is sparse. }
    \label{tab:vkitti_semidenseout}
\end{table}

\subsection{Relation to KITTI GT by Outliers}
We define outliers as pixels having depth errors greater than $1$m, contrary to KITTI outliers in Tab.~\ref{tab:disp2dep} which define errors in pixel units. Evaluated on KITTI's $2015$ stereo benchmark depth data, we found outliers of KITTI's semi-dense ground-truth at $4.4\%$ of the inlier depths. We created outliers in semi-dense VKITTI by introducing Gaussian noise in VKITTI's extrinsics. See Tab.~\ref{tab:vkitti_semidenseout} for a metric comparison with outliers. Tab.~\ref{tab:vkitti_semidenseout} shows that, as we add noisy in $R$, $t$, the semi-dense GT of VKITTI is more comparable to KITTI semi-dense GT.

\section{Video}
We provide a video in the supplementary material. The video shows point-cloud rendered from estimated depth maps of TWISE and MultiStack \cite{Li_2020_WACV}. It shows point-cloud generated from MultiStack contains significantly more mixed depth pixels (compare the floating depth pixels in the pointclouds) compared to TWISE.  

\printbibliography